\documentclass[10pt,journal]{IEEEtran}
\usepackage[T1]{fontenc}
\usepackage[utf8]{inputenc}
\usepackage{amsmath,amsfonts}
\usepackage{booktabs}
\usepackage[table]{xcolor}
\usepackage{graphicx}
\usepackage{multirow}
\usepackage{array}
\usepackage{tabularx}
\usepackage{makecell}
\usepackage{nicefrac}
\usepackage[patch=none]{microtype}
\usepackage{listings}
\usepackage{enumitem}
\usepackage{xurl}
\usepackage{xspace}
\usepackage{tcolorbox}
\usepackage{algorithm}
\usepackage{algpseudocode}
\usepackage{etoc}
\usepackage{titletoc}
\usepackage{stfloats}
\usepackage{cite}
\usepackage[hidelinks]{hyperref}
\usepackage{xr-hyper}

\definecolor{neuralRIshade}{RGB}{226,239,250}
\definecolor{hemoshade}{RGB}{238,244,249}
\definecolor{deepPurple}{RGB}{110,70,170}
\definecolor{softPurple}{RGB}{140,110,200}
\definecolor{rosePink}{RGB}{210,90,140}
\definecolor{softPink}{RGB}{230,130,170}
\definecolor{accentBlue}{RGB}{70,130,200}
\definecolor{vividPurple}{RGB}{120,0,220}
\definecolor{myPurple}{RGB}{102,51,153}
\definecolor{myPink}{RGB}{220,80,140}
\definecolor{codeblue}{RGB}{0,0,180}
\definecolor{codebg}{RGB}{245,245,245}

\newcommand{\fixedcomp}[1]{{\sffamily\itshape #1}}
\newcommand{\hemo}{\textsc{HEMO}\xspace}
\newcommand{\module}[1]{{\sffamily\bfseries\small #1}}
\newcommand{\suppheading}[1]{%
  \par\addvspace{0.45\baselineskip}%
  \noindent\textbf{#1}\enspace\ignorespaces
}

\tcbuselibrary{breakable,skins}
\newtcolorbox{promptbox}[1][]{%
  enhanced,
  colback=blue!3,
  colframe=blue!40!gray!60,
  fonttitle=\bfseries\small\sffamily,
  coltitle=blue!40!black,
  attach boxed title to top left={yshift=-2mm,xshift=4mm},
  boxed title style={
    colback=blue!10,
    colframe=blue!40!gray!60,
    boxrule=0.4pt,
    arc=1.5pt
  },
  boxrule=0.5pt,
  arc=2pt,
  left=6pt,
  right=6pt,
  top=6pt,
  bottom=4pt,
  fontupper=\small,
  breakable,
  title={#1},
}
\makeatletter
\newcommand*{\addFileDependency}[1]{%
  \typeout{(#1)}%
  \@addtofilelist{#1}%
  \IfFileExists{#1}{}{\typeout{No file #1.}}%
}

\makeatother

\title{HEMO: Harness Self-Evolution for Multi-Objective Combinatorial Optimization}
\author{Jianghan Zhu, Cong Zhang, Ni Zhang, Yifan Yang, Yew Soon Ong, and Zhiguang Cao%
\thanks{Jianghan Zhu, Ni Zhang, and Zhiguang Cao are with the School of Computing and Information Systems, Singapore Management University, Singapore. (E-mail: \href{mailto:zhuj0044@ntu.edu.sg}{zhuj0044@ntu.edu.sg})}%
\thanks{Yifan Yang, and Yew Soon Ong are with the College of Computing and Data Science, Nanyang Technological University, Singapore.}%
\thanks{Cong Zhang used to be affiliated with Nanyang Technological University, Huawei Noah’s Ark Lab,
and TikTok Singapore. Now he is an independent researcher.}
}

\begin{document}
\maketitle
\begin{abstract}
Automatic operator design (AOD) has increasingly adopted large language models (LLMs) to generate problem-specific search and variation operators for multi-objective combinatorial optimization (MOCO).
Yet most methods evolve the operators while leaving the process that discovers them fixed. This limits adaptation as new Pareto trade-offs and operator interactions change where and how the search should proceed. We introduce \textbf{\hemo}, which makes the AOD harness itself a self-evolving optimization target. \hemo{} maintains a persistent harness state and updates one module at a time, revising the logic that determines which
experience to retrieve, which operators to improve, which parents to select, and how to formulate the orchestration of search steps. Downstream solution quality and runtime guide these updates, while a discounted UCB scheduler directs adaptation toward promising modules as their utility changes. A persistent memory filesystem preserves candidate outcomes and diagnostic feedback across harness states, enabling successive states to build on accumulated optimization experience. Experiments across eight MOCO tasks and two structurally distinct evolutionary backbones demonstrate strong performance against conventional evolutionary, neural, and LLM-based AOD baselines, together with effective transfer to related tasks. \hemo{} extends automatic operator design beyond evolving individual operators to improving the search process that discovers them.

\end{abstract}

\begin{IEEEkeywords}
Evolutionary computing and genetic algorithms, heuristics design, heuristic methods, language models, multi-objective combinatorial optimization, optimization.
\end{IEEEkeywords}

\section{Introduction}
\label{sec:introduction}

\IEEEPARstart{M}{ulti-objective} combinatorial optimization (MOCO) arises in many discrete
decision problems with competing objectives, including routing, scheduling,
and resource allocation~\cite{jozefowiez2008multi,hoogeveen2005multicriteria,lust2012multiobjective,zajac2021objectives}.
A spectrum of methods has been developed for these problems, including exact approaches~\cite{ehrgott2016exact}, Pareto local search and related metaheuristics~\cite{blot2018survey}, and population-based evolutionary algorithms such as NSGA-II and 
MOEA/D~\cite{deb2002fast,zhang2007moea}. Despite their different search mechanisms, performance often relies on problem-specific representations, and variation operators, making the 
solving of a new MOCO problem closely tied to the design of an effective search procedure.

Learning-based combinatorial optimization has sought to reduce this manual effort by learning solution policies directly from data~\cite{lin2022pareto,chen2023neural,wang2024asp}
while automatic algorithm design (AAD), particularly automatic heuristic design (AHD), instead automates the generation and improvement of algorithmic
components~\cite{burke2013hyper}. 
Recent LLM-based methods further enable
evolutionary search over algorithms and heuristics~\cite{romera2024mathematical,
liu2024evolution,ye2024reevo}.
Within MOCO, a particularly relevant direction is the automatic design of
problem-specific search and variation operators, which we refer to as
automatic operator design (AOD). LLMs have been used to generate evolutionary operators with different trade-offs~\cite{huang2025autonomous,yao2025multi},
while newer frameworks further emphasize population diversity and operator interactions~\cite{hieu2026pareto,qiu2026e2oc}. These developments mark a shift from generating isolated heuristics toward designing structured optimization components.

\begin{figure*}[!t]
    \centering
    \includegraphics[width=0.85\textwidth]{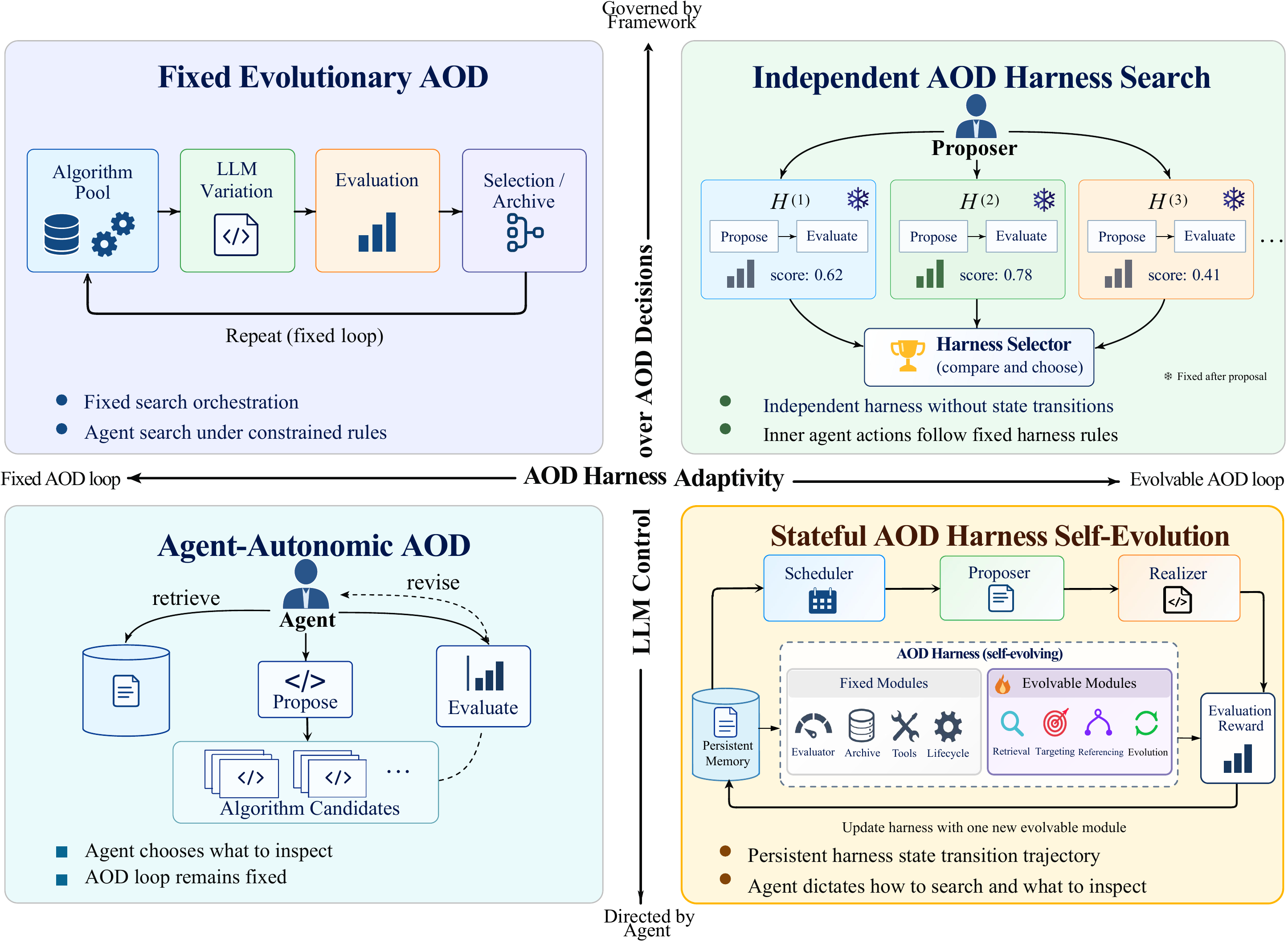}
    \caption{
    Paradigms of LLM-based AOD characterized by agent autonomy and harness adaptivity. Fixed evolutionary AOD (upper left) uses agents as role-specific tools
    within a predefined evolutionary loop, whereas independent AOD harness
    search (upper right) evaluates separately sampled but internally fixed
    harnesses.
    Agent-autonomic AOD (lower left) allows the agent to control candidate
    generation and feedback inspection within a fixed outer AOD loop, while
    stateful AOD harness self-evolution, adopted by \hemo in the lower-right
    quadrant, evolves the stateful harness itself through persistent
    module-level transitions, allowing search control and the use of
    accumulated experience to co-adapt over time under outer-agent guidance.
    }
    \label{fig:aad_paradigms}

\end{figure*}

LLM-based AOD relies on an outer control process that determines how design context is formed, when candidates are evaluated, how search state is recorded, and when search terminates. We call this control process the \emph{AOD harness}, specializing the broader notion of an LLM harness as executable infrastructure governing model context, environment interactions, and control flow~\cite{lee2026metaharnessendtoendoptimizationmodel,li2026agent} to the search over operator candidates.
Most evolutionary AOD methods use a largely fixed harness, with the main operator search steps governed by predefined procedures. For example, AlphaEvolve retains a fixed search framework in which the model context is assembled through predefined selection rules~\cite{novikov2025alphaevolve}. Greater LLM autonomy can also be introduced within a fixed search loop, whereas recent methods search over independently generated abstraction-level heuristic-optimizers that serve as fixed search controllers for the inner AHD process~\cite{shi2026generalizable}. Figure~\ref{fig:aad_paradigms} summarizes these regimes along two dimensions, namely \emph{agent autonomy in search control} and \emph{AOD harness adaptivity}.

This distinction matters for MOCO-oriented AOD, where the search state changes continuously as new trade-offs, candidate structures, and component interactions emerge. The effectiveness of a search decision can vary over time, and a strategy that works well at one stage may become less suitable as the explored regions and candidate population change. 
A fixed AOD harness cannot revise its predefined control logic, while independently generated high-level procedures lack a persistent control-state trajectory. This motivates a stateful AOD harness that evolves its search behavior as optimization experience accumulates. Such self-evolution also requires access to relevant long-term experience. An AOD run may produce hundreds or thousands of evaluated candidates with their performance, lineage, and search context, yet storing this history alone does not ensure useful evidence remains accessible. Recent harness research likewise emphasizes detailed execution records and selective retrieval in long-running searches~\cite{lee2026metaharnessendtoendoptimizationmodel,qu2026coral}. For MOCO, relevant evidence can span Pareto regions, complementary candidates, and earlier search stages, making persistent, selectively retrieved experience a stable basis for adapting harness decisions.

Motivated by these observations, we introduce \textbf{\hemo}, a stateful AOD harness for LLM-based operator design in multi-objective combinatorial optimization. \hemo maintains a persistent harness state and progressively adapts its search control through localized state transitions. Successive harness states are assessed by their downstream impact on multi-objective operator design, while optimization experience from evaluated candidates is continuously retained in a \emph{memory filesystem} and selectively reused by subsequent states. This formulation allows the AOD harness to adapt across different search stages while the memory filesystem evolves with the trajectory and preserves a consistent archive of optimization experience, thereby improving how an LLM-driven operator design process searches, learns from feedback, and reuses experience over time.

Our main contributions are summarized as follows:
\begin{itemize}
    \item We introduce a \emph{stateful harness self-evolution} perspective for LLM driven AOD. The AOD harness maintains a persistent state and evolves through local updates along a single harness design trajectory, extending beyond  fixed outer procedures, agent-directed search within fixed harnesses, and independently evaluated high-level search procedures.

    \item We develop \hemo, a framework designed specifically for MOCO using downstream multi-objective AOD performance to guide harness state adaptation while preserving long-term experience in a memory filesystem.   
    Later search decisions can thus draw on evidence from previous harness states without compressing the full history into a limited recent context.

    \item Extensive experiments on eight MOCO tasks spanning routing, traveling salesman, knapsack, and scheduling, using two representative evolutionary backbones, demonstrate the effectiveness of \hemo across diverse settings against conventional evolutionary, neural, and LLM-based AOD baselines in both within-task evaluation and cross-task transfer.
\end{itemize}

\section{Related Work}
\label{sec:related_work}

\subsection{Automated Algorithm Design}
\label{sec:related_aad}

Automated algorithm design (AAD) aims to reduce the expert effort required to construct effective algorithms by automating design processes such as algorithm selection, configuration, component design, and code execution and validation~\cite{pillay2021automated,langdon2002foundations,zhang2026agentic}. Automated heuristic design (AHD) forms an important research branch of this broader field, with methods ranging from hyper-heuristics~\cite{burke2013hyper} and genetic programming~\cite{langdon2002foundations} to the automatic generation and composition of search operators~\cite{xu2026dreamcop}. Although these approaches automate different parts of algorithm development, the underlying heuristic spaces and composition rules are commonly specified by human experts.

LLM reasoning and code generation have substantially expanded the scope of automated design. FunSearch evolves executable programs through island-based selection and generation~\cite{romera2024mathematical}, while EoH, ReEvo, and MCTS-AHD combine LLM generation with evolutionary, reflective, or tree-based search to improve one single-objective optimization heuristic component~\cite{liu2024evolution,ye2024reevo,zheng2025monte}. SimpleEvol further organizes LLM-driven heuristic generation within a lightweight agent loop designed to minimize human-specified priors~\cite{zhu2026simpleevol}. Beyond individual combinatorial heuristics, LLaMEA and LLaMEA-BO evolve
complete algorithms for continuous black-box and Bayesian optimization,
respectively~\cite{vanstein2025llamea,li2026llameabo}. Outside optimization, AlphaEvolve demonstrates evolutionary code search over broader scientific and computational algorithms~\cite{novikov2025alphaevolve}. 
More recent work has begun to move the design target beyond solely heuristics.
MoH searches over heuristic optimizers that generate task-specific
heuristics~\cite{shi2026generalizable}, while related work explores
higher-level algorithm structures and control
policies~\cite{ma2025toward}.
These developments indicate a broader shift from designing isolated heuristics toward changing increasingly large portions of the algorithm design process itself. Yet the execution structure governing LLM
access to accumulated experience, forms design context, invokes evaluation, and maintains search state is still typically specified by human designers. Designing a framework that enables the stateful self-evolution of this
surrounding control structure is therefore the focus of our work.

\subsection{Multi-Objective Optimization}
\label{sec:related_moo}

Most of recent research can be divided into two pathways. The first improves
multi-objective solvers through manually designed evolutionary mechanisms
or learned solution policies and search components. Classical multi-objective evolutionary
algorithms (MOEAs) include dominance-based methods such as
NSGA-II~\cite{deb2002fast}, decomposition-based methods like
MOEA/D~\cite{zhang2007moea}, and archive- or local-search-based approaches
such as Pareto Local Search (PLS)~\cite{paquete2004pareto}, two-phase
Pareto Local Search (2PPLS)~\cite{lust2012multiobjective}, and Simple
Evolutionary Multi-objective Optimization
(SEMO)~\cite{laumanns2004running}. Their effectiveness depends strongly
on carefully designed mechanisms for selection, population management,
neighborhood exploration, and diversity preservation, motivating
numerous modified and hybrid
variants~\cite{verma2021comprehensive,xu2023pareto}. Neural approaches
instead learn solution policies or search components from data. PMOCO employs a single preference-conditioned model to generate solutions under different objective preferences~\cite{lin2022pareto}. NHDE explicitly promotes Pareto-set diversity~\cite{chen2023neural}, while EMNH improves learning efficiency through meta-learning~\cite{chen2023efficient}.Neuro-PLS embeds learned components into Pareto local search~\cite{zhang2025neuro}, while other
work improves preference conditioning and cross-size generalization through learned weight representations~\cite{chen2025rethinking}. Although neural methods reduce repeated handcrafting and enable fast GPU-based inference, they generally require model training and retain a largely predefined optimization backbone.

The second pathway uses LLMs to design problem-specific heuristics and
variation operators within prescribed multi-objective solvers, a setting
we refer to as automatic operator design (AOD). Here, an operator is an
algorithmic component that constructs, modifies, selects, or otherwise
guides candidate solutions, including heuristics studied in AHD when they serve such a role. Huang et al.~\cite{huang2025autonomous} use LLMs
to iteratively refine evolutionary operators. MEoH treats heuristic
design as a multi-objective optimization problem and searches for
non-dominated heuristics~\cite{yao2025multi}, while REMoH integrates
LLM-generated operators into NSGA-II with clustering-based reflection
to promote diversity~\cite{fornies2025remoh}. MPaGE combines LLM-based
heuristic generation with Pareto-front-grid-guided population management
under SEMO~\cite{hieu2026pareto}. E2OC further models dependencies among
multiple search operators and uses Monte Carlo tree search to explore
their design combinations~\cite{qiu2026e2oc}. These methods extend
LLM-based AAD toward increasingly complex multi-objective solver
components. However, they primarily evolve operators \emph{within} a prescribed
search process, leaving its higher-level harness unchanged.

\subsection{Harness Engineering and Evolution}
\label{sec:related_harness}

The performance of an LLM-based system depends not only on its backbone model, but also on
the surrounding execution machinery that determines what information the model receives, which tools it can invoke, how intermediate state is recorded and updated, and how multi-step execution is guided by feedback. Mechanisms now associated with this layer were initially studied
separately, including retrieval-augmented
generation~\cite{lewis2020retrieval}, interleaved reasoning and acting~\cite{yao2022react}, and explicit self-reflection~\cite{shinn2023reflexion}. Recent work increasingly
integrates these mechanisms into a broader \emph{agent harness}. Natural-Language Agent Harnesses separate control, contracts, and
persistent state from the underlying runtime~\cite{pan2026natural}, while recent surveys characterize harness engineering as spanning execution environments, tool interfaces, context and memory management, lifecycle and orchestration, execution monitoring, and output validation~\cite{li2026agent}. 
A closely related line of work also automates broader portions of agent and optimization
systems~\cite{zeng2016structured,dong2026evolutionary}. ADAS searches over code-defined agent systems~\cite{hu2025automated}, AFlow formulates agentic workflow generation as search over executable workflow graphs~\cite{zhang2025aflow}, and the Darwin G\"odel Machine allows
coding agents to iteratively modify their own
implementations~\cite{zhang2026darwin}. In optimization, AutoEP uses an
external LLM-based controller to adapt the hyperparameters of a
structurally fixed metaheuristic~\cite{xu2026autoep}, while
OKAEM adapts learned evolutionary operators using prior and online
optimization knowledge~\cite{wang2026learning}. CORAL employs
autonomous single- and multi-agent evolution to determine what
experience to retrieve, when to evaluate candidates, and what knowledge
to preserve~\cite{qu2026coral}. Although these approaches automate
context construction, workflows, and search control, they do not
generally treat the complete task-specific execution harness as the
direct search object.

More recent studies explicitly optimize the harness surrounding an LLM.
Meta-Harness performs outer-loop search over harness code, allowing an agentic proposer to inspect previous source code, evaluation scores, and execution traces through a filesystem before constructing new harness candidates~\cite{lee2026metaharnessendtoendoptimizationmodel}.
Self-Harness similarly identifies recurring weaknesses from execution
traces, proposes targeted modifications, and validates the resulting
harness~\cite{zhang2026selfharnessharnessesimprove}. Existing harness
evolution studies primarily target general agent execution in coding,
reasoning, and interactive tasks. \hemo instead studies harness
evolution for LLM-based automatic operator design in MOCO. It maintains
a persistent AOD harness whose decision modules adapt through localized
state transitions while the execution-critical scaffold remains fixed,
connecting harness adaptation directly to downstream multi-objective
optimization performance and long-term operator design experience.

\section{Preliminaries}
\label{sec:preliminaries}

\subsection{Multi-Objective Combinatorial Optimization}
\label{sec:prelim_mocop}

\noindent\textbf{Problem Formulation and Pareto Optimality.}
A multi-objective combinatorial optimization problem (MOCOP) seeks to optimize multiple, potentially conflicting objectives over a discrete feasible space~\cite{ehrgott2005multicriteria}. Let $\mathcal{X}$ be the feasible solution set and $M$ the number of objectives. Without loss of generality, we consider the minimization form
\begin{equation}
    \min_{\mathbf{x}\in\mathcal{X}}
    \mathbf{f}(\mathbf{x})
    =
    \left(
        f_1(\mathbf{x}),
        f_2(\mathbf{x}),
        \ldots,
        f_M(\mathbf{x})
    \right).
    \label{eq:mocop}
\end{equation}
For two feasible solutions $\mathbf{x},\mathbf{y}\in\mathcal{X}$, $\mathbf{x}$ is said to \emph{Pareto-dominate} $\mathbf{y}$, denoted by $\mathbf{x}\prec\mathbf{y}$, if
\begin{equation}
    f_m(\mathbf{x}) \leq f_m(\mathbf{y}),
    \quad \forall m\in\{1,\ldots,M\},
\end{equation}
and the inequality is strict for at least one objective. A feasible solution $\mathbf{x}^{*}$ is \emph{Pareto optimal} if there exists no other feasible solution that dominates it. The collection of all Pareto-optimal solutions constitutes the \emph{Pareto set} (PS), and its image in the objective space constitutes the \emph{Pareto front} (PF).

For many MOCOPs, computing the complete Pareto set is computationally prohibitive, and practical solvers therefore seek a finite approximation of the PF under limited computational budgets~\cite{ehrgott2016exact,herzel2021approximation}. A useful approximation should simultaneously approach the true PF and maintain sufficient coverage across different trade-offs among objectives~\cite{herzel2021approximation}.

\noindent\textbf{Multi-Objective Evolutionary Algorithms.} Multi-objective evolutionary algorithms (MOEAs) construct such approximations by evolving a population or archive of candidate solutions. Early Pareto-based evolutionary approaches established selection at the population level based on dominance and diversity, with later methods developing alternative mechanisms to balance convergence and diversity. The MOEA behavior thus depends on both problem-specific variation operators and higher-level mechanisms to select, retain, and propagate solutions across iterations.
We use two structurally distinct backbones, namely SEMO~\cite{laumanns2004running}, with incremental archive updates, and NSGA-II~\cite{deb2002fast}, which uses generational population evolution with non-dominated sorting and diversity preservation. 

\subsection{AOD Harnesses}
\label{sec:prelim_harness}

\noindent\textbf{Agent Harnesses.}
An LLM-based system includes an execution layer that governs the context presented to the model, interactions with tools and evaluators, persistent state, and control across multiple model calls. We refer to this execution layer as an \emph{agent harness}. 
In automatic operator design, the harness governs the outer process through which an LLM searches over operator candidates. It determines how prior search experience is exposed, how design context is assembled, how candidate generation and evaluation are coordinated, and how search state is maintained across iterations. We use the term \emph{AOD harness} for this task-specific execution structure. The methodology in Section~\ref{sec:method} focuses on how such a harness can maintain a persistent state and adapt its search-control behavior over time. 

\section{Methodology}
\label{sec:method}

\hemo is an AOD harness with self-evolving decision modules for multi-objective combinatorial optimization (MOCO). Its state evolves through local module updates, while a memory filesystem retains experience across states. These updates adapt experience retrieval, target and parent selection, and candidate evolution, with new optimization experience continuously recorded in memory.

\begin{figure*}[t]
    \centering
    \includegraphics[width=\textwidth]{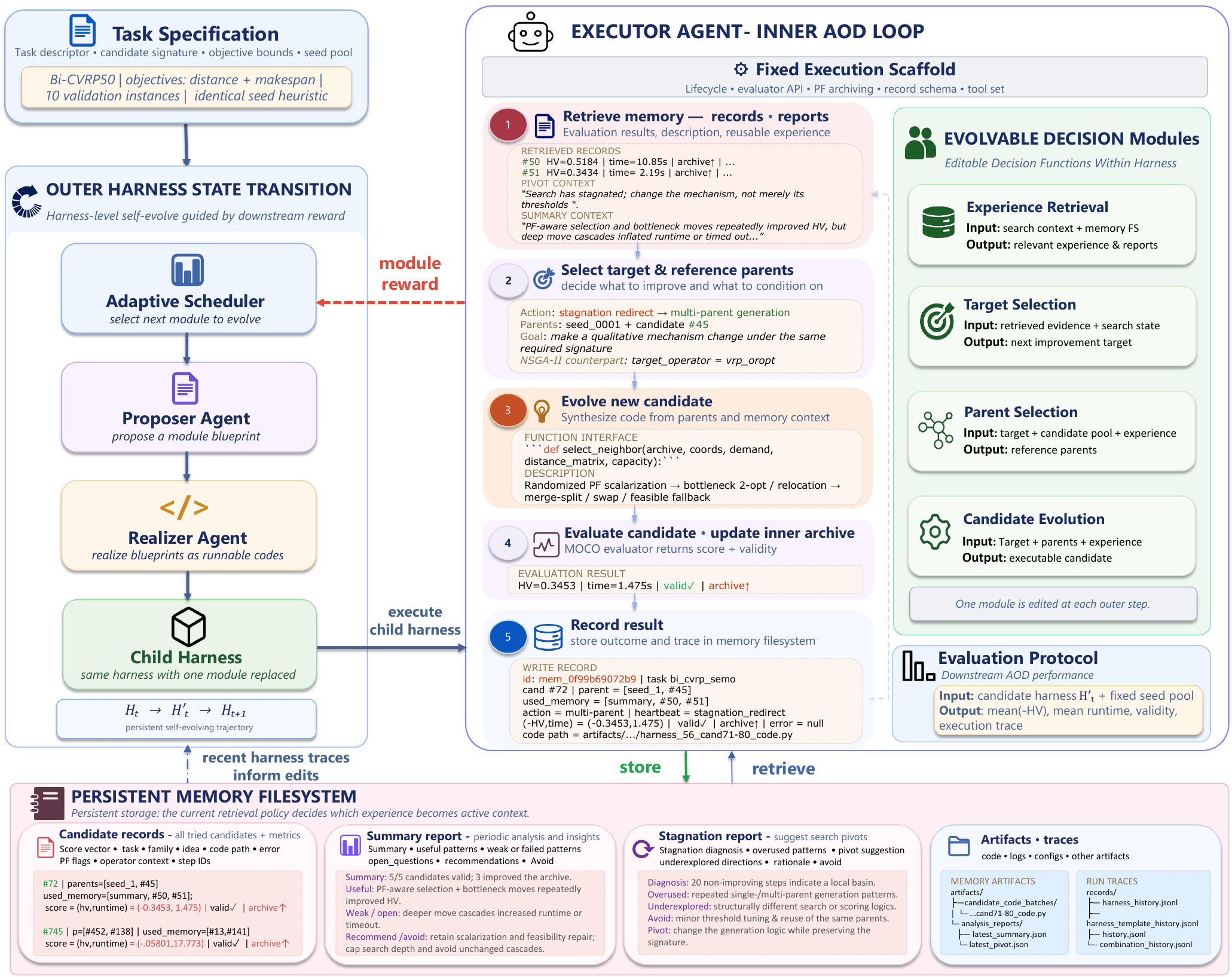}
    \caption{
    Overview of \hemo. Along the outer state transition trajectory, a selected decision module is revised and incorporated into the parent harness to form the child harness for the next evaluation episode, while optimization experience is retained in the memory filesystem.
    }
    \label{fig:hemo_overview}
\end{figure*}

Figure~\ref{fig:hemo_overview} summarizes the overall workflow. \hemo separates a fixed execution scaffold from the set of decision modules, with harness state transitions guided by the module-specific reward computed from downstream task performance. Three LLM agents play distinct roles in this process. The \textbf{Proposer Agent} specifies an updated version of one selected module in a JSON blueprint using the current harness state, task information, and recent harness traces. The \textbf{Realizer Agent} converts this blueprint into executable Python code grounded in the current harness implementation. The resulting child harness is then operationalized by the \textbf{Executor Agent} within a bounded harness evaluation episode, interacting with the memory filesystem and the task evaluator, and producing evaluated candidates and search records. Memory persists across episodes, accumulating experience along the outer trajectory.

\subsection{Stateful Harness Self-Evolution}
\label{sec:stateful_harness_evolution}

\noindent\textbf{Harness initialization.} \hemo begins from a manually specified seed harness that defines the basic execution interface and a simple reference implementation for each evolvable module. To avoid committing the search to a single manually designed starting point, a small set of alternative harness blueprints, each specifying all evolvable modules, is generated and realized to form an initial harness pool. These states are evaluated to construct an initial Pareto archive over harness states, from which the initial current state $H_0$ is selected for subsequent harness self-evolution.

\noindent\textbf{Harness state transitions.} We represent the current harness state as $H_t=(C_t,B_t)$, where $C_t$ denotes the full harness code and $B_t$ denotes its corresponding blueprint that records the current specifications of all evolvable modules. The memory filesystem at step $t$ is denoted by $M_t$ and serves as an external mechanism for retaining optimization experience across harness states. At each outer step after initialization, one evolvable module $m_t \in \mathcal{M}$ is selected for modification, where $\mathcal{M}=\{m^{(1)},\ldots,m^{(K)}\}$ denotes the set of evolvable modules. The Proposer Agent generates a module blueprint $b_t=\mathcal{P}(H_t,m_t,\mathcal{C}_t)$, where $\mathcal{C}_t$ contains the task description and recent harness evaluations. The Realizer Agent converts this blueprint into an executable replacement $\phi_t$, yielding
\begin{equation}
    H'_t
    =
    H_t[m_t \leftarrow \phi_t].
\end{equation}

After evaluating $H'_t$, it becomes the next current harness state if its parent $H_t$ does not Pareto-dominate it. The outer Pareto archive is updated with the evaluated child according to Pareto dominance. 
The resulting state
transition is
\[
(H_t,M_t)\rightarrow(H_{t+1},M_{t+1}),
\]
where an accepted harness modification updates the harness state and newly
generated experience updates the memory filesystem.
This localized transition preserves accumulated harness behavior, makes consecutive states directly comparable, and localizes credit to the modified module.

\noindent\textbf{Downstream harness evaluation.} Let $\mathcal{P}(H'_t)$ denote the final Pareto archive of operator candidates produced during the evaluation episode of $H'_t$. We use the hypervolume (HV) indicator together with runtime to summarize harness performance. The summarization score is computed by averaging these quantities over the final candidate Pareto archive:
\begin{equation}
    \mathbf{s}(H'_t)
    =
    \frac{1}{|\mathcal{P}(H'_t)|}
    \sum_{a\in\mathcal{P}(H'_t)}
    \left(
        -\mathrm{HV}(a),
        \mathrm{Runtime}(a)
    \right).
\end{equation}
Both dimensions follow a lower-is-better convention. To obtain scale-insensitive credit for the selected module, HEMO computes the normalized parent-to-child improvement along each evaluation dimension:
\begin{equation}
    r_{t,d}
    =
    \operatorname{clip}
    \left(
        \frac{
            s_d(H_t)-s_d(H'_t)
        }{
            \max\!\left(
                |s_d(H_t)|,
                |s_d(H'_t)|
            \right)
        },
        -1,
        1
    \right).
\end{equation}
The reward at module level is then aggregated as
\begin{equation}
    r_t
    =
    \frac{1}{D}
    \sum_{d=1}^{D}
    r_{t,d}.
\end{equation}
The scalar reward $r_t$ is used only for adaptive module credit assignment, while the original multi-dimensional harness evaluation is retained for Pareto-based comparison and archival throughout the outer search process.

\noindent\textbf{Adaptive Scheduler via Discounted UCB.} The Adaptive Scheduler determines which evolvable module is updated at each outer step. The utility of a module changes as the other modules in the harness are modified, making module selection inherently non-stationary~\cite{chen2012conceptual}. We therefore instantiate the scheduler with discounted UCB~\cite{garivier2011upper}, which favors recent module outcomes while retaining an exploration bonus for less frequently selected modules. For each module $m$, \hemo maintains a discounted selection count $n^{(\gamma)}_{m,t}$ and discounted reward sum $R^{(\gamma)}_{m,t}$. At each outer step, both statistics are multiplied by a discount factor $\gamma\in(0,1)$, after which the selected module receives one additional visit and its current reward. Its discounted mean reward is
$\bar r^{(\gamma)}_{m,t}=R^{(\gamma)}_{m,t}/n^{(\gamma)}_{m,t}.$
The next module is selected according to
\begin{equation}
\label{eq:discounted_ucb}
    \operatorname{UCB}_t(m)
    =
    \bar r^{(\gamma)}_{m,t}
    +
    c
    \sqrt{
        \frac{
            \log\!\left(1+\sum_{j\in\mathcal{M}} n^{(\gamma)}_{j,t}\right)
        }{
            n^{(\gamma)}_{m,t}
        }
    },
\end{equation}
where $c$ controls exploration strength. Each module is selected exactly once before applying Eq.~\eqref{eq:discounted_ucb}. Discounting is used because the reward of updating a module depends on the current harness state. A module that was unproductive earlier may become useful again after other modules have changed. Older observations are therefore gradually downweighted instead of remaining equally influential throughout the evolution trajectory. Appendix~\ref{app:discounted_credit} formally shows how discounting supports adaptive module selection under evolving harness states by attenuating stale credit and restoring exploration pressure for modules that have not been selected recently.

\subsection{Persistent Memory and Long-Horizon Feedback}
\label{sec:persistent_memory}

Long-horizon AOD can produce hundreds or thousands of evaluated candidates, making it impractical to expose the complete search history to an LLM at every step. Moreover, retaining all records in external storage does not imply that they remain effectively available to the model. Finite context windows and repeated summarization can bias the working context toward recent attempts. \hemo therefore separates \emph{persistent storage} from \emph{effective retrieval} through an external memory filesystem and explicit long-horizon feedback mechanisms.

\noindent\textbf{Structured experience memory.}
The memory filesystem stores the search experience of the evaluated operator candidates outside the transient LLM context as structured optimization events. Each record can retain candidate designs, evaluation results, parent information, validity or failure information, Pareto-related status, operator context, and other search metadata. This provides a persistent empirical history that remains reusable across successive harness states.


\noindent\textbf{Periodic long-term feedback.} \hemo periodically transforms accumulated search history into compact diagnostic feedback. These tools summarize the evolving search trajectory and identify prolonged stagnation or repeated search patterns, allowing subsequent harness modules to access abstracted signals without rescanning the complete history. We refer to each periodic invocation of these tools as a \emph{heartbeat}, consisting primarily of a summary report and a stagnation report, and the latest report of each type is maintained as the only active reference record.


\subsection{MOCO-Specific Harness Design}
\label{sec:moco_harness_design}

\hemo separates its AOD harness into a fixed execution scaffold and a set of evolvable decision roles. As shown in Fig.~\ref{fig:hemo_overview}, the fixed scaffold preserves the AOD harness lifecycle, while the evolvable roles capture decisions whose preferred behavior depends on the current search state.

\noindent\textbf{Fixed execution scaffold.} Four operational interfaces remain fixed throughout the entire stateful harness self-evolution process.
\fixedcomp{1) Evaluation interface.} It specifies solver configuration, dataset generation, evaluation instances, and feedback construction independently of the harness loop, preventing harness changes from altering the evaluation protocol or gaming the intended objective.
\fixedcomp{2) Multi-objective archive interface.} 
It merges newly generated operator candidates into the candidate Pareto archive using a fixed non-dominated sorting rule, keeping Pareto filtering and diversity preservation consistent throughout harness state transition.
\fixedcomp{3) Diagnostic and heartbeat interface.} It exposes standardized tools for summarizing the search history and diagnosing stagnation patterns. The heartbeat tools and callable interfaces remain fixed, while the decision modules determine how the resulting diagnostic feedback is retrieved and utilized.
\fixedcomp{4) Memory-writing interface.} It records evaluated candidates and search events using a common schema, ensuring a consistent empirical history across successive harness states. This separates persistent bookkeeping from the adaptive policy that determines which experience becomes active context. Examples of evolved harness states and generated operators are in Appendix~\ref{app:generated_harness_modules} and ~\ref{app:discovered_algorithms}.

\noindent\textbf{Evolvable decision roles.} Within this scaffold, \hemo exposes four high-level roles whose behavior can be modified by the harness state transition.

\module{Experience Retrieval.}
This role determines which historical candidate records and reports of the previously discovered search patterns should be surfaced from memory filesystem for the current AOD step. Its purpose is not merely to compress context, but to decide adaptively which parts of the long-horizon optimization history are currently useful.

\module{Target Selection.}
When the underlying optimization solver exposes multiple search components or evolutionary operators, this role determines which component should receive the next design intervention. Such a decision is particularly relevant to multi-component evolutionary solvers such as NSGA-II~\cite{deb2002fast}, where the marginal value of improving one operator depends on the current behavior of the surrounding operators. For solvers with only one evolvable component, this role can degenerate to a trivial choice and need not introduce additional control complexity.

\module{Parent Selection.}
This role selects previously evaluated candidates that serve as reference parents for the next generation action. These parents provide concrete working context from which the LLM can identify reusable structures, compare patterns associated with success and failures, and decide whether to refine an existing design or redirect the search toward a different region. 

\module{Candidate Evolution.}
This role determines the search strategy used to produce successive candidates within a bounded harness evaluation episode. It controls how the search progresses from one evaluated candidate to the next, coordinating candidate generation, evaluation, and subsequent iterations. The resulting strategy may follow iterative, population based, trajectory based, staged, or hybrid search behavior, allowing \hemo to adapt the dynamics of candidate evolution.

\noindent\textbf{Overall self-evolution procedure.}
Algorithm~\ref{alg:hemo} summarizes the complete outer harness
self-evolution procedure. To support controlled comparison and cleaner
credit assignment, each harness evaluation initializes its local
candidate population from the same fixed seed pool. The Executor Agent
returns the valid retained candidates, their mean score vector containing
negative HV and runtime, and an execution trace. Successfully evaluated
children update the external harness and candidate Pareto archives, while
their execution records remain in memory even when they do not replace
the current harness.

\begin{algorithm*}[!t]
\caption{Stateful Harness Self-Evolution in HEMO}
\label{alg:hemo}
\begin{algorithmic}[1]
\Require Seed harness $H_{\mathrm{seed}}$; fixed seed pool
$\mathcal{P}_{\mathrm{seed}}$; harness evaluation budget
$B_{\mathrm{outer}}$; per-harness candidate budget $B_H$

\Statex \textit{Notation:}
$\mathbf{s}(H)$ denotes the current harness score,
$\mathbf{s}'$ the proposed harness score,
$\mathcal{V}$ the evaluated candidates returned by the executor,
and $\mathrm{ok}$ evaluation success.

\State Initialize memory filesystem $\mathcal{M}$,
harness archive $\mathcal{A}_H$, and candidate archive $\mathcal{A}_A$
\State Generate initial harness states from $H_{\mathrm{seed}}$
\State Evaluate the seed and initial states using
$(\mathcal{P}_{\mathrm{seed}},\mathcal{M},B_H)$
\State Update $\mathcal{A}_H$ and $\mathcal{A}_A$
with initialization results
\State Sample current harness $H$ from $\mathcal{A}_H$
\State Initialize discounted UCB statistics for all evolvable modules

\While{harness evaluation budget remains}
    \State Select module $m$ using discounted UCB
    \State $b \gets \Call{ProposerAgent}{H,m,\text{recent traces}}$
    \Comment{revise one module specification}
    \State $H' \gets \Call{RealizerAgent}{H,m,b}$
    \Comment{instantiate the proposed revision}
    \State $(\mathcal{V},\mathbf{s}',\mathrm{ok}) \gets
    \Call{ExecutorAgent}{H',\mathcal{P}_{\mathrm{seed}},\mathcal{M},B_H}$
    \Comment{run a bounded AOD episode}

    \If{$\mathrm{ok}$}
        \State Update Pareto archive $\mathcal{A}_A$ with $\mathcal{V}$
        \State Update Pareto archive $\mathcal{A}_H$
        with $(H',\mathbf{s}')$
        \State Compute reward $r$ from the relative improvement
        of $\mathbf{s}'$ over $\mathbf{s}(H)$
        \Comment{assign credit to module $m$}
        \If{$\mathbf{s}(H)$ does not dominate $\mathbf{s}'$}
            \State $H \gets H'$
            \Comment{retain child code and blueprint}
        \EndIf
    \Else
        \State $r \gets -1$
        \Comment{retain the parent after repair retries fail}
    \EndIf

    \State Update discounted UCB statistics using module $m$ and reward $r$
\EndWhile

\State \Return $\mathcal{A}_H,\mathcal{A}_A,\mathcal{M}$
\end{algorithmic}
\end{algorithm*}

\section{Experiments}
\label{sec:experiments}


\noindent\textbf{Benchmarks and generalization protocol.} We evaluate \hemo on eight MOCO tasks spanning four representative problem families~\cite{verma2021comprehensive}, covering knapsack, vehicle routing, TSP, and scheduling. Here, TSP, KP, MDKP, CVRP, SPDP, and FJSP denote the traveling salesman
problem, knapsack problem, multi-dimensional knapsack problem,
capacitated vehicle routing problem, selective pickup and delivery
problem, and flexible job shop scheduling problem, respectively.
The prefixes Bi- and Tri- denote bi-objective and tri-objective
formulations, respectively. The source-to-target pairs are Bi-TSP $\rightarrow$ Tri-TSP, Bi-KP $\rightarrow$ Bi-MDKP, Bi-CVRP $\rightarrow$ Bi-SPDP, and Bi-FJSP $\rightarrow$ Tri-FJSP, where the source task is used for in-distribution harness self-evolution and the target task for intra-family cross-task generalization with the fixed harness state. Synthetic problems are evaluated at multiple sizes, while FJSP uses the Brandimarte benchmark~\cite{brandimarte1993routing} instances for final evaluation. Detailed task formulations are provided in Appendix~\ref{app:benchmark_problems}.

\noindent\textbf{\hemo configuration.} We instantiate \hemo with both SEMO~\cite{laumanns2004running} and NSGA-II~\cite{deb2002fast} as two structurally distinct optimization backbones. Harness self-evolution allows at most 200 harness state evaluations for both SEMO and NSGA-II. Each SEMO harness evaluation generates 50 candidates with population size 5, while each NSGA-II evaluation allows 100 complete operator combination evaluations with operator and combination population sizes of 10. 

\noindent\textbf{Objectives and evaluation metrics.} During AOD, all generated operator candidates are evaluated jointly by solution quality and computational efficiency, represented by negative Hypervolume (HV) and runtime under a common lower-is-better convention. Final performance is reported using HV, runtime, and Inverted Generational Distance (IGD), where higher HV and lower IGD indicate better Pareto-front coverage and smaller distance to the reference Pareto front, respectively. Metric definitions, reference points, and objective normalization are provided in Appendix~\ref{app:mo_metrics}.

\begin{table*}[t]
\centering
\scriptsize
\setlength{\tabcolsep}{2pt}
\renewcommand{\arraystretch}{1.15}
\caption{Average performance of the best-HV candidate selected from the final Pareto front of each of four independent runs on the three source tasks.
Time denotes runtime in seconds per instance.
The best value in each column is shown in \textbf{bold};
``--'' denotes runs that exceeded the time limit.}
\label{tab:semo_id_results}

\resizebox{\textwidth}{!}{%
\begin{tabular}{@{}lrrrrrrrrrrrrrrrrrr@{}}
\toprule
& \multicolumn{6}{c}{\textbf{Bi-TSP}}
& \multicolumn{6}{c}{\textbf{Bi-CVRP}}
& \multicolumn{6}{c}{\textbf{Bi-KP}} \\
\cmidrule(lr){2-7}
\cmidrule(lr){8-13}
\cmidrule(lr){14-19}
Method
& \multicolumn{2}{c}{$N=20$}
& \multicolumn{2}{c}{$N=50$}
& \multicolumn{2}{c}{$N=100$}
& \multicolumn{2}{c}{$N=20$}
& \multicolumn{2}{c}{$N=50$}
& \multicolumn{2}{c}{$N=100$}
& \multicolumn{2}{c}{$N=50$}
& \multicolumn{2}{c}{$N=100$}
& \multicolumn{2}{c}{$N=200$} \\
\cmidrule(lr){2-3}\cmidrule(lr){4-5}\cmidrule(lr){6-7}
\cmidrule(lr){8-9}\cmidrule(lr){10-11}\cmidrule(lr){12-13}
\cmidrule(lr){14-15}\cmidrule(lr){16-17}\cmidrule(lr){18-19}
& HV$\uparrow$ & Time$\downarrow$
& HV$\uparrow$ & Time$\downarrow$
& HV$\uparrow$ & Time$\downarrow$
& HV$\uparrow$ & Time$\downarrow$
& HV$\uparrow$ & Time$\downarrow$
& HV$\uparrow$ & Time$\downarrow$
& HV$\uparrow$ & Time$\downarrow$
& HV$\uparrow$ & Time$\downarrow$
& HV$\uparrow$ & Time$\downarrow$ \\
\midrule

SEMO
& 0.5878 & \textbf{29.411}
& 0.4612 & 35.476
& 0.3674 & 42.621
& 0.4840 & \textbf{9.726}
& 0.3380 & \textbf{8.991}
& 0.2987 & \textbf{10.691}
& 0.3543 & \textbf{14.528}
& 0.3917 & 8.146
& 0.2912 & 16.284 \\

NSGA-II
& 0.5600 & 1102.171
& 0.3860 & 1099.551
& 0.3120 & 985.188
& 0.6025 & 1178.418
& 0.5420 & 1162.623
& 0.5199 & 1530.646
& 0.3581 & 1045.565
& 0.4024 & 551.730
& 0.3207 & 529.589 \\

MOEA/D
& 0.5874 & 136.718
& 0.4385 & 146.027
& 0.3696 & 183.499
& 0.5893 & --
& 0.5288 & --
& 0.5347 & --
& 0.3541 & --
& 0.4040 & --
& 0.3285 & -- \\

PFG-MOEA
& 0.5906 & 74.886
& 0.4601 & 63.210
& 0.3847 & 86.956
& 0.5888 & 123.347
& 0.4527 & 225.457
& 0.3469 & 317.618
& 0.3562 & 68.912
& 0.4041 & 67.459
& 0.3250 & 67.129 \\

\midrule

EoH
& \textbf{0.6266} & 81.095
& 0.5961 & 35.072
& 0.6338 & 103.753
& 0.5745 & 128.232
& 0.5073 & 182.330
& 0.4979 & 169.250
& 0.3609 & 21.226
& 0.4038 & 11.891
& 0.3274 & 41.940 \\

M-EoH
& 0.6004 & 57.486
& 0.5781 & \textbf{28.429}
& 0.6155 & 88.323
& 0.5826 & 145.380
& 0.5366 & 207.620
& 0.5190 & 194.370
& 0.3576 & 14.739
& 0.3985 & 10.582
& 0.3049 & 27.431 \\

MPaGE
& 0.6206 & 78.712
& 0.6228 & 31.871
& 0.6628 & 97.819
& \textbf{0.6184} & 183.440
& 0.5532 & 243.641
& 0.5381 & 235.605
& 0.3621 & 15.783
& 0.4152 & 7.238
& 0.3350 & 19.219 \\

\rowcolor{hemoshade}
\hemo (ours)
& 0.6258 & 43.557
& \textbf{0.6336} & 36.829
& \textbf{0.6818} & \textbf{42.134}
& 0.6130 & 134.539
& \textbf{0.5777} & 191.809
& \textbf{0.5700} & 163.293
& \textbf{0.3653} & 16.053
& \textbf{0.4329} & \textbf{6.875}
& \textbf{0.3619} & \textbf{12.620} \\

\bottomrule
\end{tabular}%
}
\end{table*}

\noindent\textbf{Compared baselines and evaluation protocol.} We compare \hemo with three groups of baselines:
(1) \textbf{Conventional MOEAs}, including SEMO, NSGA-II, MOEA/D~\cite{zhang2007moea}, and PFG-MOEA~\cite{xu2023pareto}, 
(2) \textbf{Neural MOCO methods}, represented by PMOCO~\cite{lin2022pareto}
and
(3) \textbf{LLM-based AOD methods}, including EoH~\cite{liu2024evolution}, MEoH~\cite{yao2025multi}, MPaGE~\cite{hieu2026pareto}, and E2OC~\cite{qiu2026e2oc}. The \hemo Proposer, Realizer, and Executor agents use Claude Opus 4.8, GPT-5, and GPT-4.1-nano, respectively. During inference, all LLM-based AOD methods use GPT-5-nano unless otherwise stated, with identical candidate generation budgets of 2,000 for SEMO and 4,000 for NSGA-II and the same per-candidate evaluation timeout. 
Unless otherwise stated, LLM-based AOD results are averaged over four independent trials. For inference, initialization states are excluded, and the evolved harness state with the smallest normalized Euclidean distance to the empirical ideal point is selected from the remaining harness states. The selected harness is then applied to both the source task and its corresponding transfer task. For all MOEA solvers, a uniform timeout of 3,600 seconds is imposed on each test run across all test instances. Additional configuration and implementation details are provided in Appendix~\ref{app:experimental_details}.

\subsection{Main Results}
\label{sec:main_results}

\begin{table*}[t]
\centering
\scriptsize
\setlength{\tabcolsep}{2pt}
\renewcommand{\arraystretch}{1.15}
\caption{Performance averaged over the best-HV candidate of each of four independent runs on the three intra-family transfer tasks.
The best available value in each column is shown in \textbf{bold}.}
\label{tab:semo_ood_results}

\resizebox{\textwidth}{!}{%
\begin{tabular}{@{}lrrrrrrrrrrrrrrrrrr@{}}
\toprule
& \multicolumn{6}{c}{\textbf{Tri-TSP}}
& \multicolumn{6}{c}{\textbf{Bi-SPDP}}
& \multicolumn{6}{c}{\textbf{Bi-MDKP}} \\
\cmidrule(lr){2-7}
\cmidrule(lr){8-13}
\cmidrule(lr){14-19}
Method
& \multicolumn{2}{c}{$N=20$}
& \multicolumn{2}{c}{$N=50$}
& \multicolumn{2}{c}{$N=100$}
& \multicolumn{2}{c}{$N=20$}
& \multicolumn{2}{c}{$N=50$}
& \multicolumn{2}{c}{$N=100$}
& \multicolumn{2}{c}{$N=50$}
& \multicolumn{2}{c}{$N=100$}
& \multicolumn{2}{c}{$N=200$} \\
\cmidrule(lr){2-3}\cmidrule(lr){4-5}\cmidrule(lr){6-7}
\cmidrule(lr){8-9}\cmidrule(lr){10-11}\cmidrule(lr){12-13}
\cmidrule(lr){14-15}\cmidrule(lr){16-17}\cmidrule(lr){18-19}
& HV$\uparrow$ & Time$\downarrow$
& HV$\uparrow$ & Time$\downarrow$
& HV$\uparrow$ & Time$\downarrow$
& HV$\uparrow$ & Time$\downarrow$
& HV$\uparrow$ & Time$\downarrow$
& HV$\uparrow$ & Time$\downarrow$
& HV$\uparrow$ & Time$\downarrow$
& HV$\uparrow$ & Time$\downarrow$
& HV$\uparrow$ & Time$\downarrow$ \\
\midrule

SEMO
& 0.4055 & \textbf{11.166}
& 0.1880 & \textbf{22.008}
& 0.1041 & \textbf{43.133}
& 0.7510 & 1.973
& 0.8343 & \textbf{3.645}
& 0.8461 & \textbf{7.005}
& 0.2536 & \textbf{2.875}
& 0.2460 & \textbf{2.994}
& 0.4866 & \textbf{3.144} \\

NSGA-II
& 0.4020 & 814.270
& 0.2050 & 1004.854
& 0.1310 & 1207.876
& 0.7531 & 568.591
& 0.8081 & 537.841
& 0.8137 & 563.270
& 0.2596 & 1050.822
& 0.2662 & 553.071
& 0.5469 & 558.297 \\

MOEA/D
& 0.4098 & 489.636
& 0.2230 & 1104.641
& 0.1534 & 2577.410
& 0.7579 & 61.224
& 0.8219 & 107.852
& 0.8317 & 140.504
& 0.2589 & --
& 0.2729 & --
& 0.5774 & -- \\

PFG-MOEA
& 0.4520 & 265.914
& 0.2810 & 765.435
& 0.2125 & 963.652
& 0.7465 & 13.927
& 0.8029 & 31.840
& 0.8117 & 58.505
& 0.2535 & 76.271
& 0.2587 & 74.336
& 0.5420 & 70.115 \\

\midrule

EoH
& 0.4230 & 57.480
& 0.3872 & 134.550
& 0.4463 & 297.430
& 0.7608 & 10.133
& \textbf{0.8655} & 85.564
& 0.8825 & 128.023
& \textbf{0.2700} & 175.451
& 0.2915 & 284.804
& 0.6147 & 423.494 \\

M-EoH
& 0.4520 & 103.870
& 0.4235 & 187.550
& 0.4612 & 418.950
& 0.7567 & \textbf{1.874}
& 0.8523 & 8.090
& 0.8814 & 12.533
& 0.2611 & 124.392
& 0.2834 & 232.238
& 0.5835 & 312.023 \\

MPaGE
& \textbf{0.4709} & 87.400
& 0.4165 & 159.299
& 0.4702 & 359.478
& 0.7589 & 2.213
& 0.8501 & 13.221
& 0.8733 & 16.985
& 0.2653 & 93.242
& \textbf{0.3020} & 214.967
& 0.6260 & 291.463 \\

\rowcolor{hemoshade}
\hemo (ours)
& 0.4693 & 94.891
& \textbf{0.4447} & 179.706
& \textbf{0.4967} & 386.932
& \textbf{0.7617} & 2.689
& 0.8648 & 8.884
& \textbf{0.9026} & 13.429
& 0.2679 & 91.784
& 0.2980 & 185.851
& \textbf{0.6336} & 213.299 \\

\bottomrule
\end{tabular}%
}
\end{table*}

\noindent\textbf{Source-task performance with SEMO.} Table~\ref{tab:semo_id_results} reports the detailed results on the source tasks used during harness self-evolution. \hemo achieves the highest HV in seven of the nine task-size settings across Bi-TSP, Bi-CVRP, and Bi-KP. The advantage becomes clearer on larger testing instances. For example, on Bi-CVRP with $N=100$, \hemo reaches an HV of $0.5700$, compared with $0.5381$ for the strongest baseline, while on Bi-KP with $N=200$ it achieves $0.3619$ compared with $0.3350$. On Bi-TSP, the advantage of \hemo becomes more pronounced as the problem size increases. These results show that harness adaptation can further improve the quality of generated heuristics over existing LLM-based AOD frameworks. Moreover, the HV gains are not accompanied by consistently longer execution. \hemo remains competitive in runtime and is the fastest method in several larger settings, including Bi-TSP with $N=100$ and Bi-KP with $N=100$ and $N=200$, suggesting that the improvement in solution quality does not systematically come at the cost of higher heuristic runtime.

\noindent\textbf{Intra-family cross-task generalization.} Table~\ref{tab:semo_ood_results} evaluates the transferred harness in related target tasks. \hemo obtains the best HV in five of nine settings and places second in the remaining four. Although the target tasks are not seen during harness self-evolution, the transferred harness remains competitive across all three tasks. \hemo also achieves the fastest runtime among LLM-based AOD methods in several transferred settings while consistently ranking first in HV at the largest problem size, indicating that the learned harness behavior can transfer effectively across related tasks. Additional FJSP results with the SEMO backbone are provided in Appendix~\ref{app:additional_semo_fjsp}, while comparisons with expert-designed TSP operators and their
sequential combinations are provided in
Appendix~\ref{app:expert_tsp_operators}

\noindent\textbf{Performance with the NSGA-II.} \begin{table}[t]

\centering
\scriptsize
\setlength{\tabcolsep}{2pt}
\renewcommand{\arraystretch}{1}
\setlength{\abovecaptionskip}{3pt}
\setlength{\belowcaptionskip}{3pt}

\caption{NSGA-II results on TSP (averaged over 50 instances per size) and FJSP (ID: Mk15; OOD: Mk01--Mk14).}
\label{tab:nsga2_main_results}

\resizebox{\linewidth}{!}{%
\begin{tabular}{@{}l rr rr
    >{\columncolor{hemoshade}}r
    >{\columncolor{hemoshade}}r@{}}
\toprule
& \multicolumn{2}{c}{Expert}
& \multicolumn{2}{c}{E2OC}
& \multicolumn{2}{>{\columncolor{hemoshade}}c}{\hemo} \\
\cmidrule(lr){2-3}\cmidrule(lr){4-5}\cmidrule(lr){6-7}
Setting
& HV$\uparrow$ & IGD$\downarrow$
& HV$\uparrow$ & IGD$\downarrow$
& HV$\uparrow$ & IGD$\downarrow$ \\
\midrule
\multicolumn{7}{c}{\textbf{Traveling Salesman Problems}} \\
\midrule
Bi-TSP 20
& 0.5476 & 0.0579
& \textbf{0.6174} & \textbf{0.0081}
& 0.6103 & 0.0093 \\
Bi-TSP 50
& 0.3613 & 0.1330
& 0.4897 & 0.0437
& \textbf{0.4948} & \textbf{0.0215} \\
Bi-TSP 100
& 0.2835 & 0.1021
& 0.3631 & 0.0367
& \textbf{0.3763} & \textbf{0.0214} \\
\cmidrule(lr){1-7}
Tri-TSP 20
& 0.3661 & 0.0660
& 0.4132 & 0.0206
& \textbf{0.4305} & \textbf{0.0160} \\
Tri-TSP 50
& 0.1579 & 0.1514
& 0.2157 & 0.0559
& \textbf{0.2422} & \textbf{0.0300} \\
Tri-TSP 100
& 0.0915 & 0.1034
& 0.1187 & 0.0393
& \textbf{0.1262} & \textbf{0.0293} \\
\midrule
\multicolumn{7}{c}{\textbf{Flexible Job-Shop Scheduling Problems}} \\
\midrule
Bi-FJSP ID
& 0.6836 & 0.0625
& 0.7610 & \textbf{0.0086}
& \textbf{0.7660} & 0.0090 \\
Bi-FJSP OOD
& 0.5716 & 0.0580
& 0.6123 & 0.0167
& \textbf{0.6233} & \textbf{0.0126} \\
Bi-FJSP All
& 0.5790 & 0.0583
& 0.6222 & 0.0162
& \textbf{0.6328} & \textbf{0.0124} \\
\cmidrule(lr){1-7}
Tri-FJSP ID
& 0.1783 & 0.1125
& \textbf{0.2863} & 0.0059
& 0.2859 & \textbf{0.0051} \\
Tri-FJSP OOD
& 0.1712 & 0.1342
& 0.2748 & 0.0134
& \textbf{0.2836} & \textbf{0.0090} \\
Tri-FJSP All
& 0.1717 & 0.1328
& 0.2756 & 0.0129
& \textbf{0.2838} & \textbf{0.0087} \\
\bottomrule
\end{tabular}%
}

\end{table}
\noindent\textbf{TSP.} Table~\ref{tab:nsga2_main_results} directly compares \hemo with expert-designed operators and E2OC on Bi-TSP and Tri-TSP. 
\hemo achieves the best HV and IGD in five of the six settings, with E2OC only slightly ahead on Bi-TSP with $N=20$. On Bi-TSP with $N=50$ and $N=100$, \hemo improves both reported metrics over E2OC. Notably, the fixed transferred harness consistently achieves the best HV and IGD at all three Tri-TSP sizes, indicating that the evolved harness improves multi-operator design under NSGA-II and retains its effectiveness when transferred from bi-objective to tri-objective TSP.

\noindent\textbf{FJSP.}
Table~\ref{tab:nsga2_main_results} further evaluates operator design on Bi-FJSP and Tri-FJSP. On Bi-FJSP, \hemo achieves the highest HV for both ID and OOD instances and the best overall HV and IGD. On Tri-FJSP, E2OC has a marginally higher HV on Mk15, but \hemo obtains better HV and IGD on the OOD instances and consequently the best overall solution quality. This pattern shows that the evolved harness generalizes well across objective dimensionalities and different machine and job configurations. Appendix~\ref{app:additional_nsga2_results} reports NSGA-II results on knapsack and routing tasks.


\subsection{Further Analysis}
\label{sec:discussion}

\noindent\textbf{Comparison with a neural baseline.} We compare the SEMO heuristics evolved by \hemo (HEMO-SEMO)
with PMOCO~\cite{lin2022pareto}, a preference-conditioned neural solver,
across 16 (task, size) combinations. Both methods are evaluated on the same 50 test instances per setting, with consistent objective normalization
and hypervolume reference points. PMOCO uses the corresponding
pretrained checkpoint when available. For the larger TSP settings with $N=150$ and $N=200$, we reuse its $N=20$
checkpoint without retraining. For \hemo-SEMO, the compared
heuristics are designed only on 10 IID instances at a single design
size for each problem family, as summarized
in Table~\ref{tab:aad_design_instances} in the appendix. The heuristic with the
highest validation HV across all
runs is selected for reporting the test performance.


Table~\ref{tab:semo_neural_comparison} shows that \hemo-SEMO
achieves higher HV in \textbf{11 of 16 settings} and lower
empirical IGD in \textbf{12 of 16 settings}. It outperforms
PMOCO on both metrics across all Tri-TSP and Bi-KP settings,
and also performs strongly on Bi-TSP, including both larger
problem sizes. Overall, \hemo with SEMO achieves
wins in more settings under both metrics, although the relative
advantage remains problem dependent.

The larger-size TSP results further examine cross-size
generalization beyond the task sizes used by the default pretrained
configurations, where generalization across problem sizes remains
an important challenge for neural MOCO methods~\cite{lin2022pareto}.
On Bi-TSP, \hemo-SEMO achieves higher HV and lower empirical IGD
than PMOCO at both $N=150$ and $N=200$. The same pattern is observed
on Tri-TSP, where \hemo-SEMO also achieves higher HV and lower
empirical IGD at both larger sizes. Neural methods often require retraining on larger instances to improve
generalization across problem sizes. In contrast, \hemo{} with SEMO
outperforms PMOCO on both TSP tasks at $N=150$ and $N=200$ without
additional training, with both methods using the same training size.

\definecolor{neuralDiffShade}{RGB}{226,239,250}

\begin{table}[!t]
\centering
\small
\setlength{\tabcolsep}{2pt}
\renewcommand{\arraystretch}{0.96}
\caption{Solution-quality comparison with PMOCO without
instance augmentation.
Results are averaged over 50 instances per setting.
The better value for each metric within each setting is
shown in \textbf{bold}. Winning \hemo-SEMO values are
additionally shaded in blue.}
\label{tab:semo_neural_comparison}

\begin{tabular}{@{}l rr rr@{}}
\toprule
& \multicolumn{2}{c}{HV$\uparrow$}
& \multicolumn{2}{c}{IGD$\downarrow$} \\
\cmidrule(lr){2-3}\cmidrule(lr){4-5}
Setting
& PMOCO & \hemo-SEMO
& PMOCO & \hemo-SEMO \\
\midrule

Bi-TSP 20
& 0.6260
& \cellcolor{hemoshade}\textbf{0.6269}
& 0.005281
& \cellcolor{hemoshade}\textbf{0.000781} \\

Bi-TSP 50
& \textbf{0.6350} & 0.6348
& 0.006773
& \cellcolor{hemoshade}\textbf{0.001295} \\

Bi-TSP 100
& \textbf{0.6977} & 0.6853
& \textbf{0.004168} & 0.006003 \\

Bi-TSP 150
& 0.6958
& \cellcolor{hemoshade}\textbf{0.7142}
& 0.020435
& \cellcolor{hemoshade}\textbf{0.003545} \\

Bi-TSP 200
& 0.7151
& \cellcolor{hemoshade}\textbf{0.7276}
& 0.019029
& \cellcolor{hemoshade}\textbf{0.006109} \\

\midrule

Tri-TSP 20
& 0.4660
& \cellcolor{hemoshade}\textbf{0.4713}
& 0.024427
& \cellcolor{hemoshade}\textbf{0.000652} \\

Tri-TSP 50
& 0.4341
& \cellcolor{hemoshade}\textbf{0.4495}
& 0.035622
& \cellcolor{hemoshade}\textbf{0.000678} \\

Tri-TSP 100
& 0.4889
& \cellcolor{hemoshade}\textbf{0.5007}
& 0.039789
& \cellcolor{hemoshade}\textbf{0.000699} \\

Tri-TSP 150
& 0.5023
& \cellcolor{hemoshade}\textbf{0.5325}
& 0.045129
& \cellcolor{hemoshade}\textbf{0.000636} \\

Tri-TSP 200
& 0.5262
& \cellcolor{hemoshade}\textbf{0.5327}
& 0.043203
& \cellcolor{hemoshade}\textbf{0.001889} \\

\midrule

Bi-CVRP 20
& \textbf{0.6834} & 0.6281
& \textbf{0.001973} & 0.080269 \\

Bi-CVRP 50
& \textbf{0.6414} & 0.5856
& \textbf{0.002224} & 0.103187 \\

Bi-CVRP 100
& \textbf{0.6718} & 0.6000
& \textbf{0.001141} & 0.099605 \\

\midrule

Bi-KP 50
& 0.3641
& \cellcolor{hemoshade}\textbf{0.3663}
& 0.008123
& \cellcolor{hemoshade}\textbf{0.000080} \\

Bi-KP 100
& 0.4331
& \cellcolor{hemoshade}\textbf{0.4359}
& 0.008765
& \cellcolor{hemoshade}\textbf{0.000235} \\

Bi-KP 200
& 0.3622
& \cellcolor{hemoshade}\textbf{0.3645}
& 0.008241
& \cellcolor{hemoshade}\textbf{0.000249} \\

\midrule
Wins / 16
& 5 & \textbf{11}
& 4 & \textbf{12} \\
\bottomrule
\end{tabular}
\end{table}

\begin{figure*}[t]
    \centering

    \includegraphics[width=\textwidth]
        {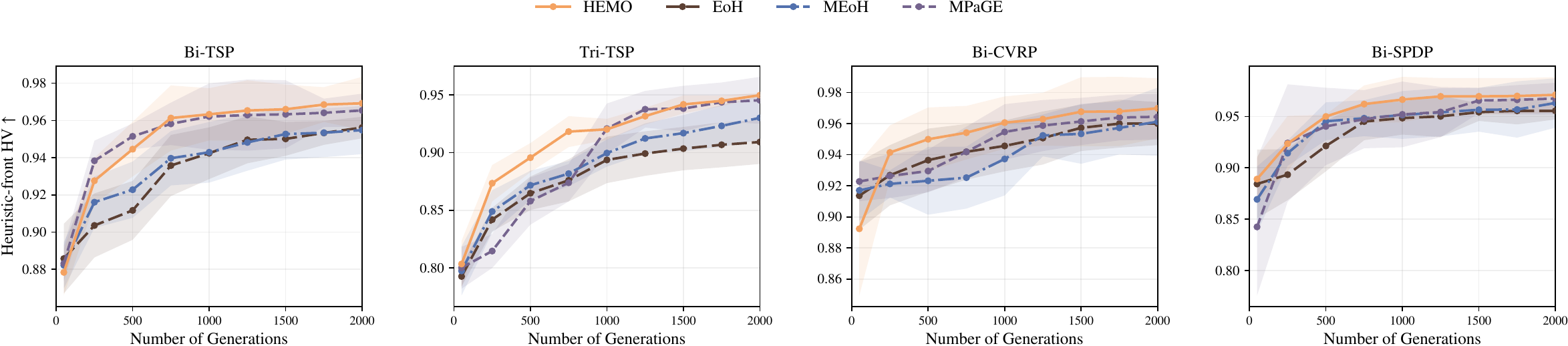}
    \par\vspace{-3pt}
    {\scriptsize (a) Evolution of heuristic-front HV.}

    \par\vspace{4pt}

    \includegraphics[width=\textwidth]
        {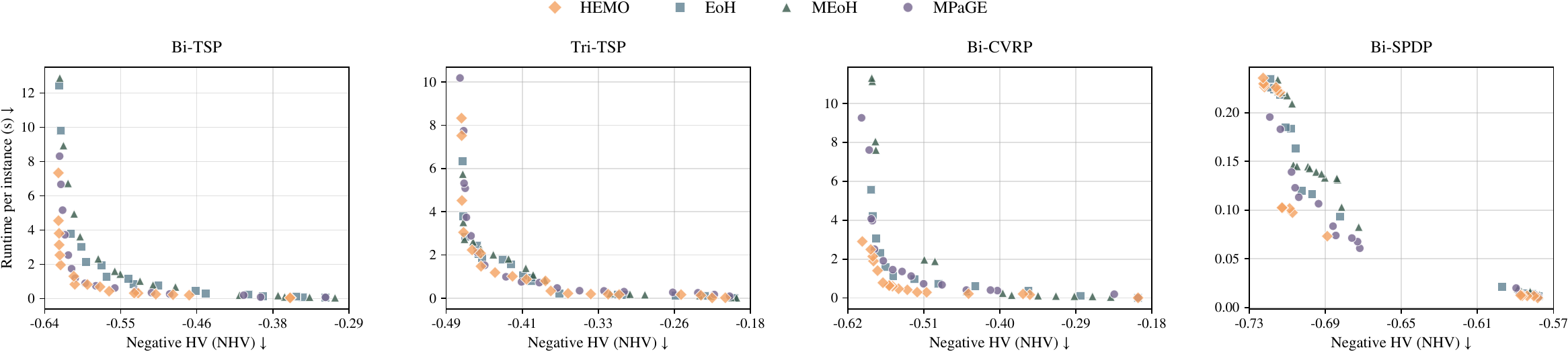}
    \par\vspace{-3pt}
    {\scriptsize (b) Final quality--runtime trade-offs.}

    \caption{
    Search dynamics and final heuristic fronts across four tasks.
    (a) Mean heuristic-front HV with $\pm$ std shading.
    (b) Aggregated final heuristic Pareto fronts obtained by nondominated filtering.
    }
    \label{fig:heuristic_search_analysis}
\end{figure*}

\noindent\textbf{Quality--runtime trade-offs.} Figure~\ref{fig:heuristic_search_analysis}(a) shows how the heuristic-front HV evolves during inference. On the ID tasks, Bi-TSP and Bi-CVRP, \hemo achieves the strongest final trade-offs despite not always leading early, indicating the benefit of continued refinement beyond the initial candidate population. 
On the OOD tasks, Tri-TSP and Bi-SPDP, the trajectories are closer, but \hemo remains competitive late in search and attains the highest final mean heuristic-front HV, suggesting that effective search guidance persists after intra-family transfer.
Figure~\ref{fig:heuristic_search_analysis}(b) shows a complementary view of the final quality--runtime trade-offs. On the ID tasks, \hemo covers favorable regions across quality levels, often matching solution quality at lower runtime and offering alternatives for different computational budgets rather than a single high-quality candidate. On the OOD tasks, the fronts overlap more and no method dominates every region, yet \hemo remains on or near the favorable boundary across quality and runtime levels. 
Appendix~\ref{app:solution_space_comparison} provides additional visualizations of the obtained solution sets.

\noindent\textbf{Ablations.} \begin{table}[t]
\centering

\caption{Ablation study of \hemo on Bi-TSP20 and Tri-TSP20,
averaged over four independent runs. HV and IGD for each variant are computed from the aggregated non-dominated solution fronts of all heuristics in the final heuristic Pareto front.}
\label{tab:hemo_ablation}

{\small
\setlength{\tabcolsep}{2pt}
\renewcommand{\arraystretch}{0.95}

\begin{tabular*}{\linewidth}{@{\extracolsep{\fill}}lcccc@{}}
\toprule
&
\multicolumn{2}{c}{Bi-TSP20}
&
\multicolumn{2}{c}{Tri-TSP20} \\
\cmidrule(lr){2-3}\cmidrule(lr){4-5}

\textbf{Variant}
& HV$\uparrow$ & IGD$\downarrow$
& HV$\uparrow$ & IGD$\downarrow$ \\
\midrule

\textbf{\hemo}
& \textbf{0.9583} & \textbf{0.0046}
& \textbf{0.9330} & \textbf{0.0084} \\

w/o harness evolution
& 0.6133 & 0.0364 & 0.5754 & 0.0462 \\

w/o discounted UCB
& 0.8766 & 0.0097 & 0.8318 & 0.0153 \\

w/o global memory
& 0.7879 & 0.0122 & 0.7942 & 0.0168 \\

w/o retrieved context
& 0.6842 & 0.0176 & 0.6366 & 0.0286 \\
\bottomrule
\end{tabular*}
}
\end{table}
We ablate the main components of \hemo on two TSP tasks under SEMO. Table~\ref{tab:hemo_ablation} shows that the unevolved seed harness performs the worst, lowering HV by $0.3450$ on Bi-TSP and $0.3576$ on Tri-TSP. Replacing the discounted UCB with sequential module selection consistently worsens HV and IGD, supporting adaptive credit assignment among the evolving decision modules. Restricting memory access to records produced by the current harness state further reduces
performance, while removing the retrieved memory context from subsequent search decisions causes an even larger drop. These results indicate that
the gains of \hemo arise jointly from stateful harness self-evolution, adaptive module scheduling, and memory filesystem, and that these benefits remain visible after cross-task transfer. 





\noindent\textbf{Cost and performance across LLM backbones.}
We further evaluate \hemo with different LLM backbones to examine the
trade-off between search performance and computational cost. Table~\ref{tab:llm_cost}
reports token consumption, API cost, wall-clock runtime, HV, and empirical
IGD on Bi-CVRP and its transferred Bi-SPDP task.

The comparison highlights a clear cost--performance trade-off across
different LLM backbones. GPT-4.1-nano is among the fastest and least expensive models, but produces weaker final heuristic performance than
the stronger backbones. This motivates its use for high-volume candidate generation during harness self-evolution, where a large number of evaluations make inference speed and cost particularly important.

For inference with a fixed selected harness state, GPT-5-nano provides a stronger
overall balance between search quality and computational cost. It achieves the
best HV and IGD on Bi-CVRP while remaining substantially less expensive to use
than Claude Haiku 4.5 and the Qwen backbones. On Bi-SPDP, several larger
models obtain slightly better final solution quality, but at considerably
higher API cost or substantially longer wall-clock runtime. We therefore use GPT-5-nano as the
common inference backbone for \hemo and all LLM-based AOD baselines in the
main experiments, providing a strong and relatively cost-efficient model
under a unified evaluation setting.

\begin{table}[!t]
\centering
\caption{
Cost and performance of \hemo under different LLM backbones on Bi-CVRP
and Bi-SPDP with $N=20$. Performance is evaluated separately on 50 test
instances for each task.
}
\label{tab:llm_cost}

{\small
\setlength{\tabcolsep}{1.2pt}
\renewcommand{\arraystretch}{1.1}

\begin{tabular*}{\columnwidth}
{@{\extracolsep{\fill}}l|ccc|cc@{}}
\noalign{\hrule height 1.2pt}
\multicolumn{6}{c}{\textbf{Bi-CVRP}} \\
\noalign{\hrule height 0.6pt}

Model
& \shortstack{Token\\(M)$\downarrow$}
& \shortstack{Cost\\(\$)$\downarrow$}
& \shortstack{Wall-clock\\(h)$\downarrow$}
& HV$\uparrow$
& IGD$\downarrow$ \\
\noalign{\hrule height 0.4pt}

\textbf{GPT-5-nano}
& 49.00
& 7.46
& 44.54
& \textbf{0.6173}
& \textbf{0.0082} \\

GPT-4.1-nano
& 29.80
& \textbf{3.82}
& 13.95
& 0.5832
& 0.0355 \\

Claude Haiku 4.5
& 45.51
& 75.15
& \textbf{12.63}
& 0.6015
& 0.0159 \\

Qwen3.5-27b
& 56.30
& 32.74
& 106.24
& 0.5960
& 0.0306 \\

Qwen3.5-122b-a10b
& 38.82
& 31.29
& 80.44
& 0.5914
& 0.0414 \\

DeepSeek-V3.2
& \textbf{28.31}
& 6.18
& 28.05
& 0.6053
& 0.0162 \\

\noalign{\hrule height 0.6pt}
\multicolumn{6}{c}{\textbf{Bi-SPDP}} \\
\noalign{\hrule height 0.6pt}

Model
& \shortstack{Token\\(M)$\downarrow$}
& \shortstack{Cost\\(\$)$\downarrow$}
& \shortstack{Wall-clock\\(h)$\downarrow$}
& HV$\uparrow$
& IGD$\downarrow$ \\
\noalign{\hrule height 0.4pt}

\textbf{GPT-5-nano}
& 29.06
& 3.83
& 18.62
& 0.7617
& 0.0167 \\

GPT-4.1-nano
& \textbf{11.05}
& \textbf{1.42}
& \textbf{7.62}
& 0.7540
& 0.0272 \\

Claude Haiku 4.5
& 52.65
& 86.94
& 14.99
& 0.7618
& 0.0165 \\

Qwen3.5-27b
& 28.41
& 16.52
& 64.95
& \textbf{0.7674}
& \textbf{0.0103} \\

Qwen3.5-122b-a10b
& 32.27
& 26.01
& 52.22
& 0.7625
& 0.0159 \\

DeepSeek-V3.2
& 20.26
& 4.42
& 15.75
& 0.7642
& 0.0141 \\

\noalign{\hrule height 1.2pt}
\end{tabular*}
}
\end{table}

\subsection{Limitations and Future Directions}

\noindent\textbf{Model accessibility.}
The Proposer and Realizer currently use proprietary language models,
making harness self-evolution dependent on external API access and cost.
Adapting smaller or open weight models to these roles, potentially
through distillation of successful harness design trajectories,
could make the framework more accessible.

\noindent\textbf{Evaluation cost and cross-task memory transfer.}
Candidate evaluation remains computationally demanding,
as assessing a designed candidate requires executing it on multiple
problem instances.
While \hemo reuses accumulated memory within a task,
our cross-task experiments transfer the harness rather than
the task-specific memory accumulated during search.
A promising extension is a transferable memory filesystem
that distinguishes reusable design knowledge from observations and evaluation records specific to each task.
Such a mechanism could reduce redundant exploration and
evaluation on new tasks, while requiring safeguards against
negative transfer from incompatible prior experience.

\noindent\textbf{Generalization across tasks.}
Although \hemo-SEMO outperforms PMOCO in several settings,
including the larger size Bi-TSP and Tri-TSP experiments,
its advantage is not uniform across problem families.
PMOCO remains stronger on Bi-CVRP, indicating that the
relative effectiveness of the two approaches depends on
the target problem.
An important direction is therefore to investigate how design datasets for operators and evaluators can better support
generalization across instance sizes and distributions.
In particular, more diverse design instances and evaluation
criteria that explicitly assess robustness to distribution
shifts may help produce more broadly transferable operators.

\section{Conclusion}
\label{sec:conclusion}
We present \hemo, a stateful harness self-evolution framework for LLM-based automatic operator design in multi-objective combinatorial optimization. By adapting the harness itself through localized state transitions while preserving long-term optimization experience in a memory filesystem, \hemo enables search control to co-evolve with the operator design process. Comparing against state-of-the-art LLM-based AOD methods, \hemo achieves strong performance across diverse tasks and two different evolutionary backbones, while also exhibiting effective intra-family cross-task transfer. We believe \hemo opens a promising direction in which the harness itself becomes a persistent and adaptive optimization object, enabling more effective long-horizon and transferable operator design.

\nocite{
mavrotas2009bomdkp,
liao2013solving,
lust2010multiobjective,
zhang2009effective,
li2010effective,
laumanns2002combining,
knowles2003properties,
chen2025reinforced,
li2024evolutionary
}
\bibliographystyle{IEEEtran}
\bibliography{references}

@article{chen2012conceptual,
  title={A conceptual modeling of meme complexes in stochastic search},
  author={Chen, Xianshun and Ong, Yew Soon},
  journal={IEEE Transactions on Systems, Man, and Cybernetics, Part C (Applications and Reviews)},
  volume={42},
  number={5},
  pages={612--625},
  year={2012},
  publisher={IEEE}
}

@article{dong2026evolutionary,
  title={Evolutionary Computation for Trustworthy AI: From Attacks and Defenses to Self-Evolving Era},
  author={Dong, Junhao and Wang, Chenkai and Lin, Xuanhui and Gong, Mingrong and Wang, Siyu and Wen, Yuqing and Liu, Jiao and Huang, Catherine and Yen, Gary G and Yao, Xin and others},
  journal={arXiv preprint arXiv:2610.02996},
  year={2026}
}

@article{zeng2016structured,
  title={Structured memetic automation for online human-like social behavior learning},
  author={Zeng, Yifeng and Chen, Xuefeng and Ong, Yew-Soon and Tang, Jing and Xiang, Yanping},
  journal={IEEE Transactions on Evolutionary Computation},
  volume={21},
  number={1},
  pages={102--115},
  year={2016},
  publisher={IEEE}
}

@article{wang2026learning,
  title={Learning Evolution via Optimization Knowledge Adaptation},
  author={Wang, Chao and Li, Lingling and Jiao, Licheng and Zhao, Jiaxuan
          and Liu, Fang and Yang, Shuyuan},
  journal={IEEE Transactions on Pattern Analysis and Machine Intelligence},
  volume={48},
  number={9},
  pages={10442--10459},
  year={2026},
  doi={10.1109/TPAMI.2026.3686919},
  publisher={IEEE}
}

@article{wang2024asp,
  title={ASP: Learn a universal neural solver!},
  author={Wang, Chenguang and Yu, Zhouliang and McAleer, Stephen and Yu, Tianshu and Yang, Yaodong},
  journal={IEEE Transactions on Pattern Analysis and Machine Intelligence},
  volume={46},
  number={6},
  pages={4102--4114},
  year={2024},
  publisher={IEEE}
}

@inproceedings{zhu2026simpleevol,
  title = {{SimpleEvol}: An Agent-Loop Framework for {LLM}-Driven Automated Heuristic Design with Minimal Human Priors},
  author = {Zhu, Jianghan and Zhang, Cong and Zhu, Rongjie and Zhang, Chi and Cao, Zhiguang},
  booktitle = {Advances in Neural Information Processing Systems},
  year = {2026}
}

@book{pillay2021automated,
  title={Automated Design of Machine Learning and Search Algorithms},
  author={Pillay, Nelishia and Qu, Rong},
  year={2021},
  publisher={Springer}
}

@article{burke2013hyper,
  title={Hyper-heuristics: A survey of the state of the art},
  author={Burke, Edmund K and Gendreau, Michel and Hyde, Matthew and Kendall, Graham and Ochoa, Gabriela and {\"O}zcan, Ender and Qu, Rong},
  journal={Journal of the Operational Research Society},
  volume={64},
  number={12},
  pages={1695--1724},
  year={2013},
  publisher={Taylor \& Francis}
}

@book{langdon2002foundations,
  title={Foundations of genetic programming},
  author={Langdon, William B and Poli, Riccardo},
  volume={90},
  year={2002},
  publisher={Springer}
}

@article{romera2024mathematical,
  title={Mathematical discoveries from program search with large language models},
  author={Romera-Paredes, Bernardino and Barekatain, Mohammadamin and Novikov, Alexander and Balog, Matej and Kumar, M Pawan and Dupont, Emilien and Ruiz, Francisco JR and Ellenberg, Jordan S and Wang, Pengming and Fawzi, Omar and others},
  journal={Nature},
  volume={625},
  number={7995},
  pages={468--475},
  year={2024},
  publisher={Nature Publishing Group UK London}
}

@article{liu2024evolution,
  title={Evolution of heuristics: Towards efficient automatic algorithm design using large language model},
  author={Liu, Fei and Tong, Xialiang and Yuan, Mingxuan and Lin, Xi and Luo, Fu and Wang, Zhenkun and Lu, Zhichao and Zhang, Qingfu},
  journal={arXiv preprint arXiv:2401.02051},
  year={2024}
}

@article{ye2024reevo,
  title={Reevo: Large language models as hyper-heuristics with reflective evolution},
  author={Ye, Haoran and Wang, Jiarui and Cao, Zhiguang and Berto, Federico and Hua, Chuanbo and Kim, Haeyeon and Park, Jinkyoo and Song, Guojie},
  journal={Advances in neural information processing systems},
  volume={37},
  pages={43571--43608},
  year={2024}
}

@article{zheng2025monte,
  title={Monte carlo tree search for comprehensive exploration in llm-based automatic heuristic design},
  author={Zheng, Zhi and Xie, Zhuoliang and Wang, Zhenkun and Hooi, Bryan},
  journal={arXiv preprint arXiv:2501.08603},
  year={2025}
}

@inproceedings{shi2026generalizable,
  title={Generalizable heuristic generation through LLMs with meta-optimization},
  author={Shi, Yiding and Zhou, Jianan and Song, Wen and Bi, Jieyi and Wu, Yaoxin and Cao, Zhiguang and Zhang, Jie},
  booktitle={International Conference on Learning Representations},
  volume={2026},
  pages={119314--119349},
  year={2026}
}

@article{deb2002fast,
  title={A fast and elitist multiobjective genetic algorithm: NSGA-II},
  author={Deb, Kalyanmoy and Pratap, Amrit and Agarwal, Sameer and Meyarivan, TAMT},
  journal={IEEE transactions on evolutionary computation},
  volume={6},
  number={2},
  pages={182--197},
  year={2002},
  publisher={Ieee}
}

@article{zhang2007moea,
  title={MOEA/D: A multiobjective evolutionary algorithm based on decomposition},
  author={Zhang, Qingfu and Li, Hui},
  journal={IEEE Transactions on evolutionary computation},
  volume={11},
  number={6},
  pages={712--731},
  year={2007},
  publisher={IEEE}
}

@incollection{paquete2004pareto,
  title={Pareto local optimum sets in the biobjective traveling salesman problem: An experimental study},
  author={Paquete, Luis and Chiarandini, Marco and St{\"u}tzle, Thomas},
  booktitle={Metaheuristics for multiobjective optimisation},
  pages={177--199},
  year={2004},
  publisher={Springer}
}

@article{laumanns2004running,
  title={Running time analysis of multiobjective evolutionary algorithms on pseudo-boolean functions},
  author={Laumanns, Marco and Thiele, Lothar and Zitzler, Eckart},
  journal={IEEE Transactions on Evolutionary Computation},
  volume={8},
  number={2},
  pages={170--182},
  year={2004},
  publisher={IEEE}
}

@article{verma2021comprehensive,
  title={A comprehensive review on NSGA-II for multi-objective combinatorial optimization problems},
  author={Verma, Shanu and Pant, Millie and Snasel, Vaclav},
  journal={IEEE access},
  volume={9},
  pages={57757--57791},
  year={2021},
  publisher={IEEE}
}

@article{chen2025reinforced,
  title={A reinforced neighborhood search method combined with genetic algorithm for multi-objective multi-robot transportation system},
  author={Chen, Peng and Liang, Jing and Qiao, Kang-Jia and Song, Hui and Suganthan, Ponnuthurai Nagaratnam and Dai, Lou-Lei and Ban, Xuan-Xuan},
  journal={IEEE Transactions on Intelligent Transportation Systems},
  volume={26},
  number={8},
  pages={12586--12599},
  year={2025},
  publisher={IEEE}
}

@article{li2024evolutionary,
  title={An evolutionary multitasking memetic algorithm for multiobjective distributed heterogeneous welding flow shop scheduling},
  author={Li, Rui and Wang, Ling and Gong, Wenyin and Ming, Fei},
  journal={IEEE Transactions on Evolutionary Computation},
  volume={29},
  number={6},
  pages={2287--2298},
  year={2024},
  publisher={IEEE}
}

@article{knowles2003properties,
  title={Properties of an adaptive archiving algorithm for storing nondominated vectors},
  author={Knowles, Joshua and Corne, David},
  journal={IEEE Transactions on Evolutionary Computation},
  volume={7},
  number={2},
  pages={100--116},
  year={2003},
  publisher={IEEE}
}

@article{lin2022pareto,
  title={Pareto set learning for neural multi-objective combinatorial optimization},
  author={Lin, Xi and Yang, Zhiyuan and Zhang, Qingfu},
  journal={arXiv preprint arXiv:2203.15386},
  year={2022}
}

@article{chen2023neural,
  title={Neural multi-objective combinatorial optimization with diversity enhancement},
  author={Chen, Jinbiao and Zhang, Zizhen and Cao, Zhiguang and Wu, Yaoxin and Ma, Yining and Ye, Te and Wang, Jiahai},
  journal={Advances in Neural Information Processing Systems},
  volume={36},
  pages={39176--39188},
  year={2023}
}

@article{zhang2009effective,
  title={An Effective Hybrid Particle Swarm Optimization Algorithm for Multi-Objective Flexible Job-Shop Scheduling Problem},
  author={Zhang, Guohui and Shao, Xinyu and Li, Peigen and Gao, Liang},
  journal={Computers \& Industrial Engineering},
  volume={56},
  number={4},
  pages={1309--1318},
  year={2009},
  publisher={Elsevier}
}

@article{li2010effective,
  title={An Effective Hybrid Tabu Search Algorithm for Multi-Objective Flexible Job-Shop Scheduling Problems},
  author={Li, Junqing and Pan, Quanke and Liang, Yun-Chia},
  journal={Computers \& Industrial Engineering},
  volume={59},
  number={4},
  pages={647--662},
  year={2010},
  publisher={Elsevier}
}

@incollection{lust2010multiobjective,
  title={The multiobjective traveling salesman problem: A survey and a new approach},
  author={Lust, Thibaut and Teghem, Jacques},
  booktitle={Advances in Multi-Objective Nature Inspired Computing},
  pages={119--141},
  year={2010},
  publisher={Springer}
}

@article{chen2023efficient,
  title={Efficient meta neural heuristic for multi-objective combinatorial optimization},
  author={Chen, Jinbiao and Wang, Jiahai and Zhang, Zizhen and Cao, Zhiguang and Ye, Te and Chen, Siyuan},
  journal={Advances in neural information processing systems},
  volume={36},
  pages={56825--56837},
  year={2023}
}

@article{zhang2025neuro,
  title={Neuro-PLS: A Generalizable Local Search Framework for Multi-objective Combinatorial Optimization},
  author={Zhang, Haotian and Shi, Jialong and Sun, Jianyong and Zhang, Qingfu and Xu, Zongben},
  journal={IEEE Transactions on Evolutionary Computation},
  year={2025},
  publisher={IEEE}
}

@inproceedings{chen2025rethinking,
  title={Rethinking neural multi-objective combinatorial optimization via neat weight embedding},
  author={Chen, Jinbiao and Cao, Zhiguang and Wang, Jiahai and Wu, Yaoxin and Qin, Hanzhang and Zhang, Zizhen and Gong, Yue-Jiao},
  booktitle={International Conference on Learning Representations},
  volume={2025},
  pages={48510--48529},
  year={2025}
}

@article{laumanns2002combining,
  title={Combining Convergence and Diversity in Evolutionary Multiobjective Optimization},
  author={Laumanns, Marco and Thiele, Lothar and Deb, Kalyanmoy and Zitzler, Eckart},
  journal={Evolutionary Computation},
  volume={10},
  number={3},
  pages={263--282},
  year={2002},
  doi={10.1162/106365602760234108}
}

@inproceedings{yao2025multi,
  title={Multi-objective evolution of heuristic using large language model},
  author={Yao, Shunyu and Liu, Fei and Lin, Xi and Lu, Zhichao and Wang, Zhenkun and Zhang, Qingfu},
  booktitle={Proceedings of the AAAI Conference on Artificial Intelligence},
  volume={39},

  pages={27144--27152},
  year={2025}
}

@article{fornies2025remoh,
  title={REMoH: A reflective evolution of multi-objective heuristics approach via large language models},
  author={Forni{\'e}s-Tabuenca, Diego and Uribe, Alejandro and Otamendi, Urtzi and Artetxe, Arkaitz and Rivera, Juan Carlos and de Lacalle, Oier Lopez},
  journal={arXiv preprint arXiv:2506.07759},
  year={2025}
}

@inproceedings{hieu2026pareto,
  title={Pareto-grid-guided large language models for fast and high-quality heuristics design in multi-objective combinatorial optimization},
  author={Hieu, Ha Minh and Phan, Hung and Doan, Tung Duy and Dao, Tung and Tran, Cong Dao and Binh, Huynh Thi Thanh},
  booktitle={Proceedings of the AAAI Conference on Artificial Intelligence},
  volume={40},

  pages={36964--36972},
  year={2026}
}

@article{brandimarte1993routing,
  title={Routing and scheduling in a flexible job shop by tabu search},
  author={Brandimarte, Paolo},
  journal={Annals of Operations research},
  volume={41},
  number={3},
  pages={157--183},
  year={1993},
  publisher={Springer}
}

@inproceedings{qiu2026e2oc,
  title={Evolving Interdependent Operators with Large Language Models for Multi-Objective Combinatorial Optimization},
  author={Qiu, Junhao and Chen, Xin and Ge, Liang and Lin, Liyong and Lu, Zhichao and Zhang, Qingfu},
  booktitle={International Conference on Machine Learning (ICML)},
  year={2026},
  url={https://arxiv.org/abs/2601.17899}
}

@article{lewis2020retrieval,
  title={Retrieval-augmented generation for knowledge-intensive nlp tasks},
  author={Lewis, Patrick and Perez, Ethan and Piktus, Aleksandra and Petroni, Fabio and Karpukhin, Vladimir and Goyal, Naman and K{\"u}ttler, Heinrich and Lewis, Mike and Yih, Wen-tau and Rockt{\"a}schel, Tim and others},
  journal={Advances in neural information processing systems},
  volume={33},
  pages={9459--9474},
  year={2020}
}

@article{yao2022react,
  title={React: Synergizing reasoning and acting in language models},
  author={Yao, Shunyu and Zhao, Jeffrey and Yu, Dian and Du, Nan and Shafran, Izhak and Narasimhan, Karthik and Cao, Yuan},
  journal={arXiv preprint arXiv:2210.03629},
  year={2022}
}

@article{shinn2023reflexion,
  title={Reflexion: Language agents with verbal reinforcement learning},
  author={Shinn, Noah and Cassano, Federico and Gopinath, Ashwin and Narasimhan, Karthik and Yao, Shunyu},
  journal={Advances in neural information processing systems},
  volume={36},
  pages={8634--8652},
  year={2023}
}

@article{pan2026natural,
  title={Natural-language agent harnesses},
  author={Pan, Linyue and Zou, Lexiao and Guo, Shuo and Ni, Jingchen and Zheng, Hai-Tao},
  journal={arXiv preprint arXiv:2603.25723},
  year={2026}
}

@article{li2026agent,
  title={Agent harness engineering: A survey},
  author={Li, Junjie and Xiao, Xi and Zhang, Yunbei and Liu, Chen and Zhao, Lin and Liao, Xiaoying and Ji, Yingrui and Wang, Janet and Gu, Jianyang and Ge, Yingqiang and others},
  journal={OpenReview preprint},
  year={2026}
}

@inproceedings{hu2025automated,
  title={Automated design of agentic systems},
  author={Hu, Shengran and Lu, Cong and Clune, Jeff},
  booktitle={International Conference on Learning Representations},
  volume={2025},
  pages={21344--21377},
  year={2025}
}

@inproceedings{zhang2025aflow,
  title={Aflow: Automating agentic workflow generation},
  author={Zhang, Jiayi and Xiang, Jinyu and Yu, Zhaoyang and Teng, Fengwei and Chen, Xionghui and Chen, Jiaqi and Zhuge, Mingchen and Cheng, Xin and Hong, Sirui and Wang, Jinlin and others},
  booktitle={International Conference on Learning Representations},
  volume={2025},
  pages={34040--34077},
  year={2025}
}

@inproceedings{zhang2026darwin,
  title={Darwin G{\"o}del machine: open-ended evolution of self-improving agents},
  author={Zhang, Jenny and Hu, Shengran and Lu, Cong and Lange, Robert and Clune, Jeff},
  booktitle={International Conference on Learning Representations},
  volume={2026},
  pages={104223--104294},
  year={2026}
}

@article{qu2026coral,
  title={Coral: Towards autonomous multi-agent evolution for open-ended discovery},
  author={Qu, Ao and Zheng, Han and Zhou, Zijian and Yan, Yihao and Tang, Yihong and Ong, Shao Yong and Hong, Fenglu and Zhou, Kaichen and Jiang, Chonghe and Kong, Minwei and others},
  journal={arXiv preprint arXiv:2604.01658},
  year={2026}
}

@misc{lee2026metaharnessendtoendoptimizationmodel,
      title={Meta-Harness: End-to-End Optimization of Model Harnesses},
      author={Yoonho Lee and Roshen Nair and Qizheng Zhang and Kangwook Lee and Omar Khattab and Chelsea Finn},
      year={2026},
      eprint={2603.28052},
      archivePrefix={arXiv},
      primaryClass={cs.AI},
      url={https://arxiv.org/abs/2603.28052},
}

@misc{zhang2026selfharnessharnessesimprove,
      title={Self-Harness: Harnesses That Improve Themselves},
      author={Hangfan Zhang and Shao Zhang and Kangcong Li and Chen Zhang and Yang Chen and Yiqun Zhang and Lei Bai and Shuyue Hu},
      year={2026},
      eprint={2606.09498},
      archivePrefix={arXiv},
      primaryClass={cs.CL},
      url={https://arxiv.org/abs/2606.09498},
}

@book{ehrgott2005multicriteria,
  title={Multicriteria optimization},
  author={Ehrgott, Matthias},
  year={2005},
  publisher={Springer}
}

@incollection{ehrgott2016exact,
  title={Exact methods for multi-objective combinatorial optimisation},
  author={Ehrgott, Matthias and Gandibleux, Xavier and Przybylski, Anthony},
  booktitle={Multiple criteria decision analysis: State of the art surveys},
  pages={817--850},
  year={2016},
  publisher={Springer}
}

@article{herzel2021approximation,
  title={Approximation methods for multiobjective optimization problems: A survey},
  author={Herzel, Arne and Ruzika, Stefan and Thielen, Clemens},
  journal={INFORMS Journal on Computing},
  volume={33},
  number={4},
  pages={1284--1299},
  year={2021},
  publisher={INFORMS}
}

@inproceedings{xu2026autoep,
  title={AutoEP: LLMs-driven automation of hyperparameter evolution for metaheuristic algorithms},
  author={Xu, Zhenxing and Zhang, Yizhe and Bao, Weidong and Wang, Hao and Chen, Ming and Ye, Haoran and Jiang, Wenzheng and Yan, Hui and Wang, Ji},
  booktitle={International Conference on Learning Representations},
  volume={2026},
  pages={139927--139945},
  year={2026}
}

@article{lust2012multiobjective,
title={The multiobjective multidimensional knapsack problem: a survey and a new approach},
author={Lust, Thibaut and Teghem, Jacques},
journal={International Transactions in Operational Research},
volume={19},
number={4},
pages={495--520},
year={2012},
publisher={Wiley Online Library}
}

@article{mavrotas2009bomdkp,
  title={Solving the bi-objective multi-dimensional knapsack problem exploiting the concept of core},
  author={Mavrotas, George and Figueira, Jos{\'e} Rui and Florios, Kostas},
  journal={Applied Mathematics and Computation},
  volume={215},
  number={7},
  pages={2502--2514},
  year={2009},
  publisher={Elsevier}
}

@article{huang2025autonomous,
  title={Autonomous multiobjective optimization using large language model},
  author={Huang, Yuxiao and Wu, Shenghao and Zhang, Wenjie and Wu, Jibin and Feng, Liang and Tan, Kay Chen},
  journal={IEEE Transactions on Evolutionary Computation},
  volume={30},
  number={2},
  pages={594--608},
  year={2025},
  publisher={IEEE}
}

@article{jozefowiez2008multi,
  title={Multi-objective vehicle routing problems},
  author={Jozefowiez, Nicolas and Semet, Fr{\'e}d{\'e}ric and Talbi, El-Ghazali},
  journal={European Journal of Operational Research},
  volume={189},
  number={2},
  pages={293--309},
  year={2008},
  publisher={Elsevier},
  doi={10.1016/j.ejor.2007.05.055}
}

@article{zajac2021objectives,
  title={Objectives and methods in multi-objective routing problems: a survey and classification scheme},
  author={Zajac, Sandra and Huber, Sandra},
  journal={European journal of operational research},
  volume={290},
  number={1},
  pages={1--25},
  year={2021},
  publisher={Elsevier}
}

@article{hoogeveen2005multicriteria,
  title={Multicriteria scheduling},
  author={Hoogeveen, Han},
  journal={European Journal of operational research},
  volume={167},
  number={3},
  pages={592--623},
  year={2005},
  publisher={Elsevier}
}

@article{blot2018survey,
  title={Survey and unification of local search techniques in metaheuristics for multi-objective combinatorial optimisation: A. Blot et al.},
  author={Blot, Aymeric and Kessaci, Marie-{\'E}l{\'e}onore and Jourdan, Laetitia},
  journal={Journal of Heuristics},
  volume={24},
  number={6},
  pages={853--877},
  year={2018},
  publisher={Springer}
}

@article{vanstein2025llamea,
  title={LLaMEA: A Large Language Model Evolutionary Algorithm for Automatically Generating Metaheuristics},
  author={van Stein, Niki and B{\"a}ck, Thomas},
  journal={IEEE Transactions on Evolutionary Computation},
  volume={29},
  number={2},
  pages={331--345},
  year={2025},
  doi={10.1109/TEVC.2024.3497793}
}

@inproceedings{li2026llameabo,
  title={Llamea-bo: A large language model evolutionary algorithm for automatically generating bayesian optimization algorithms},
  author={Li, Wenhu and van Stein, Niki and B{\"a}ck, Thomas and Raponi, Elena},
  booktitle={Proceedings of the Genetic and Evolutionary Computation Conference},
  pages={909--918},
  year={2026}
}

@article{novikov2025alphaevolve,
  title={Alphaevolve: A coding agent for scientific and algorithmic discovery},
  author={Novikov, Alexander and V{\~u}, Ng{\^a}n and Eisenberger, Marvin and Dupont, Emilien and Huang, Po-Sen and Wagner, Adam Zsolt and Shirobokov, Sergey and Kozlovskii, Borislav and Ruiz, Francisco JR and Mehrabian, Abbas and others},
  journal={arXiv preprint arXiv:2506.13131},
  year={2025}
}

@article{ma2025toward,
  title={Toward automated algorithm design: A survey and practical guide to meta-black-box-optimization},
  author={Ma, Zeyuan and Guo, Hongshu and Gong, Yue-Jiao and Zhang, Jun and Tan, Kay Chen},
  journal={IEEE Transactions on Evolutionary Computation},
  volume={30},
  number={2},
  pages={667--687},
  year={2025},
  publisher={IEEE}
}

@inproceedings{garivier2011upper,
  title={On upper-confidence bound policies for switching bandit problems},
  author={Garivier, Aur{\'e}lien and Moulines, Eric},
  booktitle={International conference on algorithmic learning theory},
  pages={174--188},
  year={2011},
  organization={Springer}
}

@inproceedings{liao2013solving,
  title={Solving the biobjective selective pickup and delivery problem with memetic algorithm},
  author={Liao, Xin-Lan and Ting, Chuan-Kang},
  booktitle={2013 IEEE Symposium on Computational Intelligence in Production and Logistics Systems (CIPLS)},
  pages={107--114},
  year={2013},
  organization={IEEE}
}

@article{xu2023pareto,
  title={A Pareto Front grid guided multi-objective evolutionary algorithm},
  author={Xu, Ying and Zhang, Huan and Huang, Lei and Qu, Rong and Nojima, Yusuke},
  journal={Applied Soft Computing},
  volume={136},
  pages={110095},
  year={2023},
  publisher={Elsevier}
}

@misc{xu2026dreamcop,
  title         = {{Dream-COP}: Guiding Algorithm Discovery for Combinatorial Optimization Problems in an Experience World},
  author        = {Xu, Kuan and Wang, Jianming and Zhang, Ni and Wu, Yaoxin and Wang, Yong and Cao, Zhiguang},
  year          = {2026},
  eprint        = {2610.36618},
  archivePrefix = {arXiv},
  primaryClass  = {cs.LG},
  url           = {https://arxiv.org/abs/2610.36618}
}

@inproceedings{zhang2026agentic,
  title={An agentic framework with LLMs for solving complex vehicle routing problems},
  author={Zhang, Ni and Cao, Zhiguang and Zhou, Jianan and Zhang, Cong and Ong, Yew-Soon},
  booktitle={International Conference on Learning Representations},
  volume={2026},
  pages={114166--114206},
  year={2026}
}

\clearpage
\onecolumn
\appendices

\startcontents[appendix]

\section*{Appendix Contents}
\setcounter{tocdepth}{2}
\printcontents[appendix]{}{1}{}

\newpage
\section{Multi-Objective Performance Metrics}
\label{app:mo_metrics}

We evaluate the quality of the Pareto-front approximations using two widely adopted multi-objective optimization metrics, namely Hypervolume (HV) and Inverted Generational Distance (IGD). Both metrics jointly reflect
convergence toward high-quality trade-off solutions and the coverage
of the obtained Pareto front, but evaluate these properties from
different perspectives. Higher HV and lower IGD indicate better
performance.

\subsection{Hypervolume}
\label{app:metric_hv}

\suppheading{Hypervolume.}
Given a non-dominated objective set
$P=\{\mathbf{y}^{1},\ldots,\mathbf{y}^{K}\}\subset\mathbb{R}^{M}$
and a reference point
$\mathbf{r}=(r_1,\ldots,r_M)^\top$ that is dominated by all points
of interest under the minimization convention, the Hypervolume (HV)
measures the volume of the objective-space region dominated by $P$
and bounded by $\mathbf{r}$.
Formally,
\begin{equation}
    \mathrm{HV}_{\mathbf{r}}(P)
    =
    \lambda_M
    \left(
        \bigcup_{\mathbf{y}\in P}
        [y_1,r_1]
        \times\cdots\times
        [y_M,r_M]
    \right),
    \label{eq:hv}
\end{equation}
where $\lambda_M(\cdot)$ denotes the $M$-dimensional Lebesgue
measure.

HV simultaneously rewards convergence and diversity because solutions closer to the Pareto front extend the dominated region toward better objective values, while well-distributed non-dominated solutions cover a broader region of the objective space. Therefore, a larger HV value indicates a better Pareto-front approximation.

Because different MOCOPs can have substantially different objective scales, making direct comparison across problem settings difficult, we report normalized HV. Given an ideal point
$\mathbf{z}=(z_1,\ldots,z_M)^\top$, the normalized HV is computed as
\begin{equation}
    \mathrm{HV}'_{\mathbf{r}}(P)
    =
    \frac{
        \mathrm{HV}_{\mathbf{r}}(P)
    }{
        \prod_{m=1}^{M}|r_m-z_m|
    }.
    \label{eq:normalized_hv}
\end{equation}
The same reference and ideal points are used for all compared methods
on a given benchmark setting. Their construction and the values used for each problem are detailed in Appendix~\ref{app:metric_normalization}.

\subsection{Inverted Generational Distance}
\label{app:metric_igd}

\suppheading{Inverted Generational Distance.}
IGD evaluates an approximation from the perspective of a reference
Pareto front. Let $Q$ denote a reference set representing the target
Pareto front and let $P$ denote the non-dominated set obtained by an
algorithm. IGD is defined as
\begin{equation}
    \mathrm{IGD}(P,Q)
    =
    \frac{1}{|Q|}
    \sum_{\mathbf{q}\in Q}
    \min_{\mathbf{p}\in P}
    \left\|
        \mathbf{q}-\mathbf{p}
    \right\|_2,
    \label{eq:igd}
\end{equation}
where $\|\cdot\|_2$ denotes Euclidean distance.

For every point on the reference front, IGD measures the distance to
its closest solution in the approximation and then averages these
distances over the entire reference front. Consequently, a low IGD requires both good convergence and sufficient coverage, since an approximation concentrated in only a small region of the Pareto front will retain large distances to uncovered reference points. Lower IGD is therefore better.

When an exact Pareto front is unavailable, we construct the reference
front from the union of the solution sets produced by all compared
methods and retain the non-dominated subset. The same reference front
is then used to evaluate every method under the corresponding
benchmark setting.

\subsection{Objective Normalization and Reference Points}
\label{app:metric_normalization}

All objectives are represented under a minimization
convention before metric computation. We use fixed
problem-specific and size-specific reference and ideal
points, as listed in Table~\ref{tab:hv_reference_points}.
These points are shared by the compared methods within
each setting.
\begin{table}[t]
\centering
\small
\caption{Reference points and ideal points for the MOCOP benchmarks.}
\label{tab:hv_reference_points}
\setlength{\tabcolsep}{12pt}
\renewcommand{\arraystretch}{1.05}
\begin{tabular}{cccc}
\toprule
Problem & Size & Reference point $\mathbf{r}$ & Ideal point $\mathbf{z}$ \\
\midrule

\multirow{3}{*}{Bi-KP}
& 50  & $(-5,-5)$   & $(-30,-30)$ \\
& 100 & $(-20,-20)$ & $(-50,-50)$ \\
& 200 & $(-30,-30)$ & $(-75,-75)$ \\
\midrule

\multirow{3}{*}{Bi-MDKP}
& 50  & $(-5,-5)$   & $(-30,-30)$ \\
& 100 & $(-20,-20)$ & $(-50,-50)$ \\
& 200 & $(-30,-30)$ & $(-85,-85)$ \\
\midrule

\multirow{3}{*}{Bi-CVRP}
& 20  & $(30,8)$ & $(0,0)$ \\
& 50  & $(45,8)$ & $(0,0)$ \\
& 100 & $(80,8)$ & $(0,0)$ \\
\midrule

\multirow{3}{*}{Bi-SPDP}
& 20  & $(31.1,68.2)$ & $(0,0)$ \\
& 50  & $(77.8,160.6)$ & $(0,0)$ \\
& 100 & $(155.6,325.6)$ & $(0,0)$ \\
\midrule

\multirow{3}{*}{Bi-TSP}
& 20  & $(20,20)$ & $(0,0)$ \\
& 50  & $(35,35)$ & $(0,0)$ \\
& 100 & $(65,65)$ & $(0,0)$ \\
\midrule

\multirow{3}{*}{Tri-TSP}
& 20  & $(20,20,20)$ & $(0,0,0)$ \\
& 50  & $(35,35,35)$ & $(0,0,0)$ \\
& 100 & $(65,65,65)$ & $(0,0,0)$ \\
\midrule

Bi-FJSP
& All & $(1.1,1.1)$ & $(0,0)$ \\
\midrule

Tri-FJSP
& All & $(1.1,1.1,1.1)$ & $(0,0,0)$ \\
\bottomrule
\end{tabular}
\end{table}

\suppheading{FJSP objective normalization.}
For Bi-FJSP and Tri-FJSP, raw objective values are divided
by an instance-specific scale determined from processing
times, independently of the solutions produced by the
compared methods. For instance $I$, this scale is
\begin{equation}
    S_I =
    \sum_{o\in\mathcal{O}_I}
    \max_{k\in\mathcal{M}_o} p_{ok},
\end{equation}
where $\mathcal{O}_I$ is the set of operations,
$\mathcal{M}_o$ contains the eligible machines for
operation $o$, and $p_{ok}$ is its processing time on
machine $k$. The normalized objectives are
\begin{equation}
    f'_m(\mathbf{x}) = \frac{f_m(\mathbf{x})}{S_I}.
    \label{eq:objective_normalization}
\end{equation}
In this normalized objective space, we use the ideal
point $\mathbf{z}=\mathbf{0}$ and the HV reference point
\begin{equation}
    \mathbf{r}^{*}=(1.1,\ldots,1.1)^\top,
    \label{eq:normalized_hv_ref}
\end{equation}
with two components for Bi-FJSP and three for Tri-FJSP.
This scaling does not use empirical ideal or nadir points
estimated from the compared methods' solution fronts.

\suppheading{Other benchmarks and HV scaling.}
For the knapsack, routing, and traveling salesman
benchmarks, HV is computed using the fixed reference
and ideal points in Table~\ref{tab:hv_reference_points}.
The reported HV is normalized by the reference box volume:
\begin{equation}
    \mathrm{HV}_{\mathrm{norm}}(\mathcal{P})
    =
    \frac{\mathrm{HV}(\mathcal{P};\mathbf{r})}
         {\prod_{m=1}^{M}(r_m-z_m)},
\end{equation}
where $\mathcal{P}$, $\mathbf{r}$, and $\mathbf{z}$
are expressed in the same objective space.

\section{Benchmark Problems}
\label{app:benchmark_problems}

We evaluate \hemo on eight multi-objective combinatorial optimization tasks spanning four well-studied families, covering \emph{knapsack}, \emph{routing}, \emph{traveling salesman}, and \emph{scheduling}. These families are among the most extensively studied classes of MOCOPs in the evolutionary optimization literature~\cite{verma2021comprehensive}. For each family, one task is used for harness self-evolution and a related task for intra-family cross-task generalization, with source-to-target pairs of Bi-KP to Bi-MDKP, Bi-CVRP to Bi-SPDP, Bi-TSP to Tri-TSP, and Bi-FJSP to Tri-FJSP. The benchmark configuration is summarized in Table~\ref{tab:benchmark_overview}. Detailed harness self-evolution and inference settings are provided in Appendix~\ref{app:training_inference_protocols}.

\begin{table}[b]
\centering
\small
\caption{Overview of the MOCOP benchmarks used in our experiments.
Each synthetic problem size contains $50$ fixed test instances. For FJSP, Mk15 is used during operator design for NSGA-II,
while all 15 Brandimarte instances are used for final evaluation.}
\label{tab:benchmark_overview}
\begin{tabularx}{\textwidth}{@{}l l X l@{}}
\toprule
Family & Problem & Objectives & Final test instances \\
\midrule
Knapsack
& Bi-KP
& two collected profits
& $n\in\{50,100,200\}$ \\

Knapsack
& Bi-MDKP
& two collected profits
& $n\in\{50,100,200\}$ \\

Routing
& Bi-CVRP
& total distance; longest route
& $n\in\{20,50,100\}$ \\

Routing
& Bi-SPDP
& total distance; required vehicle capacity
& $n\in\{20,50,100\}$ \\

TSP
& Bi-TSP
& two tour lengths
& $n\in\{20,50,100\}$ \\

TSP
& Tri-TSP
& three tour lengths
& $n\in\{20,50,100\}$ \\

Scheduling
& Bi-FJSP
& makespan; maximum machine load
& Mk01--Mk15 \\

Scheduling
& Tri-FJSP
& makespan; maximum machine load; total machine load
& Mk01--Mk15 \\
\bottomrule
\end{tabularx}
\end{table}

\subsection{Knapsack Problems}
\label{app:benchmark_knapsack}

\suppheading{Bi-objective Knapsack Problem (Bi-KP).}
The bi-objective knapsack problem extends the classical $0$--$1$
knapsack problem by associating each item with two profit values.
Given $n$ items, let $w_i$ denote the weight of item $i$, and let
$p_i^{(1)}$ and $p_i^{(2)}$ denote its profits under the two
objectives. A solution is represented by a binary vector
$\mathbf{x}\in\{0,1\}^{n}$, where $x_i=1$ indicates that item $i$
is selected. The problem is formulated as
\begin{equation}
    \max_{\mathbf{x}\in\{0,1\}^{n}}
    \left(
        \sum_{i=1}^{n}p_i^{(1)}x_i,\,
        \sum_{i=1}^{n}p_i^{(2)}x_i
    \right),
\end{equation}
subject to
\begin{equation}
    \sum_{i=1}^{n}w_i x_i \leq C,
\end{equation}
where $C$ denotes the knapsack capacity.

The two optimization backbones use different numerical coefficient
scales for Bi-KP. In the SEMO implementation, the weights $w_i$ and
the two profits $p_i^{(1)}$ and $p_i^{(2)}$ are sampled independently
from $\operatorname{Unif}[0,1)$. The capacity follows
\begin{equation}
    C =
    \begin{cases}
        12.5, & 50 \leq n < 100,\\
        25.0, & 100 \leq n \leq 200.
    \end{cases}
\end{equation}
Thus, the tested instances use capacities $12.5$, $25$, and $25$
for $n=50$, $100$, and $200$, respectively.

In the NSGA-II implementation, the weights and both profits are
sampled independently from the discrete uniform distribution
$\operatorname{Unif}\{1,\ldots,100\}$, and the capacity is determined
from the generated weights as
\begin{equation}
    C =
    \left\lfloor
        0.5\sum_{i=1}^{n}w_i
    \right\rfloor .
\end{equation}
Both implementations are evaluated at
$n\in\{50,100,200\}$, using $50$ fixed test instances for each size.

\suppheading{Bi-objective Multi-dimensional Knapsack Problem (Bi-MDKP).}
Bi-MDKP generalizes Bi-KP by introducing multiple resource
constraints~\cite{lust2012multiobjective}. Each item $i$ has two
profits $p_i^{(1)}$ and $p_i^{(2)}$ and consumes $w_{ji}$ units of
resource $j$. With $m$ resource constraints, the problem is
\begin{equation}
    \max_{\mathbf{x}\in\{0,1\}^{n}}
    \left(
        \sum_{i=1}^{n}p_i^{(1)}x_i,\,
        \sum_{i=1}^{n}p_i^{(2)}x_i
    \right),
\end{equation}
subject to
\begin{equation}
    \sum_{i=1}^{n}w_{ji}x_i \leq C_j,
    \qquad j=1,\ldots,m.
\end{equation}
We use $m=5$ resource constraints and a tightness factor
$\tau=0.5$.

In the SEMO implementation, all resource consumptions $w_{ji}$ and
profits $p_i^{(1)},p_i^{(2)}$ are independently sampled from
$\operatorname{Unif}[0,1)$, and the capacity of each resource is
\begin{equation}
    C_j
    =
    \tau\sum_{i=1}^{n}w_{ji}
    =
    0.5\sum_{i=1}^{n}w_{ji}.
\end{equation}

In the NSGA-II implementation, the resource consumptions and profits
are independently sampled from
$\operatorname{Unif}\{1,\ldots,100\}$, with
\begin{equation}
    C_j
    =
    \left\lfloor
        \tau\sum_{i=1}^{n}w_{ji}
    \right\rfloor
    =
    \left\lfloor
        0.5\sum_{i=1}^{n}w_{ji}
    \right\rfloor .
\end{equation}
We evaluate Bi-MDKP at $n\in\{50,100,200\}$, using $50$ fixed test
instances for each size. The formulation follows the standard
bi-objective multi-dimensional knapsack setting~\cite{mavrotas2009bomdkp}.

\subsection{Routing Problems}
\label{app:benchmark_routing}

\suppheading{Bi-objective Capacitated Vehicle Routing Problem (Bi-CVRP).}
Bi-CVRP considers one depot and $n$ customer nodes. Each customer $i$
has a demand $d_i$, and a homogeneous fleet of vehicles with capacity
$Q$ must serve every customer exactly once. Each route starts and ends
at the depot, and the total demand assigned to a route cannot exceed
$Q$. Given a set of routes $\mathcal{R}$, we minimize
\begin{equation}
    f_1(\mathcal{R})
    =
    \sum_{r\in\mathcal{R}}\mathrm{len}(r),
    \qquad
    f_2(\mathcal{R})
    =
    \max_{r\in\mathcal{R}}\mathrm{len}(r).
\end{equation}
The first objective measures total routing cost, whereas the second
penalizes disproportionately long individual routes and therefore
encourages workload balance among vehicles.

For every instance, depot and customer coordinates are sampled
independently and uniformly from $[0,1)^2$, and customer demands are
sampled independently from
$\operatorname{Unif}\{1,\ldots,9\}$. Vehicle capacities are set to
$Q=30$, $40$, and $50$ for $n=20$, $50$, and $100$, respectively.
We evaluate $50$ fixed instances at each size
$n\in\{20,50,100\}$.

\suppheading{Bi-objective Selective Pickup and Delivery Problem (Bi-SPDP).}
Bi-SPDP jointly minimizes travel distance and required vehicle
capacity~\cite{liao2013solving}. A single vehicle starts and ends at
a depot, visits every delivery node, and selects a subset of pickup
nodes that supplies the required commodities. Pickups and deliveries are not paired, so commodities collected at any pickup node can satisfy any subsequent delivery.

Let $q_i>0$ denote the supply at a pickup node and $q_i<0$ the
signed demand at a delivery node, with $q_0=0$ at the depot.
For a route $\pi=(0,\pi_1,\ldots,\pi_m,0)$, define the vehicle
load after the $k$th visit as
\begin{equation}
    L_k=\sum_{j=1}^{k}q_{\pi_j}, \qquad L_0=0.
\end{equation}
A feasible route visits each selected node once, includes all
delivery nodes, and satisfies $L_k\geq 0$ throughout the route.
The two minimization objectives are
\begin{equation}
    f_1(\pi)=\sum_{k=0}^{m}
        \|\mathbf{c}_{\pi_{k+1}}-\mathbf{c}_{\pi_k}\|_2,
    \qquad
    f_2(\pi)=\max_{0\leq k\leq m} L_k,
\end{equation}
where $\pi_0=\pi_{m+1}=0$ and $\mathbf{c}_i$ denotes node coordinates.
Thus, vehicle capacity is an objective determined by the maximum
route load, not a fixed capacity constraint.

We construct synthetic instances of this formulation using our own
generator. Here, size $n$ counts all nodes, including the depot.
Coordinates are sampled independently and uniformly from $[0,1)^2$.
The $n-1$ non-depot nodes are randomly partitioned into
$\lfloor(n-1)/2\rfloor$ pickup nodes and the remaining delivery nodes.
Pickup supplies are sampled uniformly from $\{1,\ldots,9\}$ and
delivery demand magnitudes from $\{1,\ldots,6\}$.
If total supply is insufficient, pickup supplies are incremented
one unit at a time in round-robin order until they cover total demand.
We generate $50$ fixed test instances for each
$n\in\{20,50,100\}$.

For each size, the test reference point is
$(1.1n\sqrt{2},\,1.1S_n)$, where $S_n$ is the largest total pickup
supply among its test instances. The ideal point is $(0,0)$.
The Bi-SPDP reference points in Table~\ref{tab:hv_reference_points}
are rounded to one decimal place for presentation only.
Evaluation uses the full-precision reference points stored with the
instances.

\subsection{Traveling Salesman Problems}
\label{app:benchmark_tsp}

\suppheading{Bi-objective and Tri-objective TSP.}
The multi-objective traveling salesman problem (MOTSP) requires a
Hamiltonian tour that visits every city exactly once and returns to
its starting city. We follow the commonly used synthetic Euclidean
MOTSP construction in which each objective is associated with an
independent two-dimensional coordinate space~\cite{lust2010multiobjective}.

For an $M$-objective instance, city $i$ is associated with
\begin{equation}
    \mathbf{c}^{(m)}_i\in[0,1)^2,
    \qquad m=1,\ldots,M,
\end{equation}
where all coordinates are sampled independently. The edge cost under
objective $m$ is
\begin{equation}
    d_{ij}^{(m)}
    =
    \left\|
        \mathbf{c}^{(m)}_i-\mathbf{c}^{(m)}_j
    \right\|_2.
\end{equation}
For a tour
$\pi=(\pi_1,\ldots,\pi_n)$, its objective value is
\begin{equation}
    f_m(\pi)
    =
    \sum_{k=1}^{n-1}
    d_{\pi_k,\pi_{k+1}}^{(m)}
    +
    d_{\pi_n,\pi_1}^{(m)}.
\end{equation}

Bi-TSP uses $M=2$ independent coordinate spaces, while Tri-TSP uses
$M=3$. Hence, one instance can equivalently be represented by
coordinates in $[0,1)^{4}$ and $[0,1)^{6}$, respectively. We
evaluate both variants at $n\in\{20,50,100\}$, with $50$ fixed test
instances at each size.

\subsection{Scheduling Problems}
\label{app:benchmark_scheduling}

\suppheading{Bi-objective and Tri-objective Flexible Job Shop Scheduling.}
The flexible job shop scheduling problem (FJSP) consists of a set of
jobs, each containing an ordered sequence of operations. Unlike the
standard job shop scheduling problem, each operation can be processed
by one of several eligible machines, with a potentially different
processing time on each machine. Operations belonging to the same job
must satisfy their precedence constraints, and each machine can
process at most one operation at any time.

Following established multi-objective FJSP formulations~\cite{zhang2009effective,li2010effective}, we use three common scheduling objectives.
Let $C_j$ denote the completion time of job $j$, and let $L_k$ denote
the total processing load assigned to machine $k$. We define
\begin{equation}
    C_{\max}
    =
    \max_j C_j,
\end{equation}
\begin{equation}
    L_{\max}
    =
    \max_k L_k,
\end{equation}
and
\begin{equation}
    L_{\mathrm{sum}}
    =
    \sum_k L_k.
\end{equation}
Bi-FJSP minimizes
$(C_{\max},L_{\max})$, whereas Tri-FJSP minimizes
$(C_{\max},L_{\max},L_{\mathrm{sum}})$.

For SEMO, heuristic design uses 15 synthetic FJSP instances,
each with 20 jobs and a machine count in $\{5,10,15\}$.
Each job contains 2 to 10 operations, each operation is eligible
for 1 to 5 machines, and processing times are sampled uniformly
from the integers 1 to 99.
For NSGA-II, operator design instead uses the Brandimarte
instance Mk15.
Consequently, all Brandimarte instances are out of distribution
for SEMO, whereas Mk15 is in distribution and Mk01--Mk14 are
held out for NSGA-II.


The held-out FJSP test suite therefore covers $n\in\{10,15,20,30\}$ jobs and machine counts ranging from $4$ to $15$. Compared with the knapsack and routing families, FJSP also introduces a qualitatively different decision structure in which each candidate jointly determines an operation sequence and a machine assignment while satisfying precedence and machine-capacity constraints.

\section{Multi-Objective Optimization Backbones}
\label{app:optimization_backbones}

\hemo is evaluated with two multi-objective evolutionary backbones,
SEMO and NSGA-II, which expose substantially different search
structures and operator design spaces. This section describes their
procedures and experimental configurations, and discusses why both
backbones are considered.

\subsection{SEMO}
\label{app:backbone_semo}

Simple Evolutionary Multi-objective Optimization (SEMO) is a
minimalist Pareto-based evolutionary algorithm that maintains a set
of mutually non-dominated solutions~\cite{laumanns2004running}. Let
$\mathcal{P}_t$ denote the Pareto archive at iteration $t$. Each SEMO
iteration consists of the following steps:

\begin{enumerate}[leftmargin=*]
    \item \textbf{Parent selection.}
    A solution $\mathbf{x}\in\mathcal{P}_t$ is selected from the
    current archive.

    \item \textbf{Variation.}
    A problem-specific mutation or neighborhood operator is applied
    to $\mathbf{x}$ to produce a candidate solution $\mathbf{x}'$.

    \item \textbf{Evaluation.}
    The candidate is checked for feasibility and its multi-objective
    fitness vector $\mathbf{f}(\mathbf{x}')$ is evaluated.

    \item \textbf{Pareto archive update.}
    If $\mathbf{x}'$ is dominated by an existing archive member, it
    is rejected. Otherwise, $\mathbf{x}'$ is inserted and all
    solutions dominated by it are removed.
\end{enumerate}

The archive update can be expressed compactly as
\begin{equation}
    \mathcal{P}_{t+1}
    =
    \mathrm{ND}
    \left(
        \mathcal{P}_t\cup\{\mathbf{x}'\}
    \right),
\end{equation}
where $\mathrm{ND}(\cdot)$ denotes the non-dominated subset.

SEMO is particularly suitable for HEMO for several reasons. First,
its deliberately simple control structure limits manually engineered
population-management mechanisms, making changes induced by the AOD
harness easier to isolate. Second, SEMO follows a lightweight
incremental update in which only one offspring is generated and evaluated
at each iteration. This is especially attractive for harness evaluation,
where each generated heuristic must be executed repeatedly across
optimization instances. The original analysis of SEMO also shows that,
despite its simplicity, it can achieve running time of the same order as
a multistart $(1+1)$-EA on the studied bi-objective model problems
~\cite{laumanns2004running}. Third, SEMO maintains a variable-size
archive containing the nondominated solutions discovered so far, without
requiring a fixed population size or additional environmental selection.
This provides a direct representation of the evolving trade-offs while
keeping the underlying solver mechanics minimal.

In our implementation, each SEMO run is initialized with $100$
randomly generated feasible solutions. Candidate solutions are then
generated and evaluated sequentially, with the Pareto archive updated
after every iteration. We distinguish the relatively small search
budget used to evaluate candidates during harness self-evolution
from the substantially larger budget used for final evaluation.
Table~\ref{tab:semo_budgets} summarizes these settings.

\begin{table}[t]
\centering
\small
\caption{SEMO initialization and iteration budgets. ``Design'' denotes
evaluation of candidates during harness self-evolution and inference of \hemo, whereas ``test''
denotes final evaluation of the selected candidates.}
\label{tab:semo_budgets}
\begin{tabular}{lrrrr}
\toprule
Problem
& Design init.
& Design steps
& Test init.
& Test steps \\
\midrule
Bi-KP    & 100 & 2,000 & 100 & 10,000 \\
Bi-MDKP  & 100 & 2,000 & 100 & 10,000 \\
Bi-CVRP  & 100 & 2,000 & 100 & 10,000 \\
Bi-SPDP  & 100 & 2,000 & 100 & 20,000 \\
Bi-TSP   & 100 & 2,000 & 100 & 20,000 \\
Tri-TSP  & 100 & 2,000 & 100 & 20,000 \\
Bi-FJSP  & 100 &   750 & 100 & 15,000 \\
Tri-FJSP & 100 &   750 & 100 & 15,000 \\
\bottomrule
\end{tabular}
\end{table}

\subsection{NSGA-II}
\label{app:backbone_nsga2}

NSGA-II is an elitist, population-based multi-objective evolutionary
algorithm~\cite{deb2002fast}. Its design combines non-dominated
sorting with diversity preservation through crowding distance. Unlike
SEMO, which accepts or rejects every candidate immediately, NSGA-II
collects an entire offspring population before performing
environmental selection at the generation level.

Given a parent population $P_t$ of size $N$, one generation proceeds
as follows:

\begin{enumerate}[leftmargin=*]
\item \textbf{Parent selection.}
Parent solutions are selected by tournament selection. For the
knapsack, routing, and TSP implementations, tournament fitness is
determined by non-dominated rank. In the FJSP implementation,
rank remains the primary criterion and crowding distance is used
to break ties in favor of solutions located in less populated
regions of objective space.

\item \textbf{Variation.}
Problem-specific crossover and mutation operators are applied to
the selected parents to construct an offspring population $Q_t$.

\item \textbf{Population merging.}
The current parent population and its offspring are combined:
\begin{equation}
    R_t=P_t\cup Q_t.
\end{equation}

\item \textbf{Non-dominated sorting.}
The combined population is partitioned into successive
non-dominated fronts
\begin{equation}
    F_1,F_2,\ldots,
\end{equation}
where $F_1$ contains the non-dominated members of $R_t$, $F_2$
contains the non-dominated members after removing $F_1$, and so
forth.

\item \textbf{Crowding-distance estimation.}
Within each front, crowding distance estimates the density of
neighboring solutions in objective space. For an interior solution
$\mathbf{x}$, it is computed as
$$
d(\mathbf{x})
=
\sum_{m=1}^{M}
\frac{
f_m(\mathbf{x}^{\mathrm{next}})
-
f_m(\mathbf{x}^{\mathrm{prev}})
}{
f_m^{\max}-f_m^{\min}
},
$$
where $\mathbf{x}^{\mathrm{prev}}$ and $\mathbf{x}^{\mathrm{next}}$
are the neighboring solutions after sorting the front by objective
$m$. Boundary solutions are assigned infinite crowding distance.
Larger values therefore indicate less densely populated regions and
are preferred when diversity must be preserved.

\item \textbf{Environmental selection.}
Complete fronts are inserted into the next population
$P_{t+1}$ in increasing Pareto-rank order. If the next front
cannot be accommodated in full, the remaining population slots
are filled by selecting the solutions with the largest crowding
distances from that front.
\end{enumerate}

These components, including non-dominated sorting, elitist
preservation, crowding distance, and selection, constitute the
central control mechanisms of NSGA-II~\cite{verma2021comprehensive}.
Compared with SEMO, NSGA-II therefore exposes a richer set of interactions at the population level and delays the effect of individual
variation decisions until environmental selection at the end of a generation.


In our implementation, the standard NSGA-II population management
procedure remains fixed, whereas the task-specific variation
operators exposed to the LLM constitute the inner operator design
space. We distinguish three evaluation stages, namely harness self-evolution, inference with the selected harness, and final testing.
During harness self-evolution, we use small populations and short
generation budgets to compare harness states at low computational
cost. During inference, both the population size and generation
budget are doubled to evaluate candidate operator combinations
more thoroughly and improve the resolution of their quality--runtime
comparisons. For example, a population size of 50 and a generation
budget of 15 become 100 and 30, respectively.
These larger solver budgets allow differences between candidate
operator combinations to emerge over a more substantial optimization process.
Final testing uses separate budgets to assess the selected operators.
Table~\ref{tab:nsga2_budgets} summarizes the harness self-evolution and
final test budgets. For comparisons with \hemo designing SEMO heuristics, all population-based MOEA baselines,
including NSGA-II, MOEA/D, and PFG-MOEA, use a population size of
300 and a generation budget of 300.
For comparisons between baseline NSGA-II and \hemo constructing NSGA-II operators,
both use the same task-specific test budgets listed in
Table~\ref{tab:nsga2_budgets}.

\suppheading{Problem-specific operator design space.}
The inner design space consists of task-specific operator
implementations injected into the fixed NSGA-II backbone.
Table~\ref{tab:nsga2_operator_sequences} lists the design targets
and their execution order, while
Table~\ref{tab:nsga2_operator_meanings} summarizes their roles.
Within each generation, the selected offspring population passes
through these operators sequentially before objective evaluation
and environmental selection. Each operator receives and returns
a chromosome matrix at the population level. Operator names identify stable interfaces and their seed reference operations.
The chromosome representation, interface contracts, and
problem-specific decoding and feasibility safeguards remain fixed.

\begin{table*}[t]
\centering
\scriptsize
\setlength{\tabcolsep}{4pt}
\renewcommand{\arraystretch}{1.18}
\caption{NSGA-II operator design targets and default execution
order across the eight tasks. Every named operator is a design
target. Arrows indicate sequential calls on the offspring
population, not alternative operator choices.}
\label{tab:nsga2_operator_sequences}
\begin{tabularx}{\textwidth}{
@{}l
>{\raggedright\arraybackslash}p{0.25\textwidth}
>{\centering\arraybackslash}X@{}
}
\toprule
Task & Chromosome representation & Design targets in execution order \\
\midrule

Bi-TSP
& City permutation
& \texttt{tsp\_oropt}
  $\rightarrow$ \texttt{tsp\_2opt}
  $\rightarrow$ \texttt{tsp\_3opt} \\

Tri-TSP
& City permutation
& \texttt{tsp\_oropt}
  $\rightarrow$ \texttt{tsp\_2opt}
  $\rightarrow$ \texttt{tsp\_3opt} \\

\midrule

Bi-CVRP
& Customer permutation (giant tour)
& \texttt{vrp\_oropt}
  $\rightarrow$ \texttt{vrp\_swap}
  $\rightarrow$ \texttt{vrp\_2opt} \\

Bi-SPDP
& Signed node permutation
& \texttt{vrp\_oropt}
  $\rightarrow$ \texttt{vrp\_swap}
  $\rightarrow$ \texttt{vrp\_2opt} \\

\midrule

Bi-KP
& Binary item-selection vector
& \texttt{kp\_bitflip}
  $\rightarrow$ \texttt{kp\_crossover}
  $\rightarrow$ \texttt{kp\_repair\_mutation} \\

Bi-MDKP
& Binary item-selection vector
& \texttt{kp\_bitflip}
  $\rightarrow$ \texttt{kp\_crossover}
  $\rightarrow$ \texttt{kp\_repair\_mutation} \\

\midrule

Bi-FJSP
& Operation-priority random keys and machine-option indices
& \texttt{operation\_crossover}
  $\rightarrow$ \texttt{operation\_mutation}
  \newline
  $\rightarrow$ \texttt{machine\_crossover}
  $\rightarrow$ \texttt{machine\_mutation} \\

Tri-FJSP
& Operation-priority random keys and machine-option indices
& \texttt{operation\_crossover}
  $\rightarrow$ \texttt{operation\_mutation}
  \newline
  $\rightarrow$ \texttt{machine\_crossover}
  $\rightarrow$ \texttt{machine\_mutation} \\

\bottomrule
\end{tabularx}
\end{table*}

\begin{table*}[t]
\centering
\scriptsize
\setlength{\tabcolsep}{3.5pt}
\renewcommand{\arraystretch}{1.18}
\caption{Operator interfaces and their intended roles.
Descriptions summarize the operation and its design scope, and task-specific feasibility safeguards are described below.}
\label{tab:nsga2_operator_meanings}
\begin{tabularx}{\textwidth}{
@{}l
>{\raggedright\arraybackslash}X@{}
}
\toprule
Operator name & Operation and design scope \\
\midrule

\texttt{tsp\_oropt}
& Relocate a contiguous tour segment by selecting its length,
original location, and reinsertion position while preserving
the order of nodes within the segment. The reference operator
moves one to three consecutive nodes.\\

\texttt{tsp\_2opt}
& Reverse a contiguous tour segment to change its boundary
connections and internal traversal direction. The design controls
segment selection and reordering while preserving a valid
city permutation. \\

\texttt{tsp\_3opt}
& Modify multiple tour segments through cut selection and
reconnection. The reference operator reverses two segments
defined by three cuts. \\

\midrule

\texttt{vrp\_oropt}
& Relocate a contiguous customer or node segment, choosing
which segment to move and where to insert it. The reference
operators move one to three nodes. \\

\texttt{vrp\_swap}
& Exchange two customer or node positions to modify visit
order while preserving node identifiers. The seed operator
uses route-quality sampling for CVRP and swaps signed
entries for SPDP. \\

\texttt{vrp\_2opt}
& Reverse a contiguous segment to modify route ordering
and neighboring connections. The seed operator acts within
one capacity-decoded route for CVRP and on the signed
node permutation for SPDP. \\

\midrule

\texttt{kp\_bitflip}
& Mutate binary item-selection decisions by choosing which
items to include or exclude. Weights, both objective profits,
and capacity information can guide the selection and extent
of these changes. \\

\texttt{kp\_crossover}
& Recombine binary item-selection vectors across individuals,
controlling pairing and inheritance of item choices.
The reference operator uses uniform crossover. \\

\texttt{kp\_repair\_mutation}
& Repair or locally improve an item subset through selective
item removal and insertion, balancing resource consumption
against objective profits. The design accounts for a single
capacity in KP or multiple resource capacities in MDKP. \\

\midrule

\texttt{operation\_crossover}
& Recombine operation-priority random keys across individuals
to alter the decoded operation sequence, preserving
$[0,1]$ bounds and leaving machine-option genes unchanged. \\

\texttt{operation\_mutation}
& Perturb operation-priority random keys to alter the decoded
sequence, controlling key selection and perturbation magnitude
while preserving $[0,1]$ bounds and leaving machine-option
genes unchanged. \\

\texttt{machine\_crossover}
& Recombine machine-option genes across individuals to alter
operation assignments, preserving eligible integer option
indices and leaving operation-priority keys unchanged. \\

\texttt{machine\_mutation}
& Change selected machine-option genes to eligible alternatives,
controlling which operations are reassigned and to which
machines while preserving valid option indices and leaving
operation-priority keys unchanged.\\

\bottomrule
\end{tabularx}
\end{table*}

\suppheading{Representation-specific constraints.}
For Bi-CVRP, operators receive customer demands, vehicle
capacity, and distances and modify customer permutations
without depot markers. The fixed decoder inserts depot
visits and splits the giant tour according to vehicle capacity.
For Bi-SPDP, operators modify signed node permutations:
absolute node identifiers specify the visit order, while
pickup node signs encode inclusion or omission.
For Bi-KP and Bi-MDKP, the same operator names act on binary
vectors, using a single capacity or multiple resource
capacities, respectively. A separate fixed feasibility repair
is applied after the operator sequence, independently of the
designed \texttt{kp\_repair\_mutation} implementation.

For TSP and FJSP, the representations and operator role decomposition follow E2OC~\cite{qiu2026e2oc}. More specifically, for TSP, the three operator interfaces,
\texttt{tsp\_oropt}, \texttt{tsp\_2opt}, and \texttt{tsp\_3opt},
act sequentially on the same offspring permutation matrix.
Their seed implementations perform segment relocation, segment
reversal, and multi-segment rearrangement, respectively.
Generated operators may redesign these transformations while
preserving the matrix shape and a complete city permutation
in each row.  For FJSP, the chromosome contains two blocks, namely continuous
operation-priority keys and discrete machine-option indices. The decoder selects the operation with the lowest priority key among the next unscheduled operations of all jobs
and schedules it subject to job precedence and machine
availability.
The four interfaces separate sequencing and assignment
decisions, with crossover and mutation provided for each block.

\begin{table}[t]
\centering
\small
\caption{NSGA-II population sizes and generation budgets.
``Evolution'' denotes candidate evaluation during harness self-evolution.
Inference with the selected harness doubles both evolution budgets,
whereas ``test'' denotes final evaluation of the selected operator combinations.}
\label{tab:nsga2_budgets}
\begin{tabular}{lrrrr}
\toprule
Problem
& Evolution pop.
& Evolution gen.
& Test pop.
& Test gen. \\
\midrule
Bi-KP    & 50 & 15 & 200 & 200 \\
Bi-MDKP  & 50 & 15 & 200 & 200 \\
Bi-CVRP  & 30 & 15 & 200 & 100 \\
Bi-SPDP  & 50 & 15 & 200 & 200 \\
Bi-TSP   & 50 & 15 & 200 & 200 \\
Tri-TSP  & 50 & 15 & 200 & 100 \\
Bi-FJSP  & 50 & 15 & 100 &  30 \\
Tri-FJSP & 50 & 15 & 100 &  30 \\
\bottomrule
\end{tabular}
\end{table}

\subsection{Why Two Backbones?}
\label{app:why_two_backbones}

SEMO and NSGA-II differ in both evolutionary control and the structure
of the operator design problem exposed to \hemo. SEMO performs an
incremental Pareto archive update after every candidate evaluation,
producing a tightly coupled loop between candidate generation and
acceptance. NSGA-II instead evaluates variation through a generational
population process with non-dominated sorting and diversity
preservation, introducing delayed interactions at the population level.

The two backbones also represent different levels of algorithmic
composition. Under SEMO, the AOD process primarily evolves a single problem-specific neighborhood
operator that determines how candidate solutions are generated from
the current archive. Under NSGA-II, the solver contains multiple
evolvable operators whose effects depend on the surrounding operator
configuration. The harness must therefore additionally decide which
component to improve and reason about interactions among operator
candidates.

These differences in solver structure and design granularity also
change the information available to the AOD harness. Under SEMO, the
harness can reason over a persistent non-dominated archive and the
direct consequence of individual neighborhood moves. Under NSGA-II, it
must additionally account for interactions at the population level, multiple
variation operators, and the delayed effect of environmental
selection. Evaluating \hemo under both settings tests whether a
stateful AOD harness can adapt effectively across both single-operator
and multi-operator design, as well as across substantially
different evolutionary control structures. This reduces the risk that
the learned harness behavior is specific to the idiosyncrasies of one
solver architecture.


\section{Algorithmic Details of HEMO}
\label{app:hemo_algorithms}

Algorithm~\ref{alg:hemo} in the main paper presents the outer harness
self-evolution procedure. This section provides the backbone-specific
execution contracts for SEMO and NSGA-II.

We distinguish bounded internal populations from external Pareto
archives. $\operatorname{Select}_K$ performs non-dominated sorting,
retains successive fronts up to capacity $K$, and uses descending
crowding distance to truncate the first front that does not fit.
In contrast, external Pareto archives retain all unique non-dominated
entries without capacity truncation.
Internal populations may therefore contain multiple dominance fronts.

The harness comprises four decision modules, with \texttt{retrieve\_memory} retrieving relevant search experience and \texttt{select\_target\_algorithm} selecting the operator component to improve, while \texttt{select\_parent\_algorithm} selects reference implementations and \texttt{evolve\_target\_algorithm} controls candidate evolution within a bounded evaluation budget.
Under SEMO, the target is a complete heuristic, so target operator selection is fixed and the remaining three modules are evolvable.
Under NSGA-II, the target is an individual operator within a combination, and all four modules are evolvable.

\setcounter{algorithm}{1} 

\begin{algorithm}[t]
\caption{SEMO Harness Execution}
\label{alg:semo_harness}
\small
\begin{algorithmic}[1]
\Require Seed heuristic operators $\mathcal{P}_{\mathrm{seed}}$;
memory filesystem $\mathcal{M}$; candidate budget $B_H$;
population capacity $K$
\Statex \textit{Notation:}
$\tau$ stores execution records,
$v$ indicates candidate validity,
and $\epsilon$ contains error information, if any.

\State $\mathcal{P} \gets \mathcal{P}_{\mathrm{seed}}$,
$\tau \gets [\,]$
\State Target algorithm is the core decision operator within
the SEMO backbone (e.g., \texttt{select\_neighbor})
\State \Return \Call{evolve\_target\_algorithm}{$B_H$}

\Statex
\Function{evolve\_target\_algorithm}{$B_H$}
\Comment{evolvable control policy}
    \State $e \gets 0$
    \While{$e < B_H$ and the evolution policy continues}
        \State $R \gets \Call{retrieve\_memory}{\mathcal{M}}$
        \Comment{evolvable}
        \State $Q \gets \Call{select\_parent\_algorithm}{\mathcal{P}}$
        \Comment{evolvable}
        \State \Call{ApplyHeartbeat}{$e,R,\tau,\mathcal{M}$}
        \State Generate heuristic $h$ with the Executor Agent
        using parents $Q$ and retrieved context $R$
        \State Evaluate $h$ to obtain score $\mathbf{s}(h)$,
        validity $v$, and error information $\epsilon$
        \State $\mathcal{P} \gets
        \operatorname{Select}_K(\mathcal{P}\cup\{h\})$
        \State Write $(h,Q,R,\mathbf{s}(h),v,\epsilon)$ to $\mathcal{M}$
        and append the execution record to $\tau$
        \State $e \gets e+1$
    \EndWhile
    \State $\mathcal{V} \gets
    \{h \in \mathcal{P} \mid h \text{ is valid}\}$
    \State \Return \Call{Finalize}{$\mathcal{V},\tau$}
\EndFunction
\end{algorithmic}
\end{algorithm}

Algorithm~\ref{alg:semo_harness} specifies the SEMO execution contract:
\texttt{evolve\_target\_algorithm} may adopt staged or adaptive search
policies within the candidate budget, while generation, evaluation,
population updates, and outcome recording use fixed helpers.
Every attempt is recorded, including the invalid attempts and failures. The heartbeat supplies periodic summaries or stagnation
feedback through the tools in Section~\ref{app:diagnostic_tool_prompts}.
Finalization excludes invalid or unscored candidates and uses a penalty
vector when no valid candidate remains.

\begin{algorithm}[t]
\caption{NSGA-II Harness Execution}
\label{alg:nsga2_harness}
\small
\begin{algorithmic}[1]
\Require Seed combinations $\mathcal{C}_{\mathrm{seed}}$;
memory filesystem $\mathcal{M}$;
per-harness candidate budget $B_H$;
inner budget $b$; capacities $K_C,K_O$
\Statex \textit{Notation:}
$\tau$ stores execution records,
$a$ denotes heartbeat feedback,
$v$ indicates candidate validity,
and $\epsilon$ contains error information, if any.

\Function{ExecuteHarness}{$B_H$}
\Comment{fixed orchestration}
    \State $\mathcal{C} \gets
    \operatorname{Select}_{K_C}(\mathcal{C}_{\mathrm{seed}})$
    \State Select an initial current combination $C$ from $\mathcal{C}$
    \State Initialize operator populations $\{\mathcal{O}_j\}$
    from $\mathcal{C}$
    \State $e \gets 0$, $t \gets 0$, $\tau \gets [\,]$
    \While{$e < B_H$}
        \State $R \gets$
        \Call{retrieve\_memory}{$\mathcal{M},t,e$}
        \Comment{evolvable}
        \State $j \gets$
        \Call{select\_target\_algorithm}{$R,t,e$}
        \Comment{evolvable}
        \State $a \gets$
        \Call{ApplyHeartbeat}{$t,j,R,\tau,\mathcal{M}$}
        \State $Q \gets$
        \Call{select\_parent\_algorithm}{$j,\mathcal{O}_j,R$}
        \Comment{evolvable}
        \State $b_t \gets \min(b,B_H-e)$
        \State Update selection statistics for target $j$
        \State $k \gets$
        \Call{evolve\_target\_algorithm}{$j,Q,R,a,b_t,e,\tau$}
        \Comment{evolvable}
        \State Require $1 \leq k \leq b_t$
        \State $e \gets e+k$, $t \gets t+1$
    \EndWhile
    \State \Return \Call{Finalize}{$\mathcal{C},C,\tau$}
\EndFunction

\Statex
\Procedure{EvaluateCandidate}{$o',j,Q,R,a,e'$}
\Comment{fixed transaction}
    \State $C' \gets C[j\leftarrow o']$
    \State Evaluate combination $C'$ to obtain score $\mathbf{s}(C')$,
    validity $v$, and error information $\epsilon$
    \State Assign $(\mathbf{s}(C'),v,\epsilon)$ to $C'$ and $o'$
    \State Store the accompanying operator identities with $o'$
    \State $\mathcal{C} \gets
    \operatorname{Select}_{K_C}(\mathcal{C}\cup\{C'\})$
    \State $\mathcal{O}_j \gets
    \operatorname{Select}_{K_O}(\mathcal{O}_j\cup\{o'\})$
    \If{\Call{AcceptAsCurrent}{$C',C,\mathcal{C},e'$}}
        \State $C \gets C'$
    \EndIf
    \State Write $(o',C',Q,R,a,\mathbf{s}(C'),v,\epsilon)$
    to $\mathcal{M}$ and append the execution record to $\tau$
\EndProcedure
\end{algorithmic}
\end{algorithm}

Under NSGA-II, Algorithm~\ref{alg:nsga2_harness} uses fixed
orchestration to allocate a bounded episode to
\texttt{evolve\_target\_algorithm}, which returns the number
of attempts consumed. Each attempt uses a fixed helper to
generate an operator with the Executor Agent, evaluate it,
and record the outcome, with the latter two steps detailed
in \textsc{EvaluateCandidate}. Only the selected operator
is replaced, and its score is measured in the resulting
complete combination, with the accompanying operator
identities retained as evaluation context.

The fixed rule for updating the current combination accepts a valid
candidate if it dominates the current combination or survives
population selection. Otherwise, acceptance depends on the mean relative score change,
with a decreasing temperature making the rule more selective
as the search progresses.
Finalization returns the valid retained combinations and their mean
score, with the valid current combination as a fallback.

\clearpage
\section{Prompts}
\label{app:user_prompts}

\providecommand{\HEMOprompttypeset}{%
  \normalfont
  \ifcsname ver@inconsolata.sty\endcsname
    \ttfamily
  \else
    \fontfamily{cmtt}\selectfont
  \fi
  \linespread{1}\fontsize{7}{8}\selectfont\raggedright
  \setlength{\parindent}{0pt}%
  \setlength{\parskip}{0pt}%
  \emergencystretch=3em\relax
}

We present the user prompt templates for harness design, harness code realization,
inner heuristic generation, and diagnostic tool calls. We use the prompts for SEMO to illustrate harness self-evolution and heuristic generation. Repeated context, detailed fixed-call protocols, API safeguards, and output boilerplate are explicitly marked where omitted.
Complete user prompts and system prompts, seed implementations and the corresponding prompts
for the NSGA-II backbone are provided in the supplementary materials.

The prompts for generating blueprints of retrieve memory, select parent algorithm, and evolve target algorithm module prepend the following shared context. It is reproduced once here and explicitly omitted from the three boxes.

\begin{promptbox}[Shared Harness Context]
\label{box:shared_harness_context}
\begingroup
\HEMOprompttypeset
You are designing a concise \textbf{JSON-like blueprint} for one replaceable Python method in an automatic heuristic design (AHD) control workflow. The workflow controls how heuristic candidates are retrieved, selected, generated, evaluated, archived, and finally returned.\par
The workflow maintains an archive of heuristic candidates and returns a set of valid heuristics. Each valid heuristic has a lower-is-better score vector. The harness is Pareto-aware: it should preserve and exploit trade-offs among heuristic candidates.\par
\par\vspace{2pt}
\# Available runtime fields inside the harness class:\par
- self.task\_id: str\par
- self.task\_descriptor: Dict[str, Any]. Use .get(...) access. Common available fields are: "problem\_name", "problem\_size", "representation", "family".\par
- self.memory\_fs: memory object; query it through self.memory\_fs.query(...)\par
- self.population: List[Heuristic]\par
- self.population\_size: int\par
\endgroup
\end{promptbox}

\subsection{Evolvable Harness-Module Prompts}
\label{app:harness_module_prompts}

The module-specific prompts guide the Proposer Agent in constructing a replacement blueprint for one decision module. The \texttt{retrieve\_memory} prompt specifies how to select compact empirical evidence, including complementary trade-offs and failure patterns, while \texttt{select\_target\_algorithm} selects which operator to evolve next in multi-operator MOEAs. The \texttt{select\_parent\_algorithm} prompt specifies how to choose diverse reference implementations, and \texttt{evolve\_target\_algorithm} defines how to organize candidate search within its budget. Each blueprint states the module's goal, signature, decision logic, and local hyperparameters. SEMO exposes a single heuristic target, making target selection trivial, so the three remaining module prompts are shown below.

\begin{promptbox}[Retrieve memory module prompt]
\label{box:retrieve_memory_prompt}
\begingroup
\HEMOprompttypeset

Design a concise JSON-like blueprint for the evolvable module retrieve\_memory(). The goal of this module is to choose which memory records are exposed to the inner heuristic generation LLM before generating the next heuristic candidate.\par
\textbf{Module name}: retrieve\_memory\par
\par\vspace{2pt}
\textbf{Expected output:}\par
- Return a List[Dict[str, Any]].\par
- Each dict should be a memory record retrieved from self.memory\_fs.\par
\par\vspace{2pt}
\textbf{Function helpers you can use:}\par
- self.memory\_fs.query(task\_id=..., query=..., top\_k=...)\par
\textit{[Additional helper details and supported memory-query fields are omitted for brevity.]}\par
\par\vspace{2pt}
\textbf{Recommended logic:}\par
- Prefer compact, relevant records that help candidate generation. Keep the number of returned records bounded. \par
- Prefer records that reveal different Pareto trade-off regions, e.g., archive\_improvement, efficient-but-lower-quality, high-quality-but-expensive.\par
- Avoid requiring analysis reports in every generation step; retrieve them only when they are useful for stagnation redirection.\par
- The retrieval logic may be adaptive rather than fixed: for example, it can change queries, tags, or top\_k according to search stage, Pareto-archive progress, recent failures, or stagnation, so that the memory context can include different evidence for exploitation, diversity, or redirection.\par
\par\vspace{2pt}
\textbf{Constraints:}\par
- Do not set top\_k above 10 in any self.memory\_fs.query(...) call.\par
- Must return a list, even if it is empty.\par
- Optional keyword arguments with default values are allowed, but every added argument must have a default value.\par
\par\vspace{2pt}
\textbf{Output format:}\par
Return only a compact JSON-like module blueprint:\par
\{"goal": "...", "func\_signature":"...", "logic": "...", "hyperparameters": "...",\}.\par
\endgroup
\end{promptbox}

\begin{promptbox}[Select parent algorithm module prompt]
\label{box:select_parents_prompt}
\begingroup
\HEMOprompttypeset

Design a concise JSON-like blueprint for the evolvable module select\_parent\_algorithm(). The goal of this module is to choose parent heuristics from the current heuristic archive/population for the next heuristic generation step.\par
\textbf{Module name}: select\_parent\_algorithm\par
\par\vspace{2pt}
\textbf{Expected output:}\par
- Return a List[Heuristic].\par
- The list may be empty if no valid parent exists.\par
\par\vspace{2pt}
\textbf{Function helpers you can use:}\par
- Standard Python list operations and random module functions such as
\texttt{random.sample(...)} and \texttt{random.choice(...)}.\par
\textit{[The optional selection helper is omitted for brevity.]}\par
\par\vspace{2pt}
\textbf{Recommended logic:}\par
- Filter valid heuristics: h.valid is True and h.score is not None\par
- Sample from a high-quality but diverse region of the archive (e.g. complementary Pareto regions).\par
- The number of returned parents only describes the generation setting:\par
\hspace*{1.28em}- two or more parents indicate a multi-parent setting;\par
\hspace*{1.28em}- one parent indicates a single-parent setting;\par
\hspace*{1.28em}- zero parents indicates open generation without parent code.\par
- The method may internally consider any number of candidates sorted, but the final returned parent list must contain at most 5 Heuristic objects.\par
\par\vspace{2pt}
\textbf{Constraints:}\par
- Must not directly modify archive storage.\par
- Optional keyword arguments with default values are allowed, but every added argument must have a default value.\par
\par\vspace{2pt}
\textbf{Output format:}\par
Return only a compact JSON-like module blueprint:\par
\{"goal": "...", "func\_signature":"...", "logic": "...", "hyperparameters": "",\}.\par
\endgroup
\end{promptbox}

\begin{promptbox}[Evolve target algorithm module prompt]
\label{box:run_prompt}
\begingroup
\HEMOprompttypeset

Design a concise JSON-like blueprint for the evolvable module evolve\_target\_algorithm(...). The goal of this module is to organize the inner AHD control loop. \par
\textbf{Module name}: evolve\_target\_algorithm\par
\par\vspace{2pt}
\textbf{Input:}\par
- budget: int, the maximum number of evaluated heuristics allowed.\par
\par\vspace{2pt}
\textbf{Expected output:}\par
Return the harness output only through the fixed finalization helper:\par
\hspace*{1.28em}return self.\_finalize\_harness\_output(trace)\par
\par\vspace{2pt}
Function helpers you can use: "self.retrieve\_memory()", "self.select\_parent\_algorithm()", "self.\_apply\_heartbeat(step, memory\_records, operator\_decision)", "self.\_generate\_evaluate\_record\_candidate(step, parents, memory\_records, operator\_decision, trace)", "self.\_finalize\_harness\_output(trace)"\par
\par\vspace{2pt}
\textbf{\# Fixed heartbeat protocol:}\par
Before \_generate\_evaluate\_record\_candidate(), evolve\_target\_algorithm() may use:\par
\par\vspace{2pt}
\hspace*{1.28em}operator\_decision = self.\_apply\_heartbeat(step=step,memory\_records=memory\_records,...)\par
\par\vspace{2pt}
which is responsible for failure analysis, periodic memory summary, and stagnation redirection.\par
\textit{[Detailed heartbeat-frequency constraints, fixed candidate-generation/evaluation/recording calls, and loop-safety constraints are omitted for brevity.]}\par
\par\vspace{2pt}
Hint: You are free to use novel self-improving optimization strategies, or design metaheuristic, hybrid, or unconventional optimization algorithms over the heuristic candidate population and selected parent heuristics, as long as they respect the fixed candidate-generation/evaluation interface and never exceed int(budget).\par
\par\vspace{2pt}
\textbf{Output format:}\par
Return only a compact JSON-like module blueprint:\par
\{"goal": "...", "func\_signature":"...", "logic": "...", "hyperparameters": "...", \}.\par
\endgroup
\end{promptbox}

\subsection{Harness Initialization and Code-Realization Prompts}
\label{app:harness_realization_prompts}

The initialization prompt requests a complete harness blueprint that coordinates the retrieve memory, select parent algorithm, and evolve target algorithm modules.
It encourages alternatives to the seed strategy while requiring concrete local hyperparameters and compatibility with the fixed runtime interfaces.
The placeholders \texttt{\_\_PROMPT\_FOR\_MODULE\_1\_\_}, \texttt{\_\_PROMPT\_FOR\_MODULE\_2\_\_}, and \texttt{\_\_PROMPT\_FOR\_MODULE\_3\_\_} are replaced at runtime by the complete module prompts, whose abridged versions appear above.

\begin{promptbox}[Initial Harness Population Prompt]
\label{box:initial_harness_population_prompt}
\begingroup
\HEMOprompttypeset
You are designing an executable automatic heuristic design (AHD) control workflow (i.e., harness) for multi-objective combinatorial optimization problems (MOCOPs). \par
Your task is to generate one concise harness blueprint covering THREE evolvable modules:\par
1. retrieve\_memory()\par
2. select\_parent\_algorithm()\par
3. evolve\_target\_algorithm()\par
\par\vspace{2pt}
\textbf{The blueprint should specify:}\par
- the intended memory retrieval strategy,\par
- the intended parent selection strategy,\par
- the intended evolution loop organization,\par
- any module-local numeric hyperparameters with concrete assigned values,\par
- how the evolution loop respects fixed heartbeat, candidate generation, evaluation recording, and finalization protocols.\par
\par\vspace{2pt}
Output a compact JSON-like blueprint string, not Python code.\par
\textbf{**The blueprint should preferably follow this structure**:}\par
\{\par
"retrieve\_memory": \{"goal": "...", "func\_signature":"...", "logic": "...", "hyperparameters": "...",\},\par
"select\_parent\_algorithm": \{"goal": "...", "func\_signature":"...",  "logic": "...", "hyperparameters": "...",\},\par
"evolve\_target\_algorithm": \{"goal": "...", "func\_signature":"...",  "logic": "...", "hyperparameters": "..."\}\par
\}\par
\par\vspace{2pt}
Field meanings: "goal": function of the module (should be a short sentence); \par
"logic" should describe the high-level logic of the module, like what information it retrieves, what choices it makes, and what it returns (should be less than 120 words). \par
"hyperparameters" should list concrete module-local parameter values; For example: "hyperparameters": \{"crossover\_prob": 0.6,\}\par
\par\vspace{2pt}
\textbf{\# Requirement for Module 1: retrieve\_memory()}\par
\_\_PROMPT\_FOR\_MODULE\_1\_\_\par
\par\vspace{2pt}
\textbf{\# Requirement for Module 2: select\_parent\_algorithm()}\par
\_\_PROMPT\_FOR\_MODULE\_2\_\_\par
\par\vspace{2pt}
\textbf{\# Requirement for Module 3: evolve\_target\_algorithm()}\par
\_\_PROMPT\_FOR\_MODULE\_3\_\_\par
\par\vspace{2pt}
Hint: You may design novel or traditional control strategies for the harness. Do not restrict yourself to the seed blueprint(template) or the example hyperparameters.\par
New local hyperparameters, exploration/exploitation strategies, parent-selection/retrieval policies, and scheduling/restart rules are allowed if they have concrete values and respect the fixed runtime fields and helper functions.\par
\par\vspace{2pt}
\textbf{Additional constraints:}\par
- Do not assume that memory\_records and parents are non-empty.\par
\endgroup
\end{promptbox}

The Realizer Agent converts blueprints into executable Python methods.
The complete harness prompt implements all evolvable modules during
initialization, whereas the single module prompt implements a proposed replacement using the current code and harness blueprint as context.
Both constrain edits to the designated modules, preserve compatible interfaces, and require explicit local parameter values, leaving the fixed infrastructure unchanged.

\begin{promptbox}[Full Harness Code Realization Prompt]
\label{box:realize_all_modules_prompt}
\begingroup
\HEMOprompttypeset
You are the harness code realization LLM. You will be given a harness blueprint (template) specifying the goal, function signatures, logic, and hyperparameters for each evolvable module.\par
Your task is to fill in the evolvable modules only. Do not rewrite unrelated fixed infrastructure.\par
\par\vspace{2pt}
The evolvable modules are:\par
1. retrieve\_memory\par
2. select\_parent\_algorithm\par
3. evolve\_target\_algorithm\par
\par\vspace{2pt}
The following helper functions are fixed and must not be renamed:\par
- self.\_apply\_heartbeat\par
- self.\_generate\_evaluate\_record\_candidate\par
- self.\_finalize\_harness\_output\par
\par\vspace{2pt}
\textbf{Important:}\par
- Preserve compatible method signatures: retrieve\_memory and select\_parent\_algorithm may add optional keyword arguments with defaults, but every added argument must have a default value.\par
- If the blueprint introduces module-local hyperparameters, define them inside the relevant method body with assigned values.\par
- Candidate heuristic score vectors are lower-is-better.\par
- Return three separate code blocks, one for each evolvable module.\par
\par\vspace{2pt}
Function helpers you can use inside each module:\par
\par\vspace{2pt}
\textit{[API signatures, supported memory-query fields, sampling safeguards, and selection helper requirements for retrieve\_memory and select\_parent\_algorithm are the same as those specified in the prompts for module-specific blueprints above and are omitted here for brevity.]}\par
\par\vspace{2pt}
- evolve\_target\_algorithm: "self.retrieve\_memory()", "self.select\_parent\_algorithm()", "self.\_apply\_heartbeat(...)", "self.\_generate\_evaluate\_record\_candidate(...)", "self.\_finalize\_harness\_output(...)"\par
\textit{[Fixed call signatures, heartbeat frequency requirements, candidate recording arguments, budget accounting, and loop safety constraints are the same as those specified in the blueprint prompt for evolve\_target\_algorithm above and are omitted here for brevity.]}\par
\par\vspace{2pt}
Seed harness blueprint(template):\par
\{seed\_harness\_template\}\par
\par\vspace{2pt}
Output format: return three separate Python code blocks implementing "retrieve\_memory(self, **kwargs)", "select\_parent\_algorithm(self, max\_returned\_parents=5, **kwargs)", and "evolve\_target\_algorithm(self, budget)".\par
\endgroup
\end{promptbox}

\begin{promptbox}[Single-Module Code Realization Prompt]
\label{box:realize_single_module_prompt}
\begingroup
\HEMOprompttypeset
You are the harness code realization LLM. You will be given:\par
1. the current implementation of one evolvable module,\par
2. the current blueprint(template) of the harness code\par
3. a blueprint specifying the replacement logic for this module.\par
Your task is to modify/improve the current module implementation according to the new logic and hyperparameters in "\# Module blueprint for replacement" with the goal of improving the search performance of the harness code.\par
\par\vspace{2pt}
Target module: \{module\_name\}\par
\par\vspace{2pt}
Current module implementation:\par
```python\par
\{current\_module\_code\}\par
```\par
\par\vspace{2pt}
Current Harness blueprint(template):\par
\{current\_harness\_template\}\par
\par\vspace{2pt}
\textbf{\# Module blueprint for replacement:}\par
\{module\_blueprint\}\par
\par\vspace{2pt}
\textbf{Requirements:}\par
- Preserve the exact method name and a compatible signature.\par
- Output exactly one Python code block.\par
- If a generated method needs an additional Python package or standard-library module, import it inside that method body.\par
- If the blueprint introduces module-local hyperparameters, define them inside the method body with assigned values.\par
\textit{[Module-specific memory-query restrictions, evolution loop constraints, and runtime safety requirements are omitted for brevity.]}\par
\par\vspace{2pt}
Output format (should be wrapped in ```python ```):\par
```python\par
def \{module\_name\}(self,...):\par
\hspace*{1.28em}...\par
```\par
\endgroup
\end{promptbox}
\clearpage

\subsection{Inner Heuristic Generation Prompt}
\label{app:heuristic_generation_prompt}

The inner generation prompt produces one candidate heuristic from the
task description and function template, selected parent implementations,
retrieved empirical records, and any diagnostic feedback. The harness supplies generation guidance, while the Executor Agent chooses a concrete design using this evidence. The prompt preserves the required interface
and feasibility constraints, discourages excessive runtime overhead, and
requests executable code alongside a short natural-language description for memory recording.

\begin{promptbox}[Heuristic Generation User Prompt]
\label{box:heuristic_generation_prompt}
\begingroup
\HEMOprompttypeset
You are an expert in LLM-based automatic heuristic design for multi-objective combinatorial optimization problems (MOCOP).\par
\textbf{\#\# Task}\par
\{task\_description\}\par
Your task is to generate ONE novel candidate heuristic function that improves the search behavior of a multi-objective COP solver.\par
\textbf{\#\# Here is an incomplete function template with the function signature and a doc-string that explains the inputs and outputs:}\par
```python\par
\{template\_program\}\par
```\par
\par\vspace{2pt}
\textbf{\#\# Parent heuristic(s) / reference solution(s)}\par
\{parents\_context\}\par
\par\vspace{2pt}
\textbf{\#\# Retrieved memory / empirical feedback}\par
\{memory\_context\}\par
\par\vspace{2pt}
\textbf{\#\# Additional analysis context}\par
\{feedback\}\par
\par\vspace{2pt}
\textbf{\#\# Function requirements:}\par
- keep the same function name and signature as the template;\par
- return an object/value with the same type and semantics required by the template;\par
- preserve all feasibility constraints described in the task description;\par
- avoid excessive runtime overhead;\par
- Important: if the function uses any package or standard-library module, write the necessary imports inside the function body.\par
\par\vspace{2pt}
\textbf{\#\# Output format}\par
Output exactly one Python code block followed by exactly one description block. \par
The Python code block must contain the complete function implementation with the same signature as the template:\par
```python\par
\textbf{\# complete function implementation}\par
```\par
\par\vspace{2pt}
After the code block, provide a concise description in this exact format:\par
\textless{}description\textgreater{}\par
A description within 80 words. Mention the selection rule, transformation/search logic, and any important parameter or threshold.\par
\textless{}/description\textgreater{}\par
\par\vspace{2pt}
Do not include any other text.\par
Do not include natural-language explanation, comments, or docstrings inside the code block.\par
\endgroup
\end{promptbox}

\clearpage
\subsection{Memory Summary and Stagnation Report Prompts}
\label{app:diagnostic_tool_prompts}

The heartbeat invokes two diagnostic tools to turn retrieved experience
into feedback for candidate generation. Each call writes the raw memory
records to a temporary file, whose path replaces \texttt{\{context\_path\}}.
Then the Executor Agent reads or searches that file using local tools and returns
a structured JSON report. The summary prompt consolidates useful
and weak patterns, unresolved questions, and recommendations. The
stagnation prompt identifies overused mechanisms and proposes a
structural change, with a concrete instruction for the next generation step. Reports are stored in memory, with each newly generated report replacing the previous active version of its type.

\begin{promptbox}[Memory Summary Tool Prompt]
\label{box:memory_summary_tool_prompt}
\begingroup
\HEMOprompttypeset
A temporary file contains the raw retrieved memory filesystem records for the current heuristic/operator-combination search.\par
First read or search this file, then synthesize actionable knowledge for the next heuristic/operator-combination generation step.\par
\par\vspace{2pt}
Temporary memory-record file:\par
\{context\_path\}\par
\par\vspace{2pt}
Return this JSON schema:\par
\{\par
\hspace*{0.8em}"analysis\_type": "memory\_summary",\par
\hspace*{0.8em}"summary": "one informative but concise synthesis",\par
\hspace*{0.8em}"useful\_patterns": ["promising patterns with evidence"],\par
\hspace*{0.8em}"weak\_or\_failed\_patterns": ["failed/risky/repeatedly weak patterns that cause a low hypervolume or a high runtime"],\par
\hspace*{0.8em}"open\_questions": ["uncertain points worth testing"],\par
\hspace*{0.8em}"recommendations": ["specific guidance for next generation"],\par
\hspace*{0.8em}"avoid": ["things the next heuristic/operator-combination should avoid"],\par
\hspace*{0.8em}"confidence": "low|medium|high"\par
\}\par
\endgroup
\end{promptbox}

\begin{promptbox}[Stagnation Report Tool Prompt]
\label{box:stagnation_report_tool_prompt}
\begingroup
\HEMOprompttypeset
A temporary file contains the raw retrieved memory filesystem records for a stagnated heuristic/operator search.\par
First read or search this file. Identify what has been overused and propose one substantially different direction.\par
Do not recommend minor parameter tuning unless it supports the structural pivot.\par
\par\vspace{2pt}
Temporary memory-record file:\par
\{context\_path\}\par
\par\vspace{2pt}
Return this JSON schema:\par
\{\par
\hspace*{0.8em}"analysis\_type": "stagnation\_pivot",\par
\hspace*{0.8em}"stagnation\_diagnosis": "why the search may be stuck",\par
\hspace*{0.8em}"overused\_patterns": ["ideas that appear repeatedly or dominate memory"],\par
\hspace*{0.8em}"underexplored\_directions": ["plausible high-EV directions not well tested"],\par
\hspace*{0.8em}"pivot\_recommendation": "one concrete structural pivot for the next candidate",\par
\hspace*{0.8em}"rationale": "why this pivot is worth trying",\par
\hspace*{0.8em}"avoid": ["local-basin behaviors to avoid"],\par
\hspace*{0.8em}"generation\_instruction": "concise instruction to pass to the heuristic/operator-generation LLM",\par
\hspace*{0.8em}"confidence": "low|medium|high"\par
\}\par
\endgroup
\end{promptbox}

\newpage

\begingroup
\definecolor{hemoHarnessBackground}{HTML}{FCF8EF}

\definecolor{hemoHarnessComment}{HTML}{549A46}
\definecolor{hemoHarnessKeyword}{HTML}{A83F50}
\definecolor{hemoHarnessString}{HTML}{996DBB}
\definecolor{hemoHarnessNumber}{HTML}{8A8175}
\lstdefinestyle{hemoGeneratedHarnessStyle}{
    language=Python,
    backgroundcolor=\color{hemoHarnessBackground},
    commentstyle=\color{hemoHarnessComment},
    keywordstyle=\color{hemoHarnessKeyword},
    stringstyle=\color{hemoHarnessString},
    numberstyle=\tiny\color{hemoHarnessNumber},
    basicstyle=\linespread{1}\fontsize{6.5}{7}\selectfont\ttfamily,
    breakatwhitespace=false,
    breaklines=true,
    captionpos=b,
    keepspaces=true,
    columns=fullflexible,
    showspaces=false,
    showstringspaces=false,
    showtabs=false,
    tabsize=2,
    numbers=none,
    numbersep=8pt,
    frame=single,
    rulecolor=\color{hemoHarnessString!45},
    mathescape=false,
    texcl=false,
    aboveskip=0.5em,
    belowskip=0.5em
}
\renewcommand{\lstlistingname}{Harness}

\section{Seed and Evolved Harness Modules}
\label{app:generated_harness_modules}
This section presents the seed harnesses and representative
LLM-generated harnesses for SEMO and NSGA-II. These examples
illustrate how HEMO evolves adaptive search procedures that
coordinate memory retrieval, parent selection, and candidate
generation using the search feedback.
The three modules of SEMO harness 141 and the four modules of
NSGA-II harness 98 originally span 416 and 1,241 source lines,
respectively. For readability, we present their core mechanisms with concise
module descriptions and summarize omitted implementation details
in comments. Captions report validation HV and runtime in seconds.

\begin{lstlisting}[style=hemoGeneratedHarnessStyle,caption={Seed SEMO harness (ID 0). Harness score vector on Bi-CVRP $(\mathrm{HV},\mathrm{runtime})=(0.316323,1.553284\,\mathrm{s})$. \texttt{retrieve\_memory} retrieves valid Pareto records for the current task. \texttt{select\_parent\_algorithm} samples one or two parents from candidates ranked by nondominated sorting and crowding distance. \texttt{evolve\_target\_algorithm} repeats retrieval, parent selection, generation, evaluation, and recording under a simple fixed loop.},label={lst:hemo_seed_semo}]
# seed retrieve_memory
"""Retrieve valid records matching the task, family, representation and objective dimensions, prioritizing Pareto candidates. The template also allows quality or runtime preferences and recent summary or stagnation reports when useful; the seed uses a fixed Pareto query."""
def retrieve_memory(self, **kwargs):
    memory_top_k = 5
    query = {'family': self.task_descriptor.get('family', self.task_descriptor.get('problem_family')),
        'representation': self.task_descriptor.get('representation'), 'tasks': [self.task_id],
            'objective_dims': [self.task_descriptor.get('objective_dim')], 'valid': True, 'on_pareto_pf': True}
    records = self.memory_fs.query(task_id=self.task_id, query=query, top_k=memory_top_k); return records

# seed select_parent_algorithm
"""Rank valid scored heuristics by nondomination and crowding, then sample one or two leading parents. Parent count sets the generation context without prescribing mutation or crossover."""
def select_parent_algorithm(self, max_returned_parents=5, **kwargs):
    crossover_prob = 0.6; parent_top_pool_min_size = 4
    valid = [h for h in self.population if h.valid and h.score is not None]
    if not valid: return []
    ordered = update_archive_by_crowding_distance(population=valid, max_size=len(valid), score_key='score')
    top_pool = ordered[:min(len(ordered), max(parent_top_pool_min_size, max_returned_parents))]
    if len(top_pool) >= 2 and random.random() < crossover_prob: num_parents = 2
    else: num_parents = 1
    num_parents = min(num_parents, max_returned_parents, len(top_pool)); return random.sample(top_pool, k=num_parents)

# seed evolve_target_algorithm
"""For each candidate, retrieve memory, select parents, apply heartbeat once, then generate, evaluate and record through fixed helpers. Finalize after the loop."""
def evolve_target_algorithm(self, budget):
    trace = []
    for step in range(int(budget)):
        memory_records = self.retrieve_memory(); parents = self.select_parent_algorithm()
        operator_decision = self._apply_heartbeat(step=step, memory_records=memory_records)
        candidate = self._generate_evaluate_record_candidate(...)
    return self._finalize_harness_output(trace)
\end{lstlisting}

\begin{lstlisting}[style=hemoGeneratedHarnessStyle,caption={Seed NSGA-II harness (ID 0). Harness score vector on Bi-TSP $(\mathrm{HV},\mathrm{runtime})=(0.458934,1.633936\,\mathrm{s})$. \texttt{retrieve\_memory} queries valid records about operators, combinations, and interactions. \texttt{select\_target\_algorithm} rotates deterministically through operator slots. \texttt{select\_parent\_algorithm} prioritizes implementations associated with combination archive improvements. \texttt{evolve\_target\_algorithm} iterates over pairs of distinct nondominated parents, followed by generation from individual parents, and refreshes the parent pool within the inner budget.},label={lst:hemo_seed_nsga2}]
# seed retrieve_memory
"""Query up to ten valid records matching task, family and representation, covering operators, combinations and interactions; return a list."""
def retrieve_memory(self, **kwargs):
    memory_top_k = 10
    query = {'family': self.task_descriptor.get('family', self.task_descriptor.get('problem_family')),
        'representation': self.task_descriptor.get('representation'), 'tasks': [self.task_id], 'valid': True,
            'text': ['operator', 'combination', 'interaction']}
    records = self.memory_fs.query(task_id=self.task_id, query=query, top_k=memory_top_k); return list(records or [])

# seed select_target_algorithm
"""Rotate deterministically through operator slots in their declared order, starting at the first slot. Return the chosen target with a reason."""
def select_target_algorithm(self, memory_records, **kwargs):
    if self.last_selected_operator not in self.operator_names: selected_idx = 0
    else:
        previous_idx = self.operator_names.index(self.last_selected_operator)
        selected_idx = (previous_idx + 1) % len(self.operator_names)
    target_operator = self.operator_names[selected_idx]
    return TargetOperatorDecision(target_operator=target_operator,
        reason='Deterministic coordinate rotation over all executable solver-operator slots.')

# seed select_parent_algorithm
"""Keep valid candidates for the target slot, favoring archive improvements. Include the current implementation as fallback and return at most six."""
def select_parent_algorithm(self, target_operator, memory_records, **kwargs):
    max_returned_parents = 6
    population = [candidate for candidate in self.operator_populations.get(target_operator,
        []) if candidate.valid and candidate.code and (candidate.operator_name == target_operator)]
    current_operator = self.current_combination.operators.get(target_operator)
    if current_operator is not None and all((candidate.id != current_operator.id for candidate in population)):
        population.append(current_operator)
    if not population: return []
    population.sort(key=lambda candidate: (not bool(candidate.meta.get('improved_combination_archive', False)),
        candidate.id))
    return population[:max_returned_parents]

# seed evolve_target_algorithm
"""Deduplicate valid nondominated parents by code and generate from each unique unordered pair. Then sample nondominated parents for individual offspring, using mutation_rate and remaining budget to limit their number. Refresh the parent pool and repeat within the inner budget."""
def evolve_target_algorithm(self, target_operator, operator_parents, memory_records, episode_context, inner_budget,
    start_eval_step, trace, **kwargs):
    mutation_rate = 0.2; outer_step = int(kwargs.get('outer_step', 0)); inner_budget = max(0, int(inner_budget))
    num_evaluated = 0; candidate_ids = []
    while num_evaluated < inner_budget:
        parents = self._get_nondominated_operator_parents(target_operator=target_operator,
            candidates=operator_parents)
        if not parents: parents = list(self.operator_populations.get(target_operator, []))
        unique_parents = []; seen_codes = set()
        for parent in parents:
            normalized_code = parent.code.strip()
            if not parent.valid or not normalized_code or normalized_code in seen_codes: continue
            seen_codes.add(normalized_code); unique_parents.append(parent)
        parent_pairs = list(itertools.combinations(unique_parents, 2)); random.shuffle(parent_pairs)
        for parent_a, parent_b in parent_pairs:
            if num_evaluated >= inner_budget: break
            pair = [parent_a, parent_b]
            candidate, _ = self._generate_evaluate_record_operator_candidate(...); candidate_ids.append(candidate.id)
            num_evaluated += 1
        if num_evaluated >= inner_budget: break
        mutation_pool = self._get_nondominated_operator_parents(target_operator=target_operator,
            candidates=self.operator_populations.get(target_operator, []))
        if not mutation_pool: mutation_pool = unique_parents
        if not mutation_pool: break
        num_mutations = max(1, int(math.ceil(mutation_rate * len(mutation_pool))))
        num_mutations = min(num_mutations, inner_budget - num_evaluated)
        selected_mutation_parents = random.sample(mutation_pool, k=min(num_mutations, len(mutation_pool)))
        for mutation_parent in selected_mutation_parents:
            if num_evaluated >= inner_budget: break
            candidate, _ = self._generate_evaluate_record_operator_candidate(...); candidate_ids.append(candidate.id)
            num_evaluated += 1
        operator_parents = list(self.operator_populations.get(target_operator, []))
    return {'target_operator': target_operator, 'num_evaluated': num_evaluated, 'candidate_ids': candidate_ids}
\end{lstlisting}

\begin{lstlisting}[style=hemoGeneratedHarnessStyle,caption={The LLM-generated SEMO harness with the best validation HV during self-evolution (ID 141). Bi-CVRP validation $(\mathrm{HV},\mathrm{runtime})=(0.429794,3.147883\,\mathrm{s})$. \texttt{retrieve\_memory} varies the context budget with search stage and stagnation, prioritizing records with high HV or low runtime and consulting additional evidence when space remains. \texttt{select\_parent\_algorithm} combines crowding tournaments with distance in normalized score space to choose complementary parents, increasing the probability of multiple parents during stagnation. \texttt{evolve\_target\_algorithm} switches among exploration, refinement, and exploitation using progress and recent outcomes, and adjusts heartbeat timing. Compared with the seed's fixed retrieval limit, parent sampling, and loop, these modules couple context selection and parent diversity to search feedback, providing explicit mechanisms to redirect stalled search.},label={lst:hemo_evolved_semo}]
# retrieve_memory
"""Match records by family, representation and objective dimensions. Set a small retrieval cap for exploration, refinement or exploitation, allowing one extra record under deep stagnation. Query the current task for high-HV Pareto records, then fast ones, followed by archive improvements from related tasks. Outside exploitation, consult sibling tasks while stagnation is moderate. If stalled and capacity remains, request pivot reports and error records, then summaries under deeper stagnation. Deduplicate IDs and stop as soon as the cap is filled, so later queries are conditional on remaining space."""
def retrieve_memory(self, stage='explore', stagnation=0, **kwargs):
    pf_explore = 2; pf_exploit = 1; impr_k = 2; sibling_k = 1; report_k = 1; fuzzy_k = 1; max_explore = 4
    max_refine = 3; max_exploit = 2; stag_threshold = 6; deep_stag = 10
    ...  # OMITTED: Read task descriptors; normalize task names; build valid base_filter and siblings.
    stage_l = str(stage).lower().strip(); is_exploit = stage_l.startswith('exploit')
    is_refine = stage_l.startswith('refine')
    if is_exploit: cap = max_exploit
    elif is_refine: cap = max_refine
    else: cap = max_explore
    if stagnation >= deep_stag: cap = min(cap + 1, 5)
    pf_k = pf_exploit if is_exploit else pf_explore
    if stagnation >= deep_stag: pf_k = 1
    records = []; seen_keys = set()
    ...  # OMITTED: Define _append: query safely, deduplicate IDs, stop at cap; return whether cap is full.
    if len(records) < cap:
        q1 = dict(base_filter); q1.update({'on_pareto_pf': True, 'high_hv': True, 'tasks': [self.task_id]})
        if _append(q1, pf_k): return records[:cap]
    if len(records) < cap:
        q2 = dict(base_filter); q2.update({'on_pareto_pf': True, 'low_runtime': True, 'tasks': [self.task_id]})
        if _append(q2, max(1, pf_k)): return records[:cap]
    if len(records) < cap:
        q3 = dict(base_filter); q3.update({'improved_archive': True, 'tasks': all_related_tasks})
        if _append(q3, impr_k): return records[:cap]
    if not is_exploit and siblings and (stagnation < deep_stag) and (len(records) < cap):
        q4 = {'family': family, 'valid': True, 'on_pareto_pf': True, 'tasks': siblings}
        if _append(q4, sibling_k): return records[:cap]
    ...  # OMITTED: If stalled outside exploitation and space remains, query pivot/error records, then summaries.
    return records[:cap]

# select_parent_algorithm
"""Filter valid finite scores, obtain the Pareto front and normalize each objective. Build a diverse pool using crowding and points from the middle of the front. As stagnation grows, raise the chance of multiple parents: choose the first by a crowding tournament and the second by maximum distance in normalized score space. Under deeper stagnation, add a third farthest from their centroid. Otherwise, sample a random valid heuristic or use a tournament among leading front members. Deduplicate and cap the returned parents."""
def select_parent_algorithm(self, max_returned_parents=5, crossover_prob=0.6, stagnation=0, diversity_boost=0.3,
    **kwargs):
    elite_pool = 6; restart_stag = 7; min_pool = 3; top_k = 3; third_parent_stag = 10; tournament_size = 2
    random_explore = 0.12; default_diversity_boost = 0.3; default_crossover_prob = 0.6
    ...  # OMITTED: Validate finite scores; obtain PF and crowding distances; build a diverse candidate_pool.
    ...  # OMITTED: Min-max normalize PF vectors into pf_norm; build pf_id_to_idx; define crowding tournament.
    try:
        scale = min(1.0, max(0.0, float(stagnation) / float(restart_stag))) if restart_stag > 0 else 0.0
    except Exception:
        scale = 0.0
    eff_cross = min(1.0, float(crossover_prob) + float(diversity_boost) * scale)
    limit = max(1, min(int(max_returned_parents), 5))
    choose_multi = random.random() < eff_cross and len(candidate_pool) >= 2
    if choose_multi:
        half = max(1, len(pf_sorted_by_cd) // 2); pool_set = {h.id for h in candidate_pool}
        high_half_pf = pf_sorted_by_cd[:half] if pf_sorted_by_cd else []
        high_half = [h for h in high_half_pf if h.id in pool_set] or list(candidate_pool)
        p1 = _crowding_weighted_tournament(high_half, pf_cd, tournament_size) or high_half[0]
        p1_idx_pf = pf_id_to_idx.get(p1.id, None); p1_vec = pf_norm[p1_idx_pf] if p1_idx_pf is not None else None
        best_idx = None; best_dist = -1.0
        if p1_vec is not None:
            for h in candidate_pool:
                if h.id == p1.id: continue
                idx_pf = pf_id_to_idx.get(h.id, None)
                if idx_pf is None: continue
                d = float(np.linalg.norm(pf_norm[idx_pf] - p1_vec))
                if d > best_dist: best_dist = d; best_idx = idx_pf
        parents = [p1]
        if best_idx is not None:
            p2 = pf[best_idx]
            if p2.id != p1.id: parents.append(p2)
        else:
            alt = next((h for h in candidate_pool if h.id != p1.id), None)
            if alt is not None: parents.append(alt)
        ...  # OMITTED: At stagnation >= third_parent_stag, add a third parent farthest from the first two's centroid.
        ...  # OMITTED: Deduplicate selected parent IDs into out and truncate to limit.
        return out
    if random.random() < float(random_explore) and valid_pop:
        parent = random.choice(valid_pop); return [parent][:limit]
    top_k_eff = max(1, min(int(top_k), len(pf_sorted_by_cd)))
    single_pool = pf_sorted_by_cd[:top_k_eff] if pf_sorted_by_cd else list(candidate_pool)
    if single_pool: parent = _crowding_weighted_tournament(single_pool, pf_cd, tournament_size) or single_pool[0]
    else: parent = random.choice(candidate_pool)
    return [parent][:limit]

# evolve_target_algorithm
"""Run the full candidate budget, using a cosine progress schedule moderated by stagnation to choose exploration, refinement or exploitation. Warmup, periodic injections and extended stagnation favor exploration. Sustained improvement promotes refinement; stalled archive updates move exploitation back to refinement, and a longer Pareto stall restores exploration. Favor exploitation near the end if progress is not stalled. Condition retrieval on stage and stagnation, and parent selection on stagnation. Apply heartbeat on periodic or stall triggers, subject to a minimum gap and a late cutoff. After each fixed generation/evaluation call, update improvement streaks and stall counters from archive membership, archive improvement and population growth; finalize the trace."""
def evolve_target_algorithm(self, budget):
    trace = []; B = int(budget) if budget is not None else 0; explore_frac = 0.22; exploit_frac = 0.56
    heartbeat_period = 8; min_hb_gap = 4; stag_threshold = 6; restart_stag = 8; stag_ref = 8; stag_boost = 0.5
    warmup = 3; cycle_len = 10; inject_frac = 0.1; hv_stall_thr = 5; pf_stall_thr = 6; lock_streak = 3
    late_hb_cutoff = 0.85; late_exploit_cut = 0.9; stagnation = 0; best_pf = len(self.population or []); last_hb = -99
    hv_stall = 0; pf_stall = 0; improve_streak = 0
    for step in range(B):
        prog = step / max(1, B)
        eff = 0.5 * (1.0 - math.cos(math.pi * prog)) * (1.0 - float(stag_boost) * min(1.0,
            float(stagnation) / float(stag_ref)))
        cyc = step % int(max(1, cycle_len)) / float(max(1, cycle_len))
        if stagnation >= int(restart_stag) or step < int(warmup) or cyc < float(inject_frac) or (eff < float(explore_frac)):
            stage = 'explore'
        elif eff < float(exploit_frac): stage = 'refine'
        else: stage = 'exploit'
        if improve_streak >= int(lock_streak) and stage == 'explore': stage = 'refine'
        if (hv_stall >= int(hv_stall_thr) or pf_stall >= int(pf_stall_thr)) and stage == 'exploit': stage = 'refine'
        if pf_stall >= int(pf_stall_thr) + 2 and stage == 'refine': stage = 'explore'
        if prog >= float(late_exploit_cut) and pf_stall < int(pf_stall_thr): stage = 'exploit'
        memory_records = self.retrieve_memory(stage=stage, stagnation=stagnation)
        parents = self.select_parent_algorithm(stagnation=stagnation)
        ...  # OMITTED: Apply heartbeat with periodic/stall triggers, minimum gap, and late-budget cutoff.
        self._generate_evaluate_record_candidate(...); last = trace[-1] if trace else {}
        hv = bool(last.get('improved_archive')); pf = bool(last.get('on_pareto_pf')); cur = len(self.population or [])
        grew = cur > best_pf; best_pf = max(best_pf, cur); improved = hv or pf or grew
        stagnation = 0 if improved else stagnation + 1; improve_streak = improve_streak + 1 if improved else 0
        hv_stall = 0 if hv else hv_stall + 1; pf_stall = 0 if pf or grew else pf_stall + 1
    return self._finalize_harness_output(trace)
\end{lstlisting}

\begin{lstlisting}[style=hemoGeneratedHarnessStyle,caption={The LLM-generated NSGA-II harness with the best validation HV during self-evolution (ID 98). Bi-TSP validation $(\mathrm{HV},\mathrm{runtime})=(0.468633,1.656327\,\mathrm{s})$. \texttt{retrieve\_memory} interleaves evidence about quality, runtime, operator interactions, and related tasks, adding failures and reports when indicated by search history. \texttt{select\_target\_algorithm} ranks operator slots using gains, failures, and exploration signals, matching past evidence to the other operators in the current combination. \texttt{select\_parent\_algorithm} balances quality, diversity, novelty, and compatibility with that combination. \texttt{evolve\_target\_algorithm} combines Beta sampling, observed rewards, and a UCB bonus to select generation modes, and updates parent credit from offspring outcomes. Compared with the seed's fixed rotation and pairing schedule, this design makes target and parent choices sensitive to operator interactions and adapts generation guidance to the observed outcomes.},label={lst:hemo_evolved_nsga2}]
# retrieve_memory
"""Read search progress, recent archive stagnation and failures. Retrieve TSP permutation records covering high HV, low runtime, archive improvements and general Pareto membership, plus relevant records from related tasks. Gather interaction evidence for recently selected operators, filling gaps with underused ones. Add failed candidates after failures or stagnation, and request pivot and summary reports when stalled or late in the search. Use a smaller total cap during stable progress. Interleave source groups and their records in turn, deduplicate IDs and stop at the cap; safely ignore unusable query results."""
def retrieve_memory(self, stagnation_window=3, **kwargs):
    pole_top_k = 4; interaction_top_k = 3; cross_task_top_k = 2; failure_top_k = 2; max_total = 10; compact_total = 7
    interaction_ops = 2; failure_lookback = 2; late_stage_fraction = 0.7
    ...  # OMITTED: Read recent history; compute stage, stagnant and recent_failure; define safe capped query.
    base = {'family': 'tsp', 'representation': 'permutation'}; pole_sources = []
    pole_sources.append(_safe_query({**base, 'valid': True, 'on_pareto_pf': True, 'high_hv': True}, pole_top_k))
    pole_sources.append(_safe_query({**base, 'valid': True, 'on_pareto_pf': True, 'low_runtime': True}, pole_top_k))
    pole_sources.append(_safe_query({**base, 'improved_archive': True}, pole_top_k))
    pole_sources.append(_safe_query({**base, 'valid': True, 'on_pareto_pf': True}, pole_top_k))
    ...  # OMITTED: Query related tasks; select recent or underused operators and query interaction evidence.
    failure_sources = []
    if recent_failure or stagnant: failure_sources.append(_safe_query({**base, 'valid': False}, failure_top_k))
    report_sources = []
    if stagnant or stage >= late_stage_fraction:
        report_sources.append(_safe_query({**base, 'include_reports': ['stagnation_pivot', 'memory_summary']}, 1))
    groups = [pole_sources, interaction_sources, cross_task_sources, failure_sources, report_sources]
    cap_total = compact_total if not stagnant and (not recent_failure) else max_total
    ...  # OMITTED: Interleave source groups round-robin, then records round-robin; deduplicate IDs up to cap_total.
    return deduped

# select_target_algorithm
"""Visit unseen operator slots first. For each target, compare the other operators in the current combination with contexts recorded in recent history and memory; retain sufficiently similar evidence and weight it by overlap. Estimate improvement and failure rates, expected gain, interaction strength, starvation, headroom and exploration bonuses. Combine these signals while penalizing plateaus and conflicts. Shift emphasis toward observed gains as the budget is consumed. During stagnation, favor improving targets without plateaus and strong interactions; also reward momentum from the latest episode. Penalize immediate repetition unless that target is strictly best. Return the highest score, breaking ties by fewer selections and then slot order, with an explanatory reason."""
def select_target_algorithm(self, memory_records, **kwargs):
    w_starv = 1.0; w_gain = 1.2; w_fail = 0.6; w_impr = 0.9; w_inter = 0.6; w_plateau = 0.85; w_ucb = 0.3
    w_head = 0.4; w_conflict = 0.65; repeat_penalty = 0.3; starv_boost = 1.35; stag_pivot_bonus = 0.5
    pivot_inter_bonus = 0.3; recent_gain_bonus = 0.45; exploit_bonus = 0.55; exploit_stage = 0.7; recent_window = 20
    plateau_window = 3; stagnation_window = 3; ctx_match_thresh = 0.66; conflict_thresh = 0.75; min_ctx_attempts = 3
    late_amplify = 0.6; shrink_prior = 2.0; remaining_budget = int(kwargs.get('remaining_budget', 0))
    total_budget = max(1, int(getattr(self, '_total_budget', 1)))
    progress = 1.0 - min(1.0, float(remaining_budget) / float(total_budget))
    amp_impr = 1.0 + late_amplify * max(0.0, min(1.0, progress))
    counts = {name: int(self.operator_selection_counts.get(name, 0)) for name in self.operator_names}
    if any((counts.get(n, 0) == 0 for n in self.operator_names)):
        best_unseen = min(self.operator_names, key=lambda n: (counts.get(n, 0), self.operator_names.index(n)))
        reason = f'Cold-start: {best_unseen}'
        return TargetOperatorDecision(target_operator=best_unseen, reason=reason)
    ...  # OMITTED: Detect stagnation from recent archive updates; extract context IDs and outcomes.
    ...  # OMITTED: For each target, retain evidence with non-target context overlap >= ctx_match_thresh; weight by
    # overlap.
    ...  # OMITTED: Compute shrink-corrected improvement/failure rates, positive gains, interaction scores, plateaus
    # and conflicts.
    ...  # OMITTED: Normalize gains; compute starvation/UCB terms and momentum from the last episode.
    scores = {}; headrooms = {}
    for name in self.operator_names:
        ...  # OMITTED: Read this target's rates, gain, starvation, UCB, plateau/conflict flags and headroom.
        score = 0.0; score += w_starv * (starv_boost if stagnant else 1.0) * starvation; score += w_gain * exp_gain
        score += w_impr * amp_impr * improvement_rate; score += w_inter * interaction_rate
        score += w_head * (1.0 - progress) * headroom; score += w_ucb * (1.0 - progress) * ucb_bonus
        score -= w_fail * failure_rate; score -= w_plateau * plateau; score -= w_conflict * conflict_flag
        if progress >= exploit_stage: score += exploit_bonus * exp_gain
        mom = last_momentum.get(name, 0.0)
        if mom > 0.0: score += recent_gain_bonus * mom
        scores[name] = score
    if stagnant:
        pivot_names = [n for n in self.operator_names if plateaus.get(n, 0.0) == 0.0 and per_op_hist_improve.get(n,
            0.0) + per_op_mem_improve.get(n, 0.0) > 0.0]
        for n in pivot_names: scores[n] = scores.get(n, 0.0) + float(stag_pivot_bonus)
        if interaction_rates:
            best_inter_op = max(self.operator_names, key=lambda n: interaction_rates.get(n, 0.0))
            scores[best_inter_op] = scores.get(best_inter_op, 0.0) + float(pivot_inter_bonus)
    last = self.last_selected_operator
    if isinstance(last, str) and last in scores:
        last_score = scores[last]
        strictly_best = all((other == last or last_score > scores[other] for other in scores))
        if not strictly_best: scores[last] = scores[last] - repeat_penalty
    def sort_key(n): return (scores.get(n, float('-inf')), -counts.get(n, 0), -self.operator_names.index(n))
    best_name = max(self.operator_names, key=sort_key)
    ...  # OMITTED: Build reason from the selected target's score components.
    return TargetOperatorDecision(target_operator=best_name, reason=reason)

# select_parent_algorithm
"""Filter valid operators for the target slot, retaining the current implementation as fallback. Score survivors using nondomination, archive improvements, compatibility with the current combination, code and context novelty, recency, crowding and objective extremes; penalize duplicate code. Under stagnation, reduce the elite fraction and strengthen novelty and diversity. Keep the current operator, leading candidates and available crowding/objective extremes. Fill remaining slots by alternating quality order with compatible candidates whose code and context are not already selected. Deduplicate IDs, enforce the target slot and cap the result."""
def select_parent_algorithm(self, target_operator, memory_records, **kwargs):
    max_returned_parents = 6; elite_fraction = 0.5; explore_min_fraction = 0.34; nd_bonus = 1.0
    context_match_bonus = 0.7; archive_improve_bonus = 0.5; dup_penalty = 0.15; novelty_weight = 0.3
    recency_bonus = 0.2; crowd_bonus = 0.4; pole_bonus = 0.45; stagnation_window = 3; explore_boost = 1.4
    force_current_slot = True
    ...  # OMITTED: Filter valid target operators; retain current implementation as fallback.
    ...  # OMITTED: Compute normalized scores, nondominated IDs, crowding, objective poles, code frequencies and
    # recency.
    ...  # OMITTED: Count code/context occurrences in memory; define context_match_ratio and candidate_ctx_sig.
    ...  # OMITTED: Compute stagnant from recent archive updates.
    eff_elite_fraction = explore_min_fraction if stagnant else elite_fraction
    if stagnant:
        novelty_weight = float(novelty_weight) * float(explore_boost)
        crowd_bonus = float(crowd_bonus) * float(explore_boost)
    def quality(cand):
        q = 0.0; cid = getattr(cand, 'id', None)
        if cid in nd_set: q += nd_bonus
        meta = getattr(cand, 'meta', {}) or {}
        if isinstance(meta, dict) and meta.get('improved_combination_archive'): q += archive_improve_bonus
        q += context_match_bonus * context_match_ratio(cand)
        dup = max(0, code_freq.get(getattr(cand, 'code', '') or '', 1) - 1); q -= dup_penalty * dup
        code = getattr(cand, 'code', '') or ''; sig = candidate_ctx_sig(cand); mem_c = mem_code_counts.get(code, 0)
        mem_s = mem_ctx_counts.get(sig, 0); novelty = 1.0 / (1.0 + float(mem_c + mem_s))
        q += novelty_weight * novelty; q += recency_bonus * recency_fresh(cand)
        q += crowd_bonus * float(dist_norm.get(getattr(cand, 'id', ''), 0.0))
        if cid in pole_ids: q += pole_bonus
        return q
    ranked = sorted(survivors, key=lambda c: (quality(c), -float(score_sum(c))), reverse=True)
    ...  # OMITTED: Keep current operator, top quality fraction, and crowding/objective poles while capacity remains.
    remaining = [c for c in ranked if getattr(c, 'id', None) not in selected_ids]
    explore_sorted = sorted(remaining, key=lambda c: (context_match_ratio(c), quality(c)), reverse=True)
    exploit_idx = 0; explore_idx = 0; take_explore = True
    while len(selected) < max_returned_parents and (exploit_idx < len(remaining) or explore_idx < len(explore_sorted)):
        picked = None
        if take_explore:
            ...  # OMITTED: Advance explore_sorted; skip selected IDs and already used code/context signatures; set
            # picked.
        else:
            ...  # OMITTED: Advance remaining by quality; skip selected IDs; set picked.
        if picked is None: take_explore = not take_explore; continue
        pid = getattr(picked, 'id', None); selected.append(picked)
        ...  # OMITTED: Record the selected ID, code and context signature.
        take_explore = not take_explore
    ...  # OMITTED: Fill unused slots by quality; filter target IDs, deduplicate and cap final_list.
    return final_list

# evolve_target_algorithm
"""Track Beta success estimates, observed rewards and counts for exploration, exploitation, recombination, repair, annealing, simplification and rebalancing guidance. Force repair after repeated non-improvement and exploration early in the budget; otherwise combine a Beta sample, mean reward and UCB bonus, with a late bias toward exploitation. Require multiple parents for recombination, simplification and rebalancing. Add the chosen instruction, best-score hint and any weakest-objective focus to the LLM prompt, then generate and evaluate. Update mode statistics, decay parent credits and credit parents of improving offspring. Refresh a diverse pool ranked by validity, improvement, contribution and score; update the best candidate and failure streak."""
def evolve_target_algorithm(self, target_operator, operator_parents, memory_records, episode_context, inner_budget,
    start_eval_step, trace, **kwargs):
    max_parent_pool = 6; elite_keep = 2; explore_frac = 0.3; repair_after = 2; c_ucb = 1.2; ts_weight = 0.6
    exploit_bias_late = 0.5; contribution_decay = 0.8; credit_gain = 0.5; improved_base = 1.0; small_valid_bonus = 0.1
    rebalance_worst_obj = True
    ...  # OMITTED: Read outer_step; define score/validity/context helpers and initialize parent credits.
    current_parents = list(operator_parents or []); candidate_ids = []; best_sum = float('inf'); best_cand = None
    best_vec = None; fail_streak = 0
    modes = ['explore', 'exploit', 'recombine', 'repair', 'anneal', 'simplify', 'rebalance']
    mode_beta = {m: {'alpha': 1.0, 'beta': 1.0} for m in modes}
    mode_stats = {m: {'reward_sum': 0.0, 'count': 0.0} for m in modes}; parent_credits = {}
    ...  # OMITTED: maintain_pool ranks by validity, improvement, credit, then score; retains elites and diverse
    # parents.
    total_iters = int(inner_budget)
    for inner_step in range(total_iters):
        ...  # OMITTED: Ensure every current parent has a credit entry.
        progress = (inner_step + 1) / float(max(1, total_iters))
        if fail_streak >= int(repair_after): mode = 'repair'
        elif progress <= float(explore_frac): mode = 'explore'
        else:
            eligible_modes = ['explore', 'exploit', 'anneal']
            if len(current_parents) >= 2: eligible_modes.extend(['recombine', 'simplify', 'rebalance'])
            total_plays = sum((int(v.get('count', 0.0)) for v in mode_stats.values())); best_val = float('-inf')
            mode = 'explore'
            for m in eligible_modes:
                b = mode_beta.get(m, {'alpha': 1.0, 'beta': 1.0}); a = float(b.get('alpha', 1.0))
                bb = float(b.get('beta', 1.0))
                try:
                    ts_sample = random.betavariate(max(1e-06, a), max(1e-06, bb))
                except Exception:
                    ts_sample = a / (a + bb) if a + bb > 0 else 0.5
                stats = mode_stats.get(m, {'reward_sum': 0.0, 'count': 0.0}); count = float(stats.get('count', 0.0))
                reward_sum = float(stats.get('reward_sum', 0.0)); mean_r = reward_sum / count if count > 0 else 0.0
                bonus = c_ucb * math.sqrt(max(0.0, math.log(float(total_plays) + 1.0)) / (count + 1.0))
                bias = exploit_bias_late * max(0.0, min(1.0, progress)) if m == 'exploit' else 0.0
                val = float(ts_weight) * float(ts_sample) + float(mean_r) + float(bonus) + float(bias)
                if val > best_val: best_val = val; mode = m
        ...  # OMITTED: Build LLM guidance: mode instruction, best-score hint, and weakest-objective focus for
        # rebalance.
        candidate, update_out = self._generate_evaluate_record_operator_candidate(...)
        candidate_ids.append(getattr(candidate, 'id', None))
        ...  # OMITTED: Compute score sum, improvement flag, and nonnegative gain delta against best_sum.
        if is_improved: reward = float(improved_base) + float(credit_gain) * float(delta)
        elif bool(getattr(candidate, 'valid', False)) and math.isfinite(ssum): reward = float(small_valid_bonus)
        else: reward = 0.0
        mb = mode_beta.get(mode, {'alpha': 1.0, 'beta': 1.0})
        if is_improved: mb['alpha'] = float(mb.get('alpha', 1.0)) + 1.0
        else: mb['beta'] = float(mb.get('beta', 1.0)) + 1.0
        mode_beta[mode] = mb
        mode_stats[mode]['reward_sum'] = float(mode_stats[mode].get('reward_sum', 0.0)) + float(reward)
        mode_stats[mode]['count'] = float(mode_stats[mode].get('count', 0.0)) + 1.0
        if bool(getattr(candidate, 'valid', False)) and math.isfinite(ssum):
            if ssum < best_sum: best_sum = ssum; best_cand = candidate; best_vec = score_vector(candidate)
        for k in list(parent_credits.keys()): parent_credits[k] = float(parent_credits[k]) * float(contribution_decay)
        if is_improved and current_parents:
            inc = float(reward) / float(max(1, len(current_parents)))
            for p in current_parents:
                pid = getattr(p, 'id', None)
                if isinstance(pid, str): parent_credits[pid] = parent_credits.get(pid, 0.0) + float(inc)
        updated_pool = list(current_parents) + [candidate]; current_parents = maintain_pool(updated_pool)
        ...  # OMITTED: Initialize credit entries for newly retained parents.
        fail_streak = 0 if is_improved else fail_streak + 1
    return {'target_operator': target_operator, 'num_evaluated': int(inner_budget), 'candidate_ids': candidate_ids}
\end{lstlisting}

\endgroup

\newpage


\section{Why Discounted Credit is Suitable for Stateful Harness Self-Evolution}
\label{app:discounted_credit}

\hemo assigns credit to the decision module updated at each outer step according to the downstream multi-objective performance of the resulting harness state. The improvement from parent to child is normalized and clipped to $[-1,1]$ for each score dimension, and the module reward is their average. Hence, $r_t\in[-1,1]$. The expected utility of a module depends on the current harness state:
$$
    \mu_{m,t}
    =
    \mathbb{E}
    \left[
        r_t
        \mid
        m_t=m,H_t
    \right].
$$
As other decision modules change, the expected benefit of updating module $m$ may also change. Module selection in \hemo is therefore naturally modeled as a nonstationary credit assignment problem.

\subsection{Attenuation of Stale Module Credit}
\label{app:discounted_stale_credit}

For each module $m$, \hemo maintains a discounted selection count and reward sum,
$$
    n^{(\gamma)}_{m,t}
    =
    \gamma n^{(\gamma)}_{m,t-1}
    +
    \mathbf{1}\{m_t=m\},
    \qquad
    R^{(\gamma)}_{m,t}
    =
    \gamma R^{(\gamma)}_{m,t-1}
    +
    \mathbf{1}\{m_t=m\}r_t,
$$
where $\gamma\in(0,1)$. Equivalently, an observation collected at outer step $s$ contributes with weight $\gamma^{t-s}$ at step $t$.

\suppheading{Proposition 1 (Geometric attenuation of stale credit).}
Consider an outer step $\tau$ after which the harness enters a different search regime. Let
$$
W^{-}_{m,t}
=
\sum_{s\leq\tau}
\mathbf{1}\{m_s=m\}\gamma^{t-s}.
$$
denote the total weight at step $t>\tau$ assigned to observations of module $m$ collected no later than $\tau$. Then
\begin{equation}
    W^{-}_{m,t}
    =
    \gamma^{t-\tau}W^{-}_{m,\tau}.
    \label{eq:stale_weight_decay}
\end{equation}

\textit{Proof.}
Factoring $\gamma^{t-\tau}$ from the definition gives
$$
W^{-}_{m,t}
=
\gamma^{t-\tau}
\sum_{s\leq\tau}
\mathbf{1}\{m_s=m\}\gamma^{\tau-s}
=
\gamma^{t-\tau}W^{-}_{m,\tau}.
$$
Since $\gamma\in(0,1)$, credit collected under earlier harness states loses weight geometrically as the outer trajectory proceeds. \hfill$\square$

This attenuation does not require module $m$ to be selected again. By comparison, an undiscounted estimator retains unit weight for every historical observation until additional observations of the same module dilute their influence.

To illustrate the effect on module utility, consider a local
model in which the mean reward is $\mu_m^-$ up to $\tau$
and $\mu_m^+$ afterward.
Let $W^+_{m,t}$ denote the discounted weight of observations
collected after $\tau$. For $W^-_{m,t}+W^+_{m,t}>0$,
define the weighted average of these means as
$$
\widetilde{\mu}_{m,t}
=
\frac{W^-_{m,t}\mu_m^-+W^+_{m,t}\mu_m^+}
{W^-_{m,t}+W^+_{m,t}}.
$$
Then
$$
\left|\widetilde{\mu}_{m,t}-\mu_m^+\right|
=
\frac{W^-_{m,t}}{W^-_{m,t}+W^+_{m,t}}
\left|\mu_m^--\mu_m^+\right|.
$$
Together with Eq.~\eqref{eq:stale_weight_decay},
this identity shows that, as new observations of module $m$
accumulate after the change, the relative contribution of
earlier observations to this weighted average decreases.

\subsection{Recovery of Exploration Pressure}
\label{app:discounted_exploration}

Discounting also affects modules that have not been selected recently. Let
$$
    N_t
    =
    \sum_{j\in\mathcal{M}}
    n^{(\gamma)}_{j,t},
    \qquad
    b_{m,t}
    =
    c
    \sqrt{
        \frac{\log(1+N_t)}
             {n^{(\gamma)}_{m,t}}
    }
$$
denote the total discounted count and the UCB exploration bonus, respectively.

\suppheading{Proposition 2 (Exploration recovery for an unselected module).}
Suppose module $m$ has previously been selected but is not selected during the following $\ell$ outer steps. Then
$$
    n^{(\gamma)}_{m,t+\ell}
    =
    \gamma^\ell n^{(\gamma)}_{m,t},
    \qquad
    \bar r^{(\gamma)}_{m,t+\ell}
    =
    \bar r^{(\gamma)}_{m,t}.
$$
Moreover, because exactly one module is selected at every outer step, $N_{t+\ell}\geq1$, giving
\begin{equation}
    b_{m,t+\ell}
    \geq
    c
    \sqrt{
        \frac{\log 2}
             {\gamma^\ell n^{(\gamma)}_{m,t}}
    }.
    \label{eq:exploration_recovery}
\end{equation}

\textit{Proof.}
When $m$ is not selected, its discounted count and reward sum are both multiplied by $\gamma$ at every outer step. After $\ell$ steps,
$$
    n^{(\gamma)}_{m,t+\ell}
    =
    \gamma^\ell n^{(\gamma)}_{m,t},
    \qquad
    R^{(\gamma)}_{m,t+\ell}
    =
    \gamma^\ell R^{(\gamma)}_{m,t}.
$$
Their ratio is unchanged, while substituting the discounted count and $N_{t+\ell}\geq1$ into the exploration bonus gives Eq.~\eqref{eq:exploration_recovery}. For $0<\gamma<1$, this lower bound for the bonus at time $t+\ell$ increases with $\ell$
at a rate proportional to $\gamma^{-\ell/2}$.
Thus, the longer module $m$ remains unselected, the larger its guaranteed exploration bonus becomes, increasing its opportunity to be selected again even without an
improvement in its estimated reward. \hfill$\square$

\suppheading{Key Takeaway.}
Discounting provides two properties that are useful for stateful harness self-evolution. Credit collected under earlier harness states is attenuated geometrically, and modules that have not been selected recently receive
larger exploration bonuses as their discounted counts decrease,
making them more likely to be selected again. Together, these effects allow the scheduler to respond when the utility of a module changes after other parts of the harness have been updated.

These results do not imply that discounted UCB universally dominates standard UCB. In a stationary setting, retaining the complete history can provide a more accurate estimate of a fixed module utility. In \hemo, however, module utility depends on the evolving harness state, so discounting exchanges some statistical efficiency under stationarity for greater responsiveness to changing module utility. This motivation is consistent with established analyses of discounted UCB for nonstationary bandit problems~\cite{garivier2011upper}.

\section{Experimental Details}
\label{app:experimental_details}

\subsection{Memory Filesystem and Record Schema}
\label{app:memory_filesystem}

\hemo maintains a filesystem-based memory that persists optimization experience across successive harness states. Fixed evaluation API and memory logging routines ensure that different harness states produce records under a common schema. Evaluated candidates are stored as structured candidate records following the \texttt{MemoryRecord} schema, while diagnostic summaries and stagnation reports are maintained as distinct report records within the same memory filesystem.  

Table~\ref{tab:memory_record_schema} summarizes the fields most relevant to experience retrieval and the memory context exposed to subsequent AOD steps. The schema combines task information, evaluation outcomes, search lineage, and multi-objective status. In particular, \texttt{improved\_archive}, \texttt{on\_pareto\_pf}, \texttt{high\_hv}, and \texttt{low\_runtime} provide compact descriptors of a candidate's contribution and current position in the candidate archive. The Pareto related annotations are updated as the active archive changes, while the original score vector remains available for detailed comparison. These Pareto annotations expose whether a candidate contributes to the current nondominated archive and which region of the quality--efficiency trade-off it represents. Such information is useful because multi-objective archives are designed not only to retain high-quality solutions, but also to preserve convergence and diversity across competing objectives~\cite{laumanns2002combining,knowles2003properties}. In \hemo, these archive statistics are retained as explicit memory attributes so that later search decisions can access this information directly.  

For backbones with multiple interacting operators, the memory additionally records the target operator through \texttt{target\_operator} and the operator composition mainly through \texttt{operator\_constitution} and \texttt{combination\_context}. This preserves the context in which an operator produced its observed performance, allowing subsequent search decisions to reason about interactions among algorithmic components instead of treating each operator candidate in isolation. Such context is important because alternative designs of the same operator can exhibit different strengths and inheritance relationships, while operators with different functional roles may interact through complementary, conflicting, or mutually exclusive effects~\cite{qiu2026e2oc}. Prior work on multi-operator evolutionary search similarly shows that combining operators with complementary behaviors and coordinating their usage can be critical to overall search performance~\cite{chen2025reinforced,li2024evolutionary}.

\begin{table}[t]
\centering
\small
\caption{Memory fields most relevant to experience retrieval and subsequent search decisions.}
\label{tab:memory_record_schema}
\begin{tabular}{p{0.16\linewidth} p{0.27\linewidth} p{0.49\linewidth}}
\toprule
\textbf{Role} & \textbf{Memory field} & \textbf{Meaning and type} \\
\midrule

Retrieval scope &
\texttt{task\_id} &
Task associated with the record (\texttt{str}), used to restrict retrieval to the current task or related scope. \\

&
\texttt{family}, \texttt{representation} &
Problem family and solution representation (\texttt{Optional[str]}), used for matching related search experience. \\

&
\texttt{objective\_dims} &
Objective dimensionalities represented by the record (\texttt{List[str]}, \texttt{List[int]}). \\

&
\texttt{timestamp} &
Record creation time (\texttt{float}), providing a weak recency preference during retrieval. \\

\midrule
\makecell[l]{Evaluation\\context} &
\texttt{score} &
Observed candidate score vector (\texttt{Any}, normally a numeric list), typically containing negative HV and per-instance runtime. \\

&
\texttt{valid}, \texttt{error} &
Evaluation validity and failure information (\texttt{Optional[bool]}, \texttt{Optional[str]}). \\

&
\texttt{improved\_archive} &
Whether the candidate improved the candidate archive when first evaluated (\texttt{Optional[bool]}). \\

&
\texttt{on\_pareto\_pf} &
Whether the candidate or operator combination currently belongs to the active Pareto archive (\texttt{bool}). \\

&
\texttt{high\_hv}, \texttt{low\_runtime} &
Indicators of relatively strong HV or runtime among active Pareto records (\texttt{bool}). Whether the record is above the median in HV or below the median in runtime among active candidate Pareto records, respectively. \\\\

\midrule
Search context &
\texttt{parent\_ids}, \texttt{used\_memory\_ids} &
Parent candidates and historical records used during generation (\texttt{List[str]}). \\

&
\texttt{description}, \texttt{summary} &
Textual description of the candidate and a compact summary of its search outcome (\texttt{str}). \\


\midrule
\makecell[l]{Operator\\interaction} &
\texttt{target\_operator} &
Operator selected for modification in a multi-operator solver (\texttt{Optional[str]}). \\

&
\texttt{context\_operator\_ids} &
Identifiers of the accompanying operators in the evaluated composition (\texttt{Dict[str, str]}). \\

&
\texttt{operator\_constitution} &
Complete operator composition associated with the observed evaluation result (\texttt{Dict[str, Any]}). \\

&
\texttt{combination\_context} &
Textual description of the operator combination and its search context (\texttt{str}). \\

\bottomrule
\end{tabular}
\end{table}


\subsection{Harness Self-Evolution and Inference Protocols for \hemo and AOD Baselines}
\label{app:training_inference_protocols}

\suppheading{Design instances.}
Table~\ref{tab:aad_design_instances} summarizes the problem sizes and numbers of instances used when evaluating generated heuristics or operator combinations during algorithm design. For source tasks, these instances are used during harness self-evolution. For the corresponding transfer tasks, they are used only after the trained harness state has been fixed for inference. Under the SEMO backbone, the instance settings and evaluation protocol for TSP, Bi-CVRP, and Bi-KP follow MPaGE~\cite{hieu2026pareto}. Under NSGA-II, the benchmark construction and design protocol for FJSP follow E2OC~\cite{qiu2026e2oc}. The remaining tasks use the same evaluation pipeline with the problem-specific configurations described in Appendix~\ref{app:benchmark_problems}.

\begin{table}[t]
\centering
\small
\setlength{\tabcolsep}{3.5pt}
\renewcommand{\arraystretch}{1.08}
\caption{Problem sizes and numbers of instances used for heuristic or operator design.}
\label{tab:aad_design_instances}

\begin{tabular}{
    @{}
    m{0.13\linewidth}
    m{0.10\linewidth}
    >{\centering\arraybackslash}m{0.20\linewidth}
    >{\centering\arraybackslash}m{0.09\linewidth}
    >{\centering\arraybackslash}m{0.23\linewidth}
    >{\centering\arraybackslash}m{0.09\linewidth}
    @{}
}
\toprule
& &
\multicolumn{2}{c}{\textbf{SEMO}} &
\multicolumn{2}{c}{\textbf{NSGA-II}} \\
\cmidrule(lr){3-4}
\cmidrule(lr){5-6}

\textbf{Family} &
\textbf{Task} &
\textbf{Design size} &
\textbf{Instances} &
\textbf{Design size} &
\textbf{Instances} \\
\midrule

\multirow{2}{*}{TSP}
& Bi-TSP
& 20 cities
& 10
& 20 cities
& 10 \\

& Tri-TSP
& 20 cities
& 10
& 20 cities
& 10 \\
\midrule

\multirow{2}{*}{Routing}
& Bi-CVRP
& 50 customers
& 10
& 50 customers
& 10 \\

& Bi-SPDP
& \makecell[c]{20 nodes incl.\ depot}
& 20
& \makecell[c]{20 nodes incl.\ depot}
& 20 \\
\midrule

\multirow{2}{*}{Knapsack}
& Bi-KP
& 50 items
& 10
& 50 items
& 10 \\

& Bi-MDKP
& 50 items
& 10
& 50 items
& 10 \\
\midrule

\multirow{2}{*}{Scheduling}
& Bi-FJSP
& \makecell[c]{20 jobs;\\5, 10, or 15 machines}
& 15
& \makecell[c]{Brandimarte Mk15;\\30 jobs, 15 machines}
& 1 \\

& Tri-FJSP
& \makecell[c]{20 jobs;\\5, 10, or 15 machines}
& 15
& \makecell[c]{Brandimarte Mk15;\\30 jobs, 15 machines}
& 1 \\

\bottomrule
\end{tabular}
\end{table}

\suppheading{\hemo self-evolution and inference settings.}
For both backbones, the initial harness population contains five harness states. The discounted UCB scheduler uses discount factor $\gamma=0.98$ and exploration coefficient $c=\sqrt{2}$. During inference, the candidate population size is 20 for SEMO, and, for NSGA-II, the population size of each operator type and the complete operator combination population size are both set to 20. For same-task inference, up to the latest 1,000 memory records collected during harness self-evolution are loaded to warm-start the memory filesystem. In intra-family cross-task inference, no source-task memory is transferred and the target task starts with a fresh memory filesystem. The LLM temperature is set to 1.0 for all models used.

For inference, we exclude harness states produced during initialization and select one state from the remaining harness state Pareto archive using normalized Euclidean distance to the empirical ideal point. Let $\mathcal{H}$ denote the eligible harness states and let $\mathbf{s}(H)=(s_1(H),\ldots,s_D(H))$ denote the lower-is-better harness score vector. For each dimension,
$$
z_d^{\mathrm{ideal}}=\min_{H\in\mathcal{H}}s_d(H),
\qquad
z_d^{\mathrm{nadir}}=\max_{H\in\mathcal{H}}s_d(H).
$$
We normalize each harness score as
$$
\tilde{s}_d(H)=
\frac{s_d(H)-z_d^{\mathrm{ideal}}}
{z_d^{\mathrm{nadir}}-z_d^{\mathrm{ideal}}},
$$
with $\tilde{s}_d(H)=0$ when the denominator is zero, and select
$$
H^*
=
\arg\min_{H\in\mathcal{H}}
\sqrt{
\sum_{d=1}^{D}
\tilde{s}_d(H)^2
}.
$$
Thus, the ideal point is empirical, formed from the best observed value of each harness score dimension among the eligible harness states. The selected harness is subsequently fixed during inference on both the source task and the related task used for intra-family transfer.

\suppheading{Traditional MOEA baselines.}
We implement SEMO, NSGA-II~\cite{deb2002fast},
MOEA/D~\cite{zhang2007moea}, and
PFG-MOEA~\cite{xu2023pareto} within our evaluation pipeline,
using task-specific representations and variation operators.
SEMO uses the fixed seed heuristic operator provided for operator design, while NSGA-II uses the corresponding seed operator
combination without modification by an LLM.
MOEA/D reuses task-specific seed operators, applying segment relocation, reversal, and multi-segment modification to Bi-TSP and Tri-TSP, relocation, swap, and segment reversal to Bi-CVRP and Bi-SPDP, and bit-flip mutation, uniform crossover, and repair-based mutation to Bi-KP and Bi-MDKP.
For PFG-MOEA, we use partially mapped crossover and swap
mutation on Bi-TSP, Tri-TSP, and Bi-CVRP, and uniform
crossover followed by bit-flip mutation and feasibility
repair on Bi-KP and Bi-MDKP. These operator types follow
the permutation and binary variation choices described
by MPaGE~\cite{hieu2026pareto}.
For Bi-SPDP, PFG-MOEA instead reuses the task-specific
relocation, swap, and reversal operators with
signed-permutation repair.
The task-specific decoders and feasibility safeguards
are retained in all implementations. For Bi-FJSP and Tri-FJSP, NSGA-II, MOEA/D, and PFG-MOEA
use the seed operation-priority crossover and mutation,
followed by machine-assignment crossover and mutation.

\suppheading{LLM-based AOD baselines.}
We preserve the search mechanisms and hyperparameters of the original LLM-based AOD frameworks, with two common controls for comparison. The heuristic or operator population size is set to 20 for EoH~\cite{liu2024evolution}, MEoH~\cite{yao2025multi}, MPaGE~\cite{hieu2026pareto}, and E2OC~\cite{qiu2026e2oc}, matching the population size used during \hemo inference. Each method terminates when the same maximum number of generated operator candidates is reached. All CPU-based evaluations were conducted on AMD EPYC 9555 processors. Traditional MOEAs and LLM-based AOD methods were configured to evaluate five instances of the same problem size in parallel. For final testing, every solver configuration is independently evaluated five times, and the mean performance over these runs is used as its reported performance to reduce stochastic variation from the underlying MOEA.

\suppheading{HEMO self-evolution cost breakdown.}
Harness self-evolution introduces a one-time computational cost from evaluating candidate harness states. For Bi-KP with the SEMO backbone, self-evolution on the source task requires approximately 37.67 wall-clock hours and \$4.22 in GPT-4.1-nano API calling cost on average. For Bi-CVRP with NSGA-II, the corresponding process requires approximately 80.08 hours and \$12.67 in API calling cost on average. Across all four source-task settings and independent runs, the largest wall-clock runtime is about 216.56 hours and the largest GPT-4.1-nano token cost is about \$29.52. The Claude Opus 4.8 Proposer and GPT-5 Realizer incur average costs of approximately \$12.53 and \$21.80 per complete run, respectively, across all source tasks.
This self-evolution cost is a one-time investment for obtaining the evolved harness and accumulated optimization memory. The resulting harness is subsequently reused during inference and intra-family transfer without repeating harness self-evolution, reducing the need to rediscover favorable quality--runtime trade-off regions from scratch.

\section{Additional Experiments}
\label{app:additional_experiments}

\subsection{Comparison with Expert-Designed TSP Operators}
\label{app:expert_tsp_operators}

To further assess the quality of the operator combinations produced by
\hemo, we compare the best-HV operator combinations of NSGA-II designed by \hemo (\hemo-NSGA-II) with several commonly used
expert-designed TSP operators and their sequential compositions.
We do not include OX-based combinations, as OX is outside the three
operator interfaces considered in our NSGA-II TSP backbone. Moreover,
the expert comparison in E2OC~\cite{qiu2026e2oc} shows that augmenting
OX and swap with additional local-search operators generally degrades
performance relative to the corresponding expert configurations. We therefore evaluate the three operators individually,
three sequential compositions, E2OC, and the operator combination
generated by \hemo on Bi-TSP with $N\in\{20,50,100\}$.

\begin{table}[t]
\centering
\scriptsize
\setlength{\tabcolsep}{3.2pt}
\renewcommand{\arraystretch}{1.08}
\caption{
Comparison of HEMO-NSGA-II with expert-designed TSP operators and
operator combinations. HV and runtime are reported for 50 test
instances at each problem size. Mean HV averages the three problem
sizes. Speedup is computed from the mean runtime across the three
sizes, using the slowest method as $1\times$. Best values in each
column are shown in \textbf{bold}.
}
\label{tab:expert_tsp_operators}

\begin{tabular}{lrrrrrrrr}
\toprule
&
\multicolumn{2}{c}{$N=20$} &
\multicolumn{2}{c}{$N=50$} &
\multicolumn{2}{c}{$N=100$} &
\multicolumn{2}{c}{Overall} \\
\cmidrule(lr){2-3}
\cmidrule(lr){4-5}
\cmidrule(lr){6-7}
\cmidrule(lr){8-9}
\textbf{Method}
& HV$\uparrow$ & Time$\downarrow$
& HV$\uparrow$ & Time$\downarrow$
& HV$\uparrow$ & Time$\downarrow$
& Mean HV$\uparrow$ & Speedup$\uparrow$ \\
\midrule

NSGA-II + 2-opt
& \textbf{0.6241} & 257.486
& 0.4607 & 249.499
& 0.3549 & 251.704
& 0.4799 & 1.531$\times$ \\

NSGA-II + 3-opt
& 0.6152 & 260.655
& 0.4607 & 253.314
& 0.3549 & 249.597
& 0.4769 & 1.521$\times$ \\

NSGA-II + Or-opt
& 0.6150 & 266.671
& 0.3647 & 262.096
& 0.2832 & 272.126
& 0.4210 & 1.450$\times$ \\

NSGA-II + 2opt-3opt-oropt
& 0.5469 & 257.407
& 0.3612 & 252.412
& 0.2836 & 244.447
& 0.3972 & 1.540$\times$ \\

NSGA-II + 3opt-2opt-oropt
& 0.5474 & 175.315
& 0.3605 & 166.707
& 0.2782 & 205.978
& 0.3954 & 2.120$\times$ \\

NSGA-II + oropt-2opt-3opt (seed)
& 0.5476 & \textbf{160.376}
& 0.3613 & \textbf{158.706}
& 0.2835 & 198.335
& 0.3975 & \textbf{2.245$\times$} \\

\midrule

NSGA-II + E2OC
& 0.6202 & 395.898
& 0.5013 & 380.750
& 0.3694 & 384.861
& 0.4970 & 1.000$\times$ \\

\rowcolor{hemoshade}
NSGA-II + \textbf{\hemo}
& 0.6158 & 210.879
& \textbf{0.5105} & 202.730
& \textbf{0.3837} & \textbf{194.252}
& \textbf{0.5033} & 1.911$\times$ \\

\bottomrule
\end{tabular}
\end{table}

\begin{table*}[t]
\centering
\scriptsize
\setlength{\tabcolsep}{3pt}
\renewcommand{\arraystretch}{1.10}
\caption{Additional FJSP results with the SEMO backbone.
ID comprises 15 synthetic instances and OOD comprises 15 Brandimarte instances.
Time is in seconds per instance.
Best displayed values are shown in \textbf{bold},
including ties, and ``--'' denotes runs exceeding the time limit.}
\label{tab:additional_semo_fjsp}

\begin{tabular*}{\textwidth}{@{\extracolsep{\fill}}l*{12}{r}@{}}
\toprule
& \multicolumn{6}{c}{Bi-FJSP}
& \multicolumn{6}{c}{Tri-FJSP} \\
\cmidrule(lr){2-7}\cmidrule(lr){8-13}
& \multicolumn{2}{c}{ID}
& \multicolumn{2}{c}{OOD}
& \multicolumn{2}{c}{All}
& \multicolumn{2}{c}{ID}
& \multicolumn{2}{c}{OOD}
& \multicolumn{2}{c}{All} \\
\cmidrule(lr){2-3}\cmidrule(lr){4-5}\cmidrule(lr){6-7}
\cmidrule(lr){8-9}\cmidrule(lr){10-11}\cmidrule(lr){12-13}
Method
& HV$\uparrow$ & Time$\downarrow$
& HV$\uparrow$ & Time$\downarrow$
& HV$\uparrow$ & Time$\downarrow$
& HV$\uparrow$ & Time$\downarrow$
& HV$\uparrow$ & Time$\downarrow$
& HV$\uparrow$ & Time$\downarrow$ \\
\midrule

SEMO
& 0.7430 & \textbf{1.482}
& 0.7789 & 14.672
& 0.7610 & 8.077
& 0.5434 & \textbf{5.014}
& 0.3210 & 12.117
& 0.4322 & 8.566 \\

NSGA-II
& 0.7166 & 1339.077
& 0.5915 & 1295.714
& 0.6540 & 1317.396
& 0.3066 & 1293.551
& 0.1784 & 1183.967
& 0.2425 & 1238.759 \\

MOEA/D
& \textbf{0.8738} & --
& 0.7801 & --
& 0.8269 & --
& 0.5516 & --
& 0.3293 & --
& 0.4404 & -- \\

PFG-MOEA
& 0.8565 & 287.431
& 0.7535 & 401.254
& 0.8050 & 344.343
& 0.5447 & 589.162
& 0.3218 & 879.100
& 0.4333 & 734.131 \\

\midrule

EoH
& 0.8226 & 4.591
& 0.7824 & 14.657
& 0.8025 & 9.624
& 0.5533 & 9.233
& 0.3259 & 21.124
& 0.4396 & 15.179 \\

MEoH
& 0.8421 & 2.985
& 0.7838 & 14.960
& 0.8130 & 8.973
& \textbf{0.5563} & 7.545
& 0.3230 & \textbf{9.253}
& 0.4397 & 8.399 \\

MPaGE
& 0.8735 & 3.754
& 0.7894 & 16.534
& \textbf{0.8315} & 10.144
& 0.5516 & 7.369
& 0.3311 & 11.129
& 0.4414 & 9.249 \\

\rowcolor{hemoshade}
\hemo
& 0.8651 & 2.035
& \textbf{0.7942} & \textbf{13.254}
& 0.8296 & \textbf{7.645}
& 0.5514 & 6.651
& \textbf{0.3369} & 9.775
& \textbf{0.4441} & \textbf{8.213} \\

\bottomrule
\end{tabular*}
\end{table*}

Table~\ref{tab:expert_tsp_operators} shows that \hemo achieves
the highest HV at $N=50$ and $N=100$, while 2-opt remains slightly
better at $N=20$. Averaged across the three problem sizes, \hemo
obtains the highest HV, clearly exceeding the
strongest expert operator, 2-opt, and outperforming E2OC. This result
is notable because several manually composed operator sequences perform
substantially worse than the strongest individual operators, indicating
that simply combining multiple neighborhood operators does not guarantee
better search performance.

The operator combination designed by HEMO also achieves a favorable
runtime profile. It achieves a $1.911\times$ speedup relative to the slowest method, while attaining a runtime close to that of the fastest operator combination at the largest problem size. Some expert combinations execute faster,
but exhibit considerably lower HV. These results provide additional
evidence that \hemo can discover
operator combinations that remain competitive with strong expert
operators in both solution quality and computational efficiency.

\subsection{Additional Results with the SEMO Backbone}
\label{app:additional_semo_fjsp}

Table~\ref{tab:additional_semo_fjsp} reports additional
results on Bi-FJSP and Tri-FJSP.
\hemo achieves the highest OOD HV on both tasks.
On Bi-FJSP, MPaGE achieves a slightly higher overall HV,
while \hemo has the lowest runtime among the LLM-based
AOD methods across all reported splits.
On Tri-FJSP, \hemo achieves the highest overall HV and
the lowest overall runtime among the LLM-based AOD
methods, although MEoH is faster on the OOD instances.
These results highlight the effectiveness of the
learned heuristics on the OOD instances, despite
not achieving the highest HV on the design distribution used during harness self-evolution.

\subsection{Additional Results with the NSGA-II Backbone}
\label{app:additional_nsga2_results}

Table~\ref{tab:additional_nsga2_results} reports results with the NSGA-II backbone on
the multi-objective routing tasks and the knapsack tasks.
Expert, E2OC, and \hemo are compared within the same
NSGA-II backbone using HV, IGD, and runtime. For the NSGA-II experiments on Bi-KP and Bi-MDKP,
each objective is the negative selected profit divided
by the total available profit for that objective:
\begin{equation}
    f_k(\mathbf{x})
    =
    -\frac{\sum_{i=1}^{n} p_i^{(k)}x_i}
    {\sum_{i=1}^{n} p_i^{(k)}},
    \qquad k\in\{1,2\}.
    \label{eq:nsga2_knapsack_normalization}
\end{equation}
HV is computed in this normalized minimization space
with reference point $\mathbf{r}=(0,0)$.
The bounding ideal point is $\mathbf{z}=(-1,-1)$.
This protocol is shared by all three methods and differs from the protocol used in the SEMO knapsack
experiments. Their HV values are therefore not directly comparable across backbones, and the test instances also follow different generation procedures.

\begin{table*}[t]
\centering
\scriptsize
\setlength{\tabcolsep}{3pt}
\renewcommand{\arraystretch}{1.10}
\caption{Additional NSGA-II-based results, averaged over
50 test instances per setting.
Time is in seconds per instance.
Best displayed values are shown in \textbf{bold} and
\hemo columns are shaded.}
\label{tab:additional_nsga2_results}

\begin{tabular*}{\textwidth}{
@{\extracolsep{\fill}}l rrr rrr
>{\columncolor{hemoshade}}r
>{\columncolor{hemoshade}}r
>{\columncolor{hemoshade}}r@{}
}
\toprule
& \multicolumn{3}{c}{Expert}
& \multicolumn{3}{c}{E2OC}
& \multicolumn{3}{>{\columncolor{hemoshade}}c}{\hemo} \\
\cmidrule(lr){2-4}
\cmidrule(lr){5-7}
\cmidrule(lr){8-10}
Setting
& HV$\uparrow$ & IGD$\downarrow$ & Time$\downarrow$
& HV$\uparrow$ & IGD$\downarrow$ & Time$\downarrow$
& HV$\uparrow$ & IGD$\downarrow$ & Time$\downarrow$ \\
\midrule

Bi-CVRP 20
& 0.6001 & \textbf{0.011265} & 173.681
& 0.5953 & 0.012716 & 139.499
& \textbf{0.6011} & 0.011267 & \textbf{119.408} \\

Bi-CVRP 50
& 0.5232 & 0.031433 & \textbf{243.641}
& 0.5179 & 0.039418 & 279.866
& \textbf{0.5467} & \textbf{0.012060} & 292.293 \\

Bi-CVRP 100
& 0.4579 & 0.095994 & \textbf{235.605}
& 0.5138 & 0.043941 & 1651.226
& \textbf{0.5457} & \textbf{0.014028} & 1361.897 \\

\midrule

Bi-SPDP 20
& 0.7447 & 0.036631 & \textbf{154.625}
& \textbf{0.7671} & 0.012824 & 158.575
& 0.7670 & \textbf{0.011057} & 162.802 \\

Bi-SPDP 50
& 0.7951 & 0.041620 & 165.747
& 0.8185 & 0.020679 & \textbf{163.603}
& \textbf{0.8268} & \textbf{0.015460} & 167.782 \\

Bi-SPDP 100
& 0.8010 & 0.024209 & \textbf{176.933}
& 0.8128 & 0.015857 & 195.670
& \textbf{0.8200} & \textbf{0.011203} & 184.557 \\

\midrule

Bi-KP 50
& 0.6237 & 0.009070 & 396.716
& 0.6323 & \textbf{0.000268} & \textbf{139.499}
& \textbf{0.6380} & 0.000372 & 158.701 \\

Bi-KP 100
& 0.6039 & 0.019835 & 350.862
& 0.6299 & 0.001442 & \textbf{156.866}
& \textbf{0.6388} & \textbf{0.000511} & 158.013 \\

Bi-KP 200
& 0.5914 & 0.028303 & 282.454
& 0.6353 & 0.000849 & 152.226
& \textbf{0.6467} & \textbf{0.000511} & \textbf{146.383} \\

\midrule

Bi-MDKP 50
& 0.5083 & 0.004763 & 397.447
& \textbf{0.5153} & \textbf{0.001644} & 254.650
& 0.5148 & 0.002123 & \textbf{172.544} \\

Bi-MDKP 100
& 0.5011 & 0.005855 & 355.499
& \textbf{0.5148} & 0.001965 & 289.035
& 0.5138 & \textbf{0.001831} & \textbf{192.561} \\

Bi-MDKP 200
& 0.4891 & 0.007465 & 279.699
& 0.4957 & 0.003953 & 379.050
& \textbf{0.5034} & \textbf{0.001404} & \textbf{187.168} \\

\bottomrule
\end{tabular*}
\end{table*}

\hemo achieves the highest HV in nine of the twelve
settings and the lowest IGD in nine settings.
On Bi-CVRP, it achieves the highest HV at every size
and the lowest IGD at sizes 50 and 100.
On Bi-SPDP, \hemo still achieves the lowest IGD at all
three sizes and the highest HV at sizes 50 and 100, though E2OC has a slightly higher HV at size 20.
On Bi-KP, \hemo achieves the highest HV at all sizes,
while E2OC achieves the lowest IGD at size 50 and
\hemo achieves the lowest IGD at sizes 100 and 200.
On Bi-MDKP, E2OC has the highest HV at sizes 50 and 100
and the lowest IGD at size 50.
\hemo achieves the lowest IGD at sizes 100 and 200,
the highest HV at size 200, and the lowest runtime
at all three sizes.
Overall, both quality indicators generally favor \hemo,
highlighting its effectiveness and generalization ability across both routing and knapsack tasks.


\subsection{Visualization of Obtained Solutions}
\label{app:solution_space_comparison}

\begin{figure*}[p]
    \centering
    \includegraphics[
        width=\textwidth,
        height=0.86\textheight,
        keepaspectratio
    ]{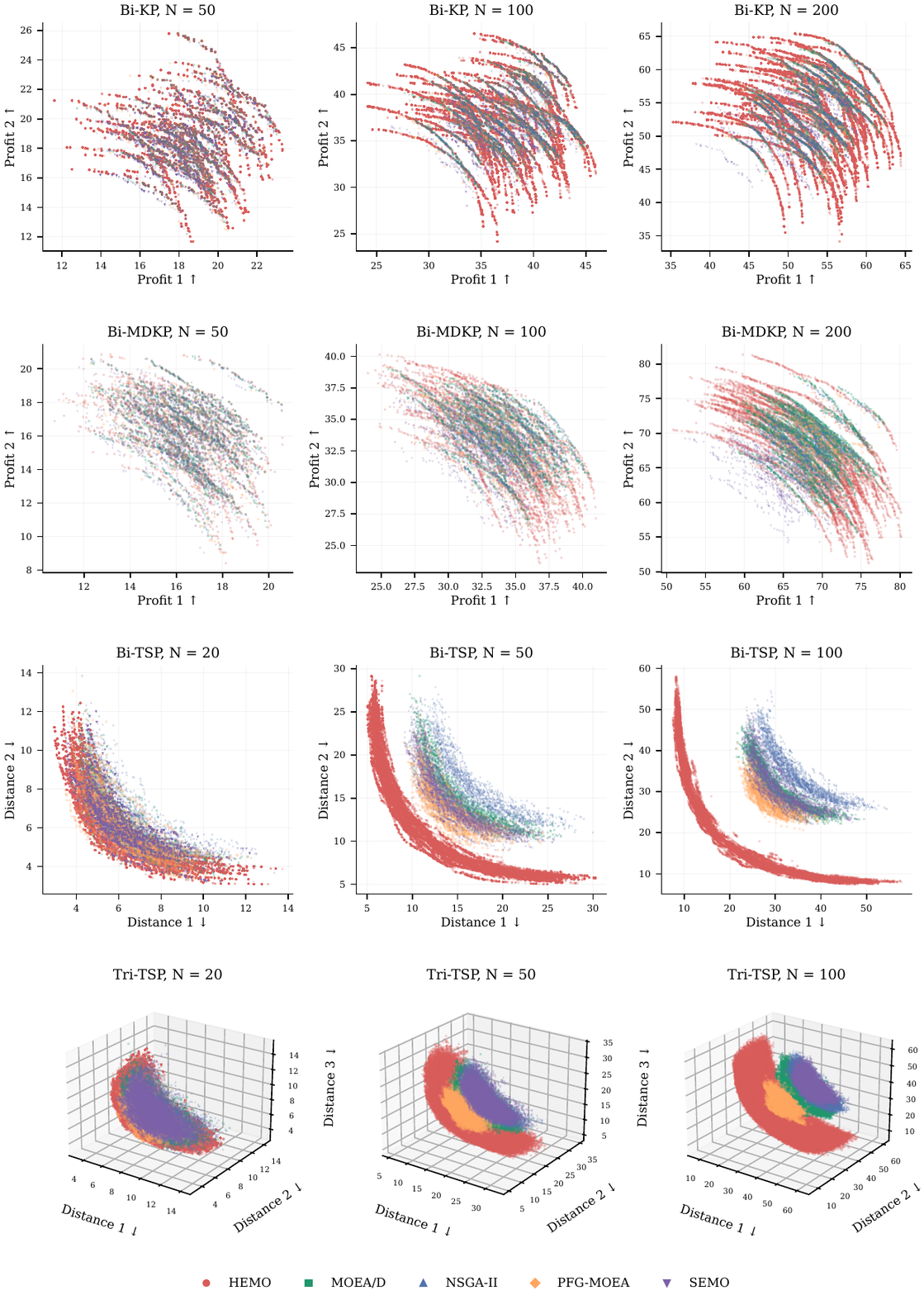}
    \caption{
    Final solution archives of \hemo-SEMO and
    expert-designed baselines across 50 test instances
    per setting.
    Rows correspond to Bi-KP, Bi-MDKP, Bi-TSP, and Tri-TSP.
    Columns correspond to increasing problem sizes.
    Knapsack objectives are shown as positive profits
    (higher is better), and TSP objectives as distances
    (lower is better).
    All saved final archive points are retained.
    }
    \label{fig:semo_solution_fronts}
\end{figure*}

\begin{figure*}[p]
    \centering
    \includegraphics[
        width=\textwidth,
        height=0.86\textheight,
        keepaspectratio
    ]{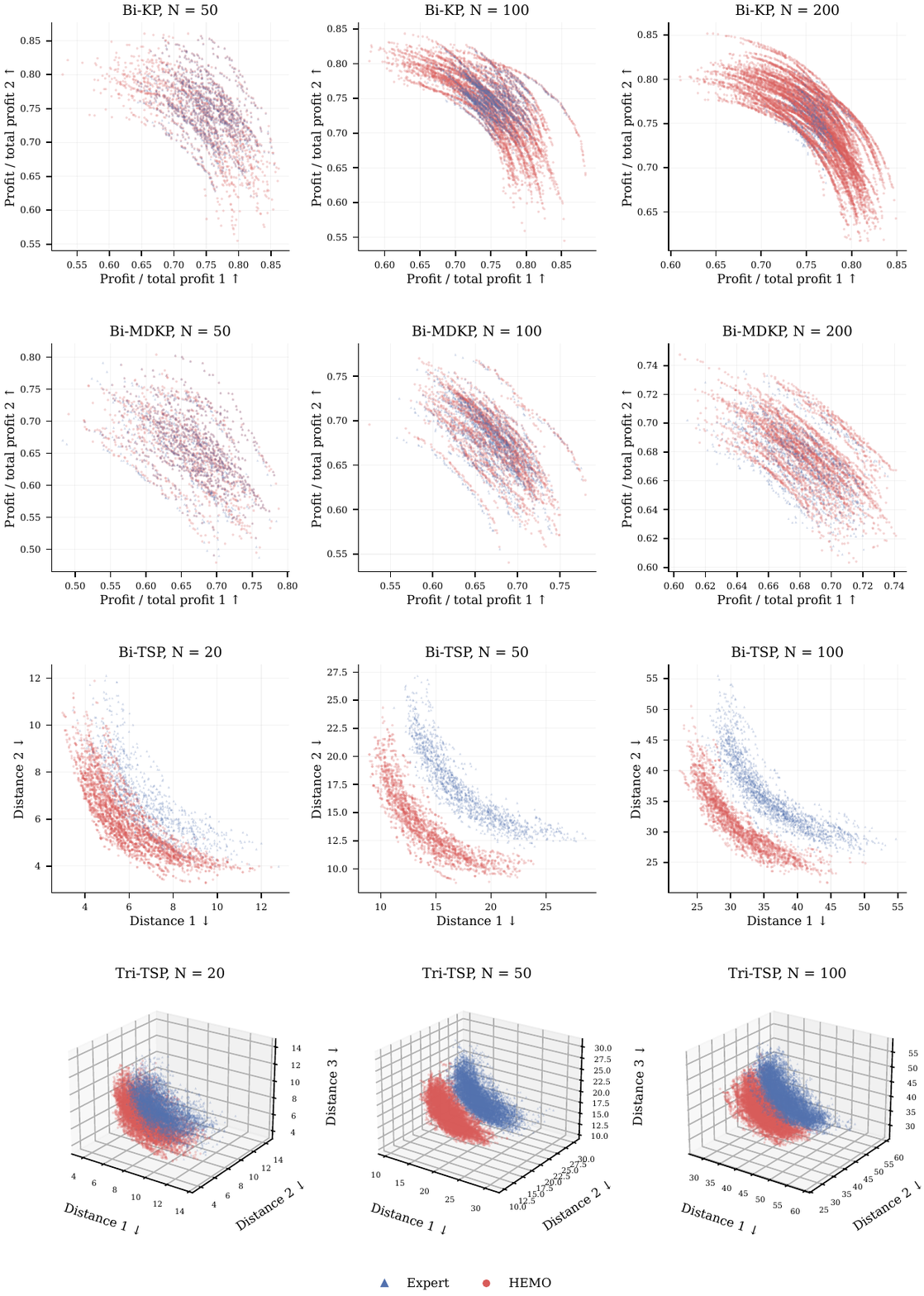}
    \caption{
    Final solution archives of operator combinations designed by \hemo and
    expert-designed operator combinations under NSGA-II, using matched population and generation budgets.
    Each panel includes 50 test instances.
    Rows correspond to Bi-KP, Bi-MDKP, Bi-TSP, and Tri-TSP.
    Columns correspond to increasing problem sizes.
    Knapsack profits are divided by the total available profit of the corresponding objective (higher is better).
    TSP objectives are distances (lower is better).
    All saved final archive points are retained.
    }
    \label{fig:nsga2_solution_fronts}
\end{figure*}

Figures~\ref{fig:semo_solution_fronts}
and~\ref{fig:nsga2_solution_fronts} compare the final solution fronts
obtained using operators evolved by \hemo{} with those produced by
traditional MOEAs on Bi-KP, Bi-MDKP, Bi-TSP, and Tri-TSP at three
problem sizes. These comparisons complement
the aggregate HV and IGD results by illustrating the
objective trade-offs achieved by the designed operators
on both source and transfer tasks.

With the SEMO backbone,
Figure~\ref{fig:semo_solution_fronts} compares \hemo
with SEMO, NSGA-II, MOEA/D, and PFG-MOEA.
The advantage is particularly visible on the TSP tasks:
\hemo reaches regions with shorter distances across
objectives, with a clear separation from several baselines
at larger problem sizes.
The knapsack solution distributions overlap more strongly,
showing that the advantage depends on the specific tasks.

With the NSGA-II backbone,
Figure~\ref{fig:nsga2_solution_fronts} compares
\hemo-designed operators with the expert-designed
operator combination.
The improved solution distributions on Bi-TSP and Tri-TSP
show that the learned operators remain effective when
transferred from two to three objectives.
On Bi-KP and Bi-MDKP, the distributions overlap more,
while the quantitative results show higher HV for \hemo
at every displayed size.
Together, these results illustrate the ability of
\hemo-designed heuristics to discover competitive trade-offs
beyond those obtained with fixed expert-designed heuristics.

\clearpage
\clearpage

\clearpage

\section{Representative Discovered Heuristics and Operator Combinations}
\label{app:discovered_algorithms}

In this section, we give representative examples of two LLM-evolved
operator designs, with Bi-TSP under SEMO and Bi-KP under NSGA-II
providing comparisons with MPaGE and E2OC, respectively. For HEMO, each
box pairs a concise natural-language explanation with the complete
executable heuristic or operator implementation. Code formatting and
comments are condensed without changing the algorithmic logic. For each
candidate, we provide its design idea, reported score, unique identifier,
and an excerpt of the algorithm code. These examples characterize
particular discovered designs, while aggregate performance is evaluated
in the broader experimental results.

\definecolor{hemoCaseBackground}{RGB}{244,250,245}
\definecolor{hemoCaseFrame}{RGB}{91,139,103}
\definecolor{hemoCaseTitleBackground}{RGB}{225,240,227}
\definecolor{hemoCaseTitleText}{RGB}{38,85,51}
\newtcolorbox{hemoalgorithmbox}[1][]{%
  enhanced, breakable,
  colback=hemoCaseBackground,
  colframe=hemoCaseFrame,
  coltitle=hemoCaseTitleText,
  fonttitle=\bfseries\small\sffamily,
  attach boxed title to top left={yshift=-2mm,xshift=4mm},
  boxed title style={colback=hemoCaseTitleBackground,
    colframe=hemoCaseFrame,boxrule=0.4pt,arc=1.5pt},
  boxrule=0.5pt,arc=2pt,
  left=6pt,right=6pt,top=6pt,bottom=4pt,
  fontupper=\small,
  title={#1}
}

\providecommand{\HEMOcasecodefont}{%
  \normalfont
  \ifcsname ver@inconsolata.sty\endcsname
    \fontfamily{zi4}%
  \else
    \fontfamily{cmtt}%
  \fi
  \linespread{1}\fontsize{7}{7.5}\selectfont
}
\lstdefinestyle{hemoCasePython}{
  language=Python,
  basicstyle=\HEMOcasecodefont,
  keywordstyle=\color{blue!55!black}\bfseries,
  commentstyle=\color{black!55},
  stringstyle=\color{green!35!black},
  numbers=none,
  columns=fullflexible,
  keepspaces=true,
  showstringspaces=false,
  showspaces=false,
  showtabs=false,
  breaklines=true,
  breakatwhitespace=false,
  breakautoindent=true,
  breakindent=1em,
  tabsize=4,
  emptylines=1,
  aboveskip=3pt,
  belowskip=0pt,
  xleftmargin=0pt,
  xrightmargin=0pt,
  frame=none,
  mathescape=false,
  texcl=false
}

\newenvironment{baselinealgorithmbox}[1][]{%
  \begingroup
  \definecolor{hemoCaseBackground}{RGB}{243,248,253}%
  \definecolor{hemoCaseFrame}{RGB}{100,145,185}%
  \definecolor{hemoCaseTitleBackground}{RGB}{222,236,249}%
  \definecolor{hemoCaseTitleText}{RGB}{35,75,115}%
  \begin{hemoalgorithmbox}[#1]%
}{%
  \end{hemoalgorithmbox}%
  \endgroup
}

\subsection{Bi-TSP with SEMO: HEMO and MPaGE}
\label{app:case_bitsp_semo}

Both selected heuristics first filter the solution archive by Pareto dominance and use randomly weighted, normalized tour costs to select a parent tour. They differ mainly in neighborhood construction, with MPaGE sampling from a broader menu, whereas HEMO pairs block relocation with segment reversal and explicitly removes duplicate neighbors. Neither implementation requires improvement over the parent, as each returns the best sampled neighbor with a swap fallback if it coincides with the parent. Thus, the comparison concerns how a bounded neighborhood is explored.

\begin{baselinealgorithmbox}[MPaGE / Bi-TSP / SEMO -- Design Idea]
\small
\textbf{Idea.} Draw scalarization weights from
$\operatorname{Dirichlet}(0.6,0.4)$ and select the best normalized
nondominated parent. Sample at most $\min(28,\max(6,n))$ trials from block
relocation, segment reversal, three-block reordering, single-node insertion,
prefix rotation, and a no-operation option. Rank valid sampled tours by the
same scalarized cost and return the best one, using a random swap fallback.
\par\smallskip
\textbf{Score:} $(\mathrm{HV},\mathrm{runtime}) = (0.629644,\,1.500210\,\mathrm{s})$.
\par\smallskip
\textbf{Core code excerpt.} Omitted portions are marked explicitly.
\begin{lstlisting}[style=hemoCasePython]
def select_neighbor(archive, instance, distance_matrix_1, distance_matrix_2):
    # Initialization, Pareto filtering and cost helpers omitted.
    ...
    w = np.random.dirichlet([0.6, 0.4])
    scalar_vals = []
    for idx in nondom_indices:
        o1 = (objs[idx][0] - min1) / range1
        o2 = (objs[idx][1] - min2) / range2
        scalar_vals.append(w[0] * o1 + w[1] * o2)
    # Parent extraction, safeguards and move helpers omitted.
    ...
    perturb_ops = [op_block_relocate, op_reverse_segment,
                   op_three_block_reorder, op_insertion,
                   op_rotate_prefix, lambda t: None]
    candidates = []
    trials = min(28, max(6, N))
    for _ in range(trials):
        op = random.choice(perturb_ops)
        cand = op(parent)
        # Validity checks and normalized cost calculation omitted.
        ...
        candidates.append((sc, cand))
    if candidates:
        candidates.sort(key=lambda x: x[0])
        best_candidate = candidates[0][1]
        best_scalar = candidates[0][0]
    # Return checks and swap fallback omitted.
    ...
\end{lstlisting}
\end{baselinealgorithmbox}

\begin{hemoalgorithmbox}[HEMO / Bi-TSP / SEMO -- Idea and Code]
\label{box:case_bitsp_hemo}
\small
\textbf{Idea.} Select a nondominated parent using normalized tour costs and
$\operatorname{Dirichlet}(0.5,0.5)$ weights. Run at most
$\min(14,\max(6,\lfloor 2n/3\rfloor))$ perturbation rounds, proposing a
block relocation and, when that proposal passes the checks, a reversal of
the same segment. Deduplicate tours and return the sampled neighbor with
the lowest scalarized cost; retain a swap fallback.
\par\smallskip
\textbf{Score:} $(\mathrm{HV},\mathrm{runtime}) = (0.631584,\,1.322606\,\mathrm{s})$.
\par\smallskip
\textbf{Candidate.} {\footnotesize\texttt{h\_239}}
\par\smallskip
\textbf{Complete implementation.}
\begin{lstlisting}[style=hemoCasePython]
def select_neighbor(archive, instance, distance_matrix_1, distance_matrix_2):
    N = int(distance_matrix_1.shape[0]) if distance_matrix_1 is not None else 0
    if N <= 0: return np.array([], dtype=int)
    if archive is None or len(archive) == 0:
        perm = np.arange(N, dtype=int)
        random.shuffle(perm)
        return perm
    sols = [np.array(sol, dtype=int) for sol, _ in archive]
    objs = [tuple(map(float, ob)) for _, ob in archive]; M = len(archive); dominated = [False] * M
    for i in range(M):
        for j in range(M):
            if i == j or dominated[i]: continue
            oi = objs[i]; oj = objs[j]
            if (oj[0] <= oi[0] and oj[1] <= oi[1]) and (oj[0] < oi[0] or oj[1] < oi[1]):
                dominated[i] = True
                break
    nondom_indices = [idx for idx, d in enumerate(dominated) if not d]
    if not nondom_indices:
        nondom_indices = list(range(M))
    min1 = min(objs[i][0] for i in nondom_indices); max1 = max(objs[i][0] for i in nondom_indices)
    min2 = min(objs[i][1] for i in nondom_indices); max2 = max(objs[i][1] for i in nondom_indices)
    range1 = max1 - min1; range2 = max2 - min2
    if range1 == 0:
        range1 = 1e-9
    if range2 == 0:
        range2 = 1e-9
    w = np.random.dirichlet([0.5, 0.5]); scalar_vals = []
    for idx in nondom_indices:
        o1 = (objs[idx][0] - min1) / range1; o2 = (objs[idx][1] - min2) / range2
        scalar_vals.append(w[0] * o1 + w[1] * o2)
    best_parent_index = nondom_indices[int(np.argmin(scalar_vals))]
    parent = sols[best_parent_index].copy()
    if parent.size != N:
        parent = np.arange(N, dtype=int)
        random.shuffle(parent)
    if N <= 3: return parent.astype(int)
    def tour_costs(tour):
        t = tour.astype(int); c1 = 0.0; c2 = 0.0
        for k in range(N - 1):
            a = int(t[k]); b = int(t[k + 1])
            c1 += distance_matrix_1[a, b]; c2 += distance_matrix_2[a, b]
        a = int(t[-1]); b = int(t[0])
        c1 += distance_matrix_1[a, b]; c2 += distance_matrix_2[a, b]
        return c1, c2
    c1_p, c2_p = tour_costs(parent); norm1_p = (c1_p - min1) / range1
    norm2_p = (c2_p - min2) / range2; current_scalar = w[0] * norm1_p + w[1] * norm2_p
    candidates = []; seen = set(); max_perturb = min(14, max(6, N * 2 // 3))
    for _ in range(max_perturb):
        if N < 4: break
        i = random.randint(1, N - 3); j = random.randint(i + 1, N - 1); block = parent[i:j + 1]
        rest = list(parent[:i]) + list(parent[j + 1:]); p = random.randint(0, len(rest))
        cand = np.array(rest[:p] + block.tolist() + rest[p:], dtype=int)
        if cand.size != N or len(set(cand.tolist())) != N: continue
        key = tuple(cand.tolist())
        if key in seen: continue
        seen.add(key)
        c1, c2 = tour_costs(cand); n1 = (c1 - min1) / range1; n2 = (c2 - min2) / range2
        sc = w[0] * n1 + w[1] * n2
        candidates.append((sc, cand))
        cand_rev = parent.copy(); cand_rev[i:j + 1] = cand_rev[i:j + 1][::-1]
        key2 = tuple(cand_rev.tolist())
        if key2 not in seen:
            c1r, c2r = tour_costs(cand_rev); nr1 = (c1r - min1) / range1
            nr2 = (c2r - min2) / range2; sc2 = w[0] * nr1 + w[1] * nr2
            candidates.append((sc2, cand_rev))
            seen.add(key2)
    best_candidate = parent; best_scalar = current_scalar
    if candidates:
        candidates.sort(key=lambda x: x[0])
        best_scalar, best_candidate = candidates[0]
    if best_candidate.size != N or len(set(best_candidate.tolist())) != N:
        best_candidate = parent
    if not np.array_equal(best_candidate, parent): return best_candidate.astype(int)
    a, b = random.sample(range(N), 2); tmp = parent.copy(); tmp[a], tmp[b] = tmp[b], tmp[a]
    return tmp.astype(int)
\end{lstlisting}
\end{hemoalgorithmbox}

Compared with MPaGE, \hemo{} uses a similar sampling budget to explore
two complementary changes to each sampled segment, moving it elsewhere
in the tour or reversing its order. This gives each segment two ways
to improve the tour, while deduplication avoids evaluating the same
neighbor repeatedly. Together with removing the explicit no-operation
option, this focuses computation on distinct alternatives. Symmetric
scalarization weights also give both objectives equal emphasis across
search steps. These mechanisms make more effective use of the sampling budget,
consistent with the selected heuristic's higher HV and lower runtime.

\clearpage
\subsection{Bi-KP with NSGA-II: HEMO and E2OC}
\label{app:case_bikp_nsga2}

The selected combinations contain three binary-chromosome operators:
\texttt{kp\_bitflip}, \texttt{kp\_crossover}, and
\texttt{kp\_repair\_mutation}. Both designs use profit and weight
information. The principal distinction is where randomized profit
scalarization enters the combination and whether repair preserves the
incoming chromosome. E2OC constructs new greedy solutions in its repair
operator, whereas HEMO combines scalarized construction with mutation and
uses its final repair operator to modify the incoming solution.

\begin{baselinealgorithmbox}[E2OC / Bi-KP / NSGA-II -- Combination Design]
\small
\textbf{Mutation.} Build a greedy core using summed-profit density, insert
core items when capacity permits, and prune overweight selections. With
probability $0.3$, perturb approximately $4\%$ of the bits and repair/refill.
\par\smallskip
\textbf{Crossover.} Use min--max-normalized summed-profit density as the
per-item probability of copying from the first parent. Generate the two
children with separate masks, then prune and greedily refill each child.
\par\smallskip
\textbf{Repair/mutation.} With valid problem data, rebuild each row from
an empty selection using profit weights $(\lambda,1-\lambda)$ with
$\lambda\sim U(0.2,0.8)$. Optionally perturb approximately $4\%$ of the bits,
then repair and refill. This branch reconstructs the solution rather than
retaining the input chromosome's selected items.
\par\smallskip
\textbf{Score:} $(\mathrm{HV},\mathrm{runtime}) = (0.620616,\,1.727510\,\mathrm{s})$ for the complete operator combination.
\par\smallskip
\textbf{Core code excerpt.} Omitted portions are marked explicitly.
\begin{lstlisting}[style=hemoCasePython]
def kp_bitflip(offspring_chromosome, weights=None, profits_1=None, profits_2=None, capacity=None):
    # Input handling, greedy-core construction and repair omitted.
    ...
    for r in range(pop):
        if rng.random() < 0.3:
            k = max(1, int(0.04 * L))
            idxs = rng.choice(L, size=min(k, L), replace=False)
            sol[r, idxs] = 1 - sol[r, idxs]
            # Capacity repair and density-guided refill omitted.
            ...
    # Shape restoration, return and exception handling omitted.
    ...

def kp_crossover(offspring_chromosome, weights=None, profits_1=None, profits_2=None, capacity=None):
    # Input/context preparation omitted.
    ...
    def cross(a, b):
        if density_available and w is not None and w.size == L:
            d = density.copy()
            dmin = d.min(); dmax = d.max()
            rngv = dmax - dmin
            if rngv > 0:
                dnorm = (d - dmin) / rngv
            else:
                dnorm = np.zeros_like(d)
            mask = rng.random(L) < dnorm
            child = np.where(mask, a, b)
        else:
            child = np.where(rng.random(L) < 0.5, a, b)
        # Binary conversion, repair, refill and return omitted.
        ...
    # Pair iteration, independent cross(a,b)/cross(b,a) calls omitted.
    ...

def kp_repair_mutation(offspring_chromosome, weights=None, profits_1=None, profits_2=None, capacity=None):
    # Input/context preparation omitted.
    ...
    if dens_ready:
        for r in range(pop):
            s1 = rng.random() * 0.6 + 0.2
            s2 = 1.0 - s1
            density_scalar = p1 * s1 + p2 * s2
            density_per_weight = density_scalar / np.maximum(w, 1e-12)
            order = np.argsort(-density_per_weight)
            totw = 0.0
            row = np.zeros(L, dtype=original.dtype)
            for idx in order:
                if totw + w[idx] <= cap:
                    row[idx] = 1
                    totw += float(w[idx])
            # Perturbation, repair/refill and output assignment omitted.
            ...
    # Missing-data fallback, shape restoration and return omitted.
    ...
\end{lstlisting}
\end{baselinealgorithmbox}

E2OC's final repair rebuilds from an empty
selection, discarding the mutation and crossover results. In \hemo{},
mutation mixes the incoming solution with a greedy selection under
a random objective preference, while repair preserves and refines
the result. Wider objective weights encourage exploration near either
objective extreme, and complementary crossover preserves both parents'
bit values across the two children before repair. Final repair fills
unused capacity and allows feasible exchanges favoring higher-density
items. Together, these choices explore diverse trade-offs while
retaining useful inherited combinations.

\begin{hemoalgorithmbox}[HEMO / Bi-KP -- kp\_bitflip]
\label{box:case_bikp_kp_bitflip}
\small
\textbf{Idea.} For each chromosome, draw $\lambda\sim U(0,1)$ and build a
greedy feasible selection ranked by scalarized profit per unit weight.
Mix it with the input through random union/intersection choices. If
overweight, remove items in ascending scalarized profit, followed by an
optional single-bit flip. Without problem data, flip a short contiguous block.
\par\smallskip
\textbf{Score:} $(\mathrm{HV},\mathrm{runtime}) = (0.624848,\,0.873728\,\mathrm{s})$ for the complete combination.
\par\smallskip
\textbf{Candidate.} {\footnotesize\texttt{h99\_kp\_bitflip\_001891}}
\par\smallskip
\textbf{Complete implementation.}
\begin{lstlisting}[style=hemoCasePython]
def kp_bitflip(offspring_chromosome, weights=None, profits_1=None, profits_2=None, capacity=None):
    original = np.asarray(offspring_chromosome)
    if original.ndim == 1:
        original = original.reshape(1, -1); flatten_out = True
    else:
        flatten_out = False
    pop, chrom_len = original.shape; binary = (original > 0.5).astype(original.dtype, copy=True)
    try:
        rng = np.random.default_rng(); w = None; p1 = None; p2 = None; cap = None; has_data = False
        if weights is not None and profits_1 is not None and profits_2 is not None and capacity is not None:
            w = np.asarray(weights, dtype=float).reshape(-1)
            p1 = np.asarray(profits_1, dtype=float).reshape(-1)
            p2 = np.asarray(profits_2, dtype=float).reshape(-1); cap = float(capacity)
            if w.size == chrom_len and p1.size == chrom_len and p2.size == chrom_len:
                has_data = True
        if not has_data:
            for r in range(pop):
                if chrom_len < 2: continue
                L = max(1, int(round(chrom_len * rng.uniform(0.04, 0.12))))
                start = rng.integers(0, max(1, chrom_len - L + 1)); block = binary[r, start:start+L]
                binary[r, start:start+L] = 1 - block
            return binary if not flatten_out else binary.ravel()
        eps = 1e-12
        for r in range(pop):
            lam = rng.random(); scalar = lam * p1 + (1 - lam) * p2
            order = np.argsort(-scalar / (w + eps)); greedy = np.zeros(chrom_len, dtype=float)
            total_w = 0.0
            for idx in order:
                if total_w + w[idx] <= cap + 1e-12:
                    greedy[idx] = 1.0; total_w += float(w[idx])
            mix = binary[r].astype(float); prob = rng.random(chrom_len)
            for i in range(chrom_len):
                if prob[i] < 0.5:
                    mix[i] = max(mix[i], greedy[i])
                else:
                    mix[i] = min(mix[i], 1.0 if greedy[i] > 0.5 else 0.0)
            row = (mix > 0.5).astype(original.dtype, copy=False); total_w = float(np.dot(row, w))
            if total_w > cap + 1e-12:
                inc = np.flatnonzero(row > 0.5)
                if inc.size > 0:
                    dens = scalar[inc]; order2 = inc[np.argsort(dens)]
                    for idx in order2:
                        row[idx] = 0.0; total_w -= float(w[idx])
                        if total_w <= cap + 1e-12: break
            if chrom_len >= 2 and rng.random() < 0.15:
                idx = int(rng.integers(0, chrom_len)); row[idx] = 1 - int(row[idx])
            binary[r] = row
        result = binary.astype(original.dtype, copy=False)
        return result if not flatten_out else result.ravel()
    except Exception:
        return np.asarray(offspring_chromosome).copy()
\end{lstlisting}
\end{hemoalgorithmbox}

\begin{hemoalgorithmbox}[HEMO / Bi-KP -- kp\_crossover]
\label{box:case_bikp_kp_crossover}
\small
\textbf{Idea.} Transform median-centered summed-profit density through
a sigmoid and use the resulting probabilities to sample a shared mask.
Produce paired children with complementary parent assignments, repair
overweight rows by dropping low-density items and refilling, and allow rare
single-bit perturbations. An odd final row is paired with the first row.
\par\smallskip
\textbf{Score:} $(\mathrm{HV},\mathrm{runtime}) = (0.624848,\,0.873728\,\mathrm{s})$ for the complete combination.
\par\smallskip
\textbf{Candidate.} {\footnotesize\texttt{h99\_kp\_crossover\_000640}}
\par\smallskip
\textbf{Complete implementation.}
\begin{lstlisting}[style=hemoCasePython]
def kp_crossover(offspring_chromosome, weights=None, profits_1=None, profits_2=None, capacity=None):
    original = np.asarray(offspring_chromosome)
    if original.ndim == 1:
        original = original.reshape(1, -1); flatten_out = True
    else:
        flatten_out = False
    pop_size, chrom_length = original.shape
    offspring = original.copy().astype(original.dtype, copy=True)
    try:
        w = None; cap = None; density = None; med_density = None; std_density = 1.0
        has_density = False
        if weights is not None:
            w = np.asarray(weights, dtype=float).reshape(-1)
            if w.size == chrom_length:
                if capacity is not None:
                    cap = float(capacity)
                    if cap is not None:
                        if profits_1 is not None and profits_2 is not None:
                            p1 = np.asarray(profits_1, dtype=float).reshape(-1)
                            p2 = np.asarray(profits_2, dtype=float).reshape(-1)
                            if p1.size == chrom_length and p2.size == chrom_length:
                                density = (p1 + p2) / np.maximum(w, 1e-12)
                                med_density = float(np.median(density))
                                std_density = float(np.std(density)) if density.size > 1 else 1.0
                                has_density = True
        rng = np.random.default_rng()
        def repair_row(row):
            if cap is None or w is None: return row
            total_w = float(np.dot(row, w))
            if total_w <= cap + 1e-12: return row
            inc = np.flatnonzero(row > 0.5)
            if inc.size == 0: return row
            if has_density:
                dens = density[inc]; order = inc[np.argsort(dens)]
            else:
                order = inc
            for idx in order:
                if total_w <= cap + 1e-12: break
                row[idx] = 0.0; total_w -= float(w[idx])
            remaining = cap - total_w
            if remaining > 1e-12:
                not_in = np.flatnonzero(row <= 0.5)
                if not_in.size > 0:
                    if has_density:
                        dens_not = density[not_in]; order_not = not_in[np.argsort(-dens_not)]
                    else:
                        order_not = not_in
                    for idx in order_not:
                        if w[idx] <= remaining + 1e-12:
                            row[idx] = 1.0; remaining -= float(w[idx]); total_w += float(w[idx])
                            if remaining <= 1e-12: break
            return row
        for i in range(0, pop_size - 1, 2):
            A = original[i].astype(float).copy(); B = original[i + 1].astype(float).copy()
            if chrom_length <= 0: continue
            if has_density:
                delta = density - med_density
                x = 1.0 / (1.0 + np.exp(-delta / (std_density + 1e-12)))
                mask = rng.random(chrom_length) < (0.5 + 0.5 * x)
            else:
                mask = rng.random(chrom_length) < 0.5
            C1 = np.where(mask, A, B); C2 = np.where(mask, B, A)
            C1 = (C1 > 0.5).astype(original.dtype, copy=False)
            C2 = (C2 > 0.5).astype(original.dtype, copy=False)
            if cap is not None and w is not None:
                C1 = repair_row(C1.astype(float)).astype(original.dtype, copy=False)
                C2 = repair_row(C2.astype(float)).astype(original.dtype, copy=False)
            if rng.random() < 0.05:
                idx = int(rng.integers(0, chrom_length)); C1[idx] = 1 - int(C1[idx])
            if rng.random() < 0.05:
                idx = int(rng.integers(0, chrom_length)); C2[idx] = 1 - int(C2[idx])
            offspring[i] = C1; offspring[i + 1] = C2
        if pop_size % 2 == 1:
            i0 = pop_size - 1; A = original[i0].astype(float).copy()
            B = original[0].astype(float).copy()
            if chrom_length > 0:
                if has_density:
                    delta = density - med_density
                    x = 1.0 / (1.0 + np.exp(-delta / (std_density + 1e-12)))
                    mask = rng.random(chrom_length) < (0.5 + 0.5 * x)
                else:
                    mask = rng.random(chrom_length) < 0.5
                C = np.where(mask, A, B); C = (C > 0.5).astype(original.dtype, copy=False)
                if cap is not None and w is not None:
                    C = repair_row(C.astype(float)).astype(original.dtype, copy=False)
                if rng.random() < 0.05:
                    idx = int(rng.integers(0, chrom_length)); C[idx] = 1 - int(C[idx])
                offspring[i0] = C
        result = offspring.astype(original.dtype, copy=False)
        return result if not flatten_out else result.ravel()
    except Exception:
        return np.asarray(offspring_chromosome).copy()
\end{lstlisting}
\end{hemoalgorithmbox}

\begin{hemoalgorithmbox}[HEMO / Bi-KP -- kp\_repair\_mutation]
\label{box:case_bikp_kp_repair_mutation}
\small
\textbf{Idea.} Retain the incoming binary selection, remove
low-density included items until capacity is satisfied, and fill remaining
capacity with high-density excluded items. With probability $0.15$, attempt
one capacity-feasible exchange between the lowest-density included item
and the highest-density excluded item.
\par\smallskip
\textbf{Score:} $(\mathrm{HV},\mathrm{runtime}) = (0.624848,\,0.873728\,\mathrm{s})$ for the complete combination.
\par\smallskip
\textbf{Candidate.} {\footnotesize\texttt{h99\_kp\_repair\_mutation\_000380}}
\par\smallskip
\textbf{Complete implementation.}
\begin{lstlisting}[style=hemoCasePython]
def kp_repair_mutation(offspring_chromosome, weights=None, profits_1=None, profits_2=None, capacity=None):
    original = np.asarray(offspring_chromosome)
    if original.ndim == 1:
        original = original.reshape(1, -1); flatten_out = True
    else:
        flatten_out = False
    binary = (original > 0.5).astype(original.dtype, copy=True)
    try:
        pop_size, chrom_len = binary.shape
        if chrom_len <= 0: return binary if not flatten_out else binary.ravel()
        cap = capacity; w = None
        if weights is not None:
            w = np.asarray(weights, dtype=float).reshape(-1)
        if w is None or w.size != chrom_len or cap is None:
            return binary if not flatten_out else binary.ravel()
        p1 = None; p2 = None; density = None
        if profits_1 is not None and profits_2 is not None:
            p1 = np.asarray(profits_1, dtype=float).reshape(-1)
            p2 = np.asarray(profits_2, dtype=float).reshape(-1)
            if p1.size == chrom_len and p2.size == chrom_len:
                density = (p1 + p2) / np.maximum(w, 1e-12)
        for r in range(pop_size):
            row = binary[r].astype(float); total_w = float(np.dot(row, w))
            if total_w > cap + 1e-12:
                inc = np.flatnonzero(row > 0.5)
                if inc.size > 0:
                    if density is not None:
                        order = inc[np.argsort(density[inc])]
                    else:
                        order = inc
                    for idx in order:
                        row[idx] = 0.0; total_w -= float(w[idx])
                        if total_w <= cap + 1e-12: break
            remaining = cap - total_w
            if remaining > 1e-12:
                not_in = np.flatnonzero(row <= 0.5)
                if not_in.size > 0:
                    if density is not None:
                        order_not = not_in[np.argsort(-density[not_in])]
                    else:
                        order_not = not_in
                    for idx in order_not:
                        if w[idx] <= remaining + 1e-12:
                            row[idx] = 1.0; remaining -= float(w[idx]); total_w += float(w[idx])
                            if remaining <= 1e-12: break
            if density is not None and np.random.rand() < 0.15:
                inc = np.flatnonzero(row > 0.5); not_in = np.flatnonzero(row <= 0.5)
                if inc.size > 0 and not_in.size > 0:
                    in_idx = int(inc[np.argmin(density[inc])])
                    out_idx = int(not_in[np.argmax(density[not_in])])
                    if density[out_idx] > density[in_idx] + 1e-12:
                        if total_w - w[in_idx] + w[out_idx] <= cap + 1e-12:
                            row[in_idx] = 0.0; row[out_idx] = 1.0
            binary[r] = row
        result = binary.astype(original.dtype, copy=False)
        return result if not flatten_out else result.ravel()
    except Exception:
        return np.asarray(offspring_chromosome).copy()
\end{lstlisting}
\end{hemoalgorithmbox}

\newpage

\end{document}